\documentclass[pdflatex,sn-mathphys-num]{sn-jnl}% Math and Physical Sciences Numbered Reference Style
\usepackage{graphicx}%
\usepackage{multirow}%
\usepackage{amsmath,amssymb,amsfonts}%
\usepackage{amsthm}%
\usepackage{mathrsfs}%
\usepackage[title]{appendix}%
\usepackage{xcolor}%
\usepackage{textcomp}%
\usepackage{manyfoot}%
\usepackage{booktabs}%
\usepackage{algorithm}%
\usepackage{algorithmicx}%
\usepackage{algpseudocode}%
\usepackage{listings}%
\usepackage{tabularx}%
\usepackage{dirtree}%
\usepackage{array}%
\newcolumntype{C}[1]{>{\centering\arraybackslash}p{#1}}
\theoremstyle{thmstyleone}%
\theoremstyle{thmstyletwo}%

\theoremstyle{thmstylethree}%

\begin{document}

\title[MultiAnimate: A Unified Framework for Controllable Multi-Character Animation]{MultiAnimate: A Unified Framework for Controllable Multi-Character Animation}


\author[1]{\fnm{Zhongyi} \sur{Zhang}}\email{ericzhang@mail.ustc.edu.cn}
\equalcont{These authors contributed equally to this work.}

\author[2]{\fnm{Guangyuan} \sur{Wang}}\email{yixuan.wgy@alibaba-inc.com}
\equalcont{These authors contributed equally to this work.}

\author[2]{\fnm{Li} \sur{Hu}}\email{hooks.hl@alibaba-inc.com}

\author[3]{\fnm{Tianyi} \sur{Wei}}
\email{tianyi.wei@ntu.edu.sg}

\author[2]{\fnm{Peng} \sur{Zhang}}
\email{futian.zp@alibaba-inc.com}

\author[1]{\fnm{Weiming} \sur{Zhang}}
\email{zhangwm@ustc.edu.cn}

\author[1]{\fnm{Nenghai} \sur{Yu}}
\email{ynh@ustc.edu.cn}

\author[2]{\fnm{Bang} \sur{Zhang}}
\email{zhangbang.zb@alibaba-inc.com}

\author*[1]{\fnm{Wenbo} \sur{Zhou}}
\email{WenboZhou@ustc.edu.cn}

\affil*[1]{\orgdiv{School of Cyber Science and Technology}, \orgname{University
of Science and Technology of China}, \orgaddress{\street{Fuxing Road 100}, \city{Hefei}, \postcode{230026}, \state{Anhui}, \country{China}}}

\affil[2]{\orgdiv{Tongyi Lab}, \orgname{Alibaba Group}, \orgaddress{\street{Dengcai Road 1008}, \city{Hangzhou}, \postcode{310023}, \state{Zhejiang}, \country{China}}}

\affil[3]{\orgdiv{College of Computing and Data Science}, \orgname{Nanyang Technological University}, \orgaddress{\street{Jurong West}, \city{Singapore}, \postcode{639798}, \state{Singapore}, \country{Singapore}}}


\abstract{Recent advances in generative models and technological innovations have significantly addressed the fundamental challenges of character image animation. However, existing approaches predominantly focus on character animation from a single reference image, substantially limiting their applicability in scenarios such as multiple character interaction animation. To fill this gap, this paper introduces MultiAnimate, a comprehensive framework that enables concurrent animation of multiple characters within a shared environment while preserving both identity consistency and spatial relationships. The framework achieves these objectives through multiple well-designed mechanisms. First, we incorporate an identity-specific reference net that enables appearance extraction from multiple reference images, distinguishing MultiAnimate from existing approaches constrained to single reference inputs. Second, we implement an identity-aware pose encoder to address the character-pose binding challenge, wherein an attention mechanism enables the network to accurately differentiate and process multiple pose sequences during generation. Third, we introduce an interaction guider module that enhances the framework's capability to handle complex inter-character interactions by leveraging character-specific mask information, serving as an optional component that refines the pose sequences. Extensive experiments and ablation analyses demonstrate our framework's superiority in multiple character animation, particularly in scenarios involving complex motion sequences.}




\maketitle

\section{Introduction}
Character image animation aims to generate video sequences of the characters from reference images and various control signals such as pose sequences and depth information, while maintaining consistent visual appearance throughout the generated video. This emerging field has garnered significant research attention due to its extensive practical applications across numerous domains, including animation production, cinematic content creation, and commercial advertising. 

Early approaches to character image animation predominantly relied on GAN-based feature map warping techniques~\cite{wang2022latent, Siarohin_2019_NeurIPS,Gao_2023_CVPR}. However, these methods were inherently limited by the fundamental constraints of GAN architectures, particularly their training instability. Recently, the emergence of diffusion models has catalyzed a paradigm shift in this field, leading to numerous innovative approaches~\cite{chang2023magicdance,hu2023animateanyone,mimicmotion2024,ma2023follow,xu2023magicanimate,wang2023disco,zhu2024champ,karras2023dreamposefashionimagetovideosynthesis,yoon2025tpctesttimeprocrustescalibration,kim2024tcananimatinghumanimages,wang2024unianimate} that achieve animation by incorporating information from the reference image into video diffusion models~\cite{blattmann2023stablevideodiffusionscaling, guo2023animatediff}. These approaches typically employ convolutional networks~\cite{hu2023animateanyone,mimicmotion2024} to compress and encode control signals into the latent space of the diffusion model, enabling precise control of character animation during the denoising process. Leveraging the robust generative capabilities of diffusion models, % combined with sophisticated architectural designs, 
these methods demonstrate exceptional video synthesis quality while maintaining both precise fidelity to input control signals and accurate preservation of reference image attributes. 
 
Despite significant advances in these diffusion-based animation approaches, the challenge of animating multiple characters simultaneously has still not been well resolved. Specifically, when attempting to animate multiple characters with their respective motion sequences within a shared environment, existing approaches encounter significant challenges, including appearance inconsistencies, inadequate identity preservation, and imprecise character interaction dynamics.

\begin{figure*}[h]
  \centering
  \includegraphics[width=\textwidth]{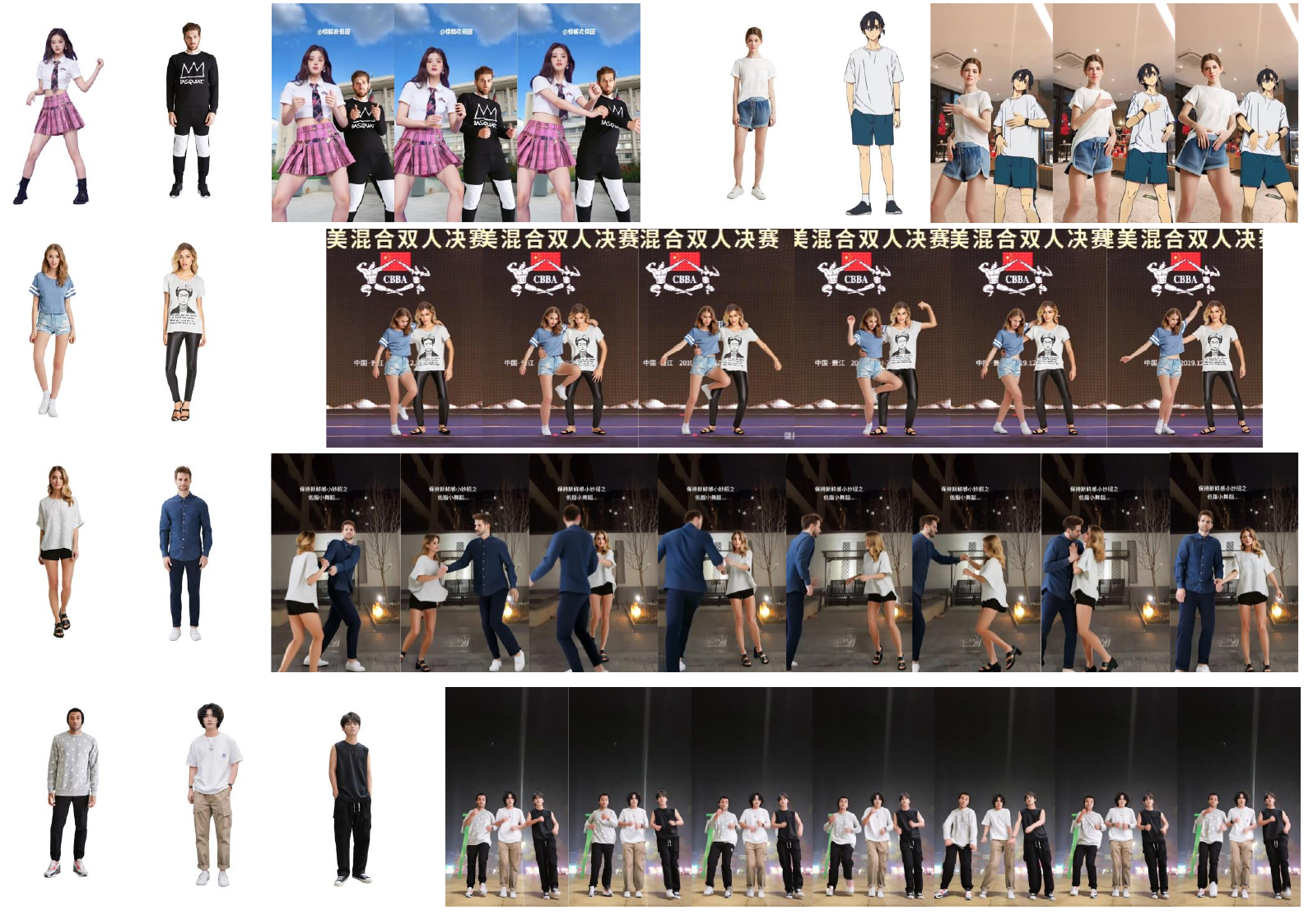}
  \caption{Performance of MultiAnimate. Our framework demonstrates successful multi-character animation across diverse scenarios, including dual-character scenarios (first row), complex spatial interactions (second row), maintains identity consistency during position interchange operations (third row), and multiple character scenarios (last row).}
  \label{fig:teaser}
\end{figure*}

Several recent approaches~\cite{xue2025multiplecharacterimageanimation,wang2025multiidentityhumanimageanimation} have partially addressed multiple character animation, however, these methods only work for some simple scenarios and may fail when encountering complex scenarios such as position interchange between characters. Moreover, a common limitation among these methods is their fundamental requirement that all characters should be presented within a single reference image, which makes it challenging for these methods to extend to multiple character scenarios due to the restricted selection of reference images.

In this paper, we present MultiAnimate, a comprehensive framework that addresses the aforementioned challenges. Our approach introduces novel framework capable of simultaneously animating multiple characters from distinct reference images and their corresponding pose sequences within a shared environment, while maintaining robust identity preservation throughout the generated video sequence. Several well-designed modules are proposed to achieve this, specifically: (1) \textit{Identity-specific reference net}. We incorporate positional encoding mechanisms with reference net to accurately extract appearance features while preserving distinct identity information from multiple reference images, thereby preventing feature attribution errors and maintaining distinct character identities. (2) \textit{Identity-aware pose encoder}. To solve the identity-pose binding problem of pose sequences, we utilize the cross-attention mechanism to inject identity information after extracting the features of pose sequences with convolutional networks to make the pose feature identity-aware. (3) \textit{Interaction Guider}. To enhance the management of inter-character spatial relationships during animation, we incorporate an interaction guider module that assists the model's generation process, which acts as an optional component when processing the pose sequences, enabling the occlusion awareness between pose sequences. In addition, we also introduce a background branch in the framework to achieve animation in the same scenario, which makes the whole framework more controllable.

Our framework has been evaluated through qualitative
and quantitative experiments, demonstrating its superiority in terms of the quality of the synthesis images, preservation of identity throughout the video, and correct spatial relationships between characters, as shown in Figure~\ref{fig:teaser}. We also conduct some extensive ablation experiments to verify the effectiveness of the modules in our framework. 

Generally, the key contributions of our method consist of the following:
\begin{itemize}
    \item We present MultiAnimate, a novel framework for multiple character animation that enables concurrent animation of multiple characters using their respective reference images and pose sequences within a shared environment, while maintaining appearance consistency and ensuring naturalistic character interactions.
    
    \item We develop a sophisticated animation framework comprising several innovative components, including an identity-specific reference net that facilitates multiple reference image processing, an identity-aware pose encoder that establishes robust character-pose bindings, and an interaction guider module that orchestrates coherent character interactions.
    
    \item We validate our approach through comprehensive experimental evaluations and ablation studies, demonstrating both the effectiveness of our framework and the essential nature of each architectural component.
\end{itemize}

\section{Related Work}
\label{sec:rw}

\subsection{Character Image Animation}

Distinguished from GAN-based methods~\cite{wang2022latent, Siarohin_2019_NeurIPS,Gao_2023_CVPR,chan2019everybodydance,zhang2022exploringdualtaskcorrelationpose,zhao2022thinplatesplinemotionmodel,siarohin2021motionrepresentationsarticulatedanimation,vougioukas2020realistic,pumarola2020ganimation}, the emergence of diffusion models, particularly the Stable Diffusion family~\cite{blattmann2023stablevideodiffusionscaling, rombach2022highresolutionimagesynthesislatent}, has catalyzed extensive research incorporating auxiliary modules to leverage their powerful generative capabilities for character image animation~\cite{chang2023magicdance,hu2023animateanyone,mimicmotion2024,ma2023follow,xu2023magicanimate,wang2023disco,zhu2024champ,karras2023dreamposefashionimagetovideosynthesis,yoon2025tpctesttimeprocrustescalibration,kim2024tcananimatinghumanimages,wang2024unianimate,tan2024animatexuniversalcharacterimage}. LEO~\cite{wang2025leo} models motion distributions in latent space to enable character animation from still images. LaMD~\cite{hu2025lamd} proposes a latent motion diffusion framework that factorizes video generation into motion synthesis and content reconstruction, leveraging a computationally efficient 2D UNet architecture. Animate Anyone~\cite{hu2023animateanyone} introduced a reference net architecture to capture and integrate reference features through self-attention mechanisms in the denoising net. DreamPose~\cite{karras2023dreamposefashionimagetovideosynthesis} implemented an adapter mechanism to incorporate reference image information into the original Stable Diffusion UNet. UniAnimate~\cite{wang2024unianimate} achieved temporal aggregation of reference images through the Mamba~\cite{gu2024mambalineartimesequencemodeling} architecture, while MimicMotion~\cite{mimicmotion2024} introduced a PoseNet with confidence-aware pose guidance to enhance motion realism. Further architectural innovations include Champ~\cite{zhu2024champ}, which incorporated the 3D parametric human model SMPL~\cite{loper2023smpl} as a control signal, integrating additional multi-modal signals such as depth and normal maps throughout the generation process. MIMO~\cite{men2024mimocontrollablecharactervideo} also proposed a 3D-aware approach, separately encoding and compositing human, background, and occlusion information.

Recent work on multiple character animation includes Xue et al.~\cite{xue2025multiplecharacterimageanimation}, which enables animation of multiple characters within a single reference image and pose sequence. Similarly, Wang et al.~\cite{wang2025multiidentityhumanimageanimation} developed a mask-aware identity encoder for multiple character animation. However, these approaches share a significant limitation: they require all characters to be present within a single reference image and their poses to be contained within a unified sequence, substantially constraining their practical applicability. In contrast, our methods proposed several modules to solve the above limitations, which allows multiple input of reference images and pose sequences.

\subsection{Multiple Subject Generation}

In the domain of image and video generation, the synthesis of content incorporating multiple subjects based on diverse reference images or concepts, commonly known as multi-concept personalization, represents a significant research direction. Several approaches have emerged to address this challenge. Some methods~\cite{kwon2024conceptweaverenablingmulticoncept, kumari2023multiconceptcustomizationtexttoimagediffusion, jang2024identitydecouplingmultisubjectpersonalization, Avrahami_2023} achieve multi-concept personalization through model fine-tuning with multiple text embeddings corresponding to distinct reference images. Training-free methodologies~\cite{xiao2023fastcomposertuningfreemultisubjectimage,wang2024moamixtureofattentionsubjectcontextdisentanglement, kim2024instantfamilymaskedattentionzeroshot, he2024uniportraitunifiedframeworkidentitypreserving,xiao2025fastcomposer,zhao2026freercustom} alternatively implement direct injection of multiple image embeddings during the inference phase. Mix-of-show~\cite{gu2023mixofshowdecentralizedlowrankadaptation} introduces an innovative approach utilizing multiple concept LoRAs and gradient fusion during the denoising process, enabling effective composition of customized concepts. Notably, Movie Weaver~\cite{liang2025movieweavertuningfreemulticoncept} presents a novel framework implementing anchored prompts and concept embeddings to establish direct associations between concepts and their corresponding reference images without architectural modifications, which has partially informed our framework.

\section{Method}
\label{sec:method}

\subsection{Preliminary}
\label{sec:preliminary}

%\paragraph{Diffusion Models.} 
AnimateDiff~\cite{guo2023animatediff} extends the conventional 2D UNet architecture~\cite{ronneberger2015unetconvolutionalnetworksbiomedical} of the traditional Latent Diffusion Model (LDM)~\cite{rombach2022highresolutionimagesynthesislatent} to a 3D UNet incorporating a temporal motion module, enabling video generation capabilities. VAE encoders~\cite{kingma2022autoencodingvariationalbayes,oord2018neuraldiscreterepresentationlearning} $\mathcal{E}$ are employed to transform video frames $x_{1:n}$ to latent representation as $z_{1:n}=\mathcal{E}(x_{1:n})$. Subsequently, gaussian noise $\epsilon \sim \mathcal{N}(0,1)$ will be added to $z_{1:n}$ according to the time step $t$, and the training objective of the framework could be formulated as:

\begin{equation}
    \mathcal{L}=\mathbb{E}_{z_{1:n},c,t}(\Vert \epsilon-\epsilon_\theta(z^t_{1:n},c,t) \Vert_2^2)
\end{equation}
where $\epsilon_\theta$ represents the denoising function of the denoising UNet, $c$ represents the conditional inputs, and $z^t_{1:n}$ represents the intermediate latent feature during denoising at timestep $t$.

%\paragraph{Problem Formulation.}

\begin{figure*}
  \centering
  \includegraphics[width=\textwidth]{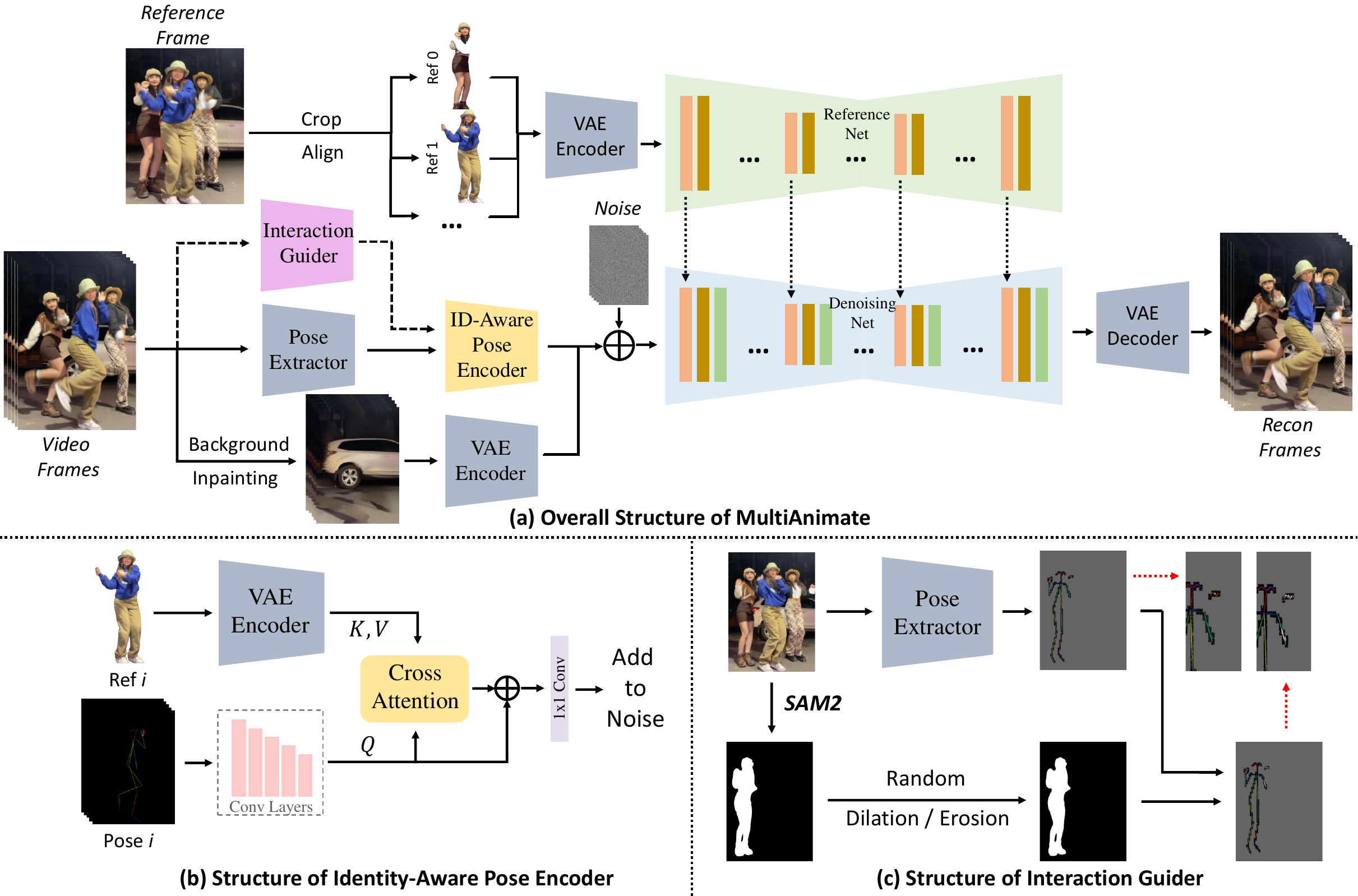}
  \caption{(a) The framework of MultiAnimate, where the Reference Net indicates our identity-specific reference net (detailed in following sections). The Denoising Net receives composite inputs comprising background features and identity-aware pose features. (b) The structure of identity-aware pose encoder, utilizing cross-attention mechanism for identity-pose binding. (c) The structure of interaction guider. Background color of 2D skeleton image modified for visualization clarity, with inset showing detailed interaction guider effects.}
  \label{fig:pipeline}
\end{figure*}

\subsection{Overview}
\label{sec:overview}

Conventional character image animation frameworks are limited to processing single reference images with corresponding control signals. Even recent approaches that support multiple character animation impose significant limitations: they require all subjects to be presented within a single reference image while simultaneously demanding comprehensive pose sequences containing poses of each character, substantially limiting their practical applications. This limitation motivates our fundamental research question: \textit{How can we enable independent reference images and pose sequences for multiple characters within a unified animation framework?} To address this challenge systematically, we propose a comprehensive problem formulation.

Given a set of reference character images $\{I^k\}_{k=1}^N$ and their corresponding pose sequences $\{P_{1:n}^k\}_{k=1}^N$, where each pose sequence $P_{1:n}^j\in\mathbb{R}^{n\times H\times W\times 3}$ comprises $n$ frames of 2D human skeleton representations serving as the control signals for animating the respective reference image $I^j\in\mathbb{R}^{H\times W\times 3}$. Our objective is to simultaneously animate these characters from distinct reference images within a shared scenario input ${B_{1:n}}\in\mathbb{R}^{n\times H\times W\times 3}$, synthesizing a video sequence $x_{1:n}\in\mathbb{R}^{n\times H\times W\times 3}$. Additionally, we define optional interaction mask sequences $\{M_{1:n}^k\}_{k=1}^N$, where $M_{1:n}^j\in\mathbb{R}^{n\times H\times W}$ denotes the spatial occupancy region of character $j$ in the animation result. The comprehensive conditional generation process $g(\cdot)$ can be formulated as:
\begin{equation}
    g(\cdot):(\{I^k\}_{k=1}^N, \{P_{1:n}^k\}_{k=1}^N, B_{1:n}, [\{M_{1:n}^k\}_{k=1}^N])\rightarrow x_{1:n}
\end{equation}

To achieve this, our framework incorporates fundamental modifications to both the reference net architecture and the integration of the pose image. Specifically, we introduce an identity-specific reference net that enables accurate feature extraction from multiple reference images. Additionally, we implement an identity-aware pose encoder that establishes explicit identity-pose bindings after feature extraction, thereby preventing identity ambiguity during the generation process. To facilitate naturalistic inter-character interactions, we also implement an interaction guider module as an optional component that orchestrates proper occlusion relationships and spatial interactions between characters during the animation process. Furthermore, inspired by MIMO~\cite{men2024mimocontrollablecharactervideo}, we integrate background information into the network architecture to enhance the framework's practical utility. The overall pipeline of our framework is illustrated in Figure~\ref{fig:pipeline}(a).

\subsection{Identity-Specific Reference Net}
\label{sec:penet}
Since its introduction in Animate Anyone~\cite{hu2023animateanyone}, the reference net architecture has gained widespread adoption in character image animation tasks, owing to its robust feature extraction capabilities from reference images. However, when applied to multiple character animation, traditional reference net imposes a significant constraint: it requires all characters to be present within a single reference image. This limitation substantially restricts both reference image selection and the practical applicability of the framework. When the number of characters to be animated increases, it is difficult to find a suitable reference image. To address this limitation, we propose an enhanced approach to reference net utilization, wherein character features are extracted independently from \textbf{separate reference images} and subsequently integrated during the denoising process. This design will enable a freer choice of reference images and enhance the scalability of the framework.

\begin{figure}
  \centering
  \includegraphics[width=\linewidth]{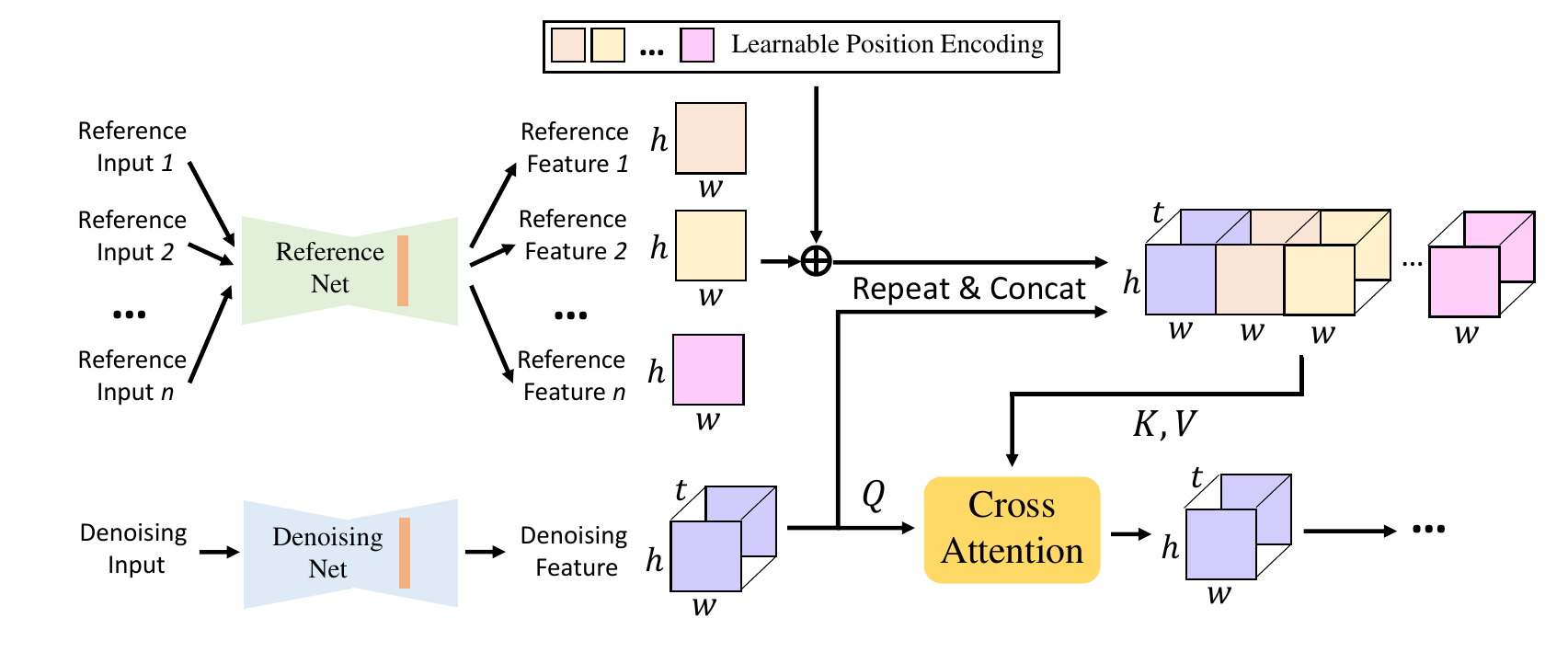}
  %\vspace{-2em}
  \caption{The detailed structure of our identity-specific reference net. Learnable position encoding are added to each reference features before they are replicated and concatenated with the denoising feature.}
  \label{fig:positionencoding}
\end{figure}

The conventional reference net architecture replicates and concatenates reference features with denoising features, a naive extension is directly concatenating features from multiple reference images. However, since the inherent spatial-agnostic nature of the reference net, a significant challenge arises: during the cross-attention operations, the denoising feature queries the entire concatenated reference feature without spatial discrimination. This lack of spatial awareness can lead to feature attribution ambiguities, as the denoising net cannot effectively determine which portions of the reference features should be utilized, thus leading to feature attribution errors, such as incorrect clothing assignments. To address this challenge, we propose an identity-specific reference net, as illustrated in Figure~\ref{fig:positionencoding}, which \textbf{introduces spatial awareness into the feature processing pipeline}. The whole process can be formulated as:
\begin{equation}
    f^{enc}_{i,l} = f_{i,l} +PE_{i,l}
\end{equation}
where $f_{i,l}$ represents the feature representation $i$ after the $l$-th layer of the reference net, and $PE_{i,l}$ denotes the corresponding position encoding, which is learned during the training process. Each encoded feature $f^{enc}_{i,l}\in\mathbb{R}^{h\times w\times c}$ is subsequently replicated along temporal dimension $t$, and then concatenated with the denoising feature of the $l$-th layer $f^{denoise}_{l}\in\mathbb{R}^{t\times h\times w\times c}$ along width dimension $w$, yielding a composite feature representation $f_l\in\mathbb{R}^{t\times h\times (n+1)w\times c}$, detailed visualization is presented in Figure~\ref{fig:positionencoding}. The denoising feature then queries the composite feature through a cross-attention mechanism to extract information from distinct reference features. Specifically,
\begin{equation}
\begin{gathered}
    f^{out}_{l} = softmax(\frac{QK^T}{\sqrt{d}})V \\
    Q=q(f^{denoise}_{l}),\quad K=k(f_l),\quad V=v(f_l)
\end{gathered}
\end{equation}
the resulting feature with size $\mathbb{R}^{t\times h\times w\times c}$ serves as input to the subsequent module, following the approach proposed in Animate Anyone~\cite{hu2023animateanyone}.

\subsection{Identity-Aware Pose Encoder}
\label{sec:idaware}

In character image animation, numerous approaches~\cite{hu2023animateanyone,mimicmotion2024,wang2024unianimate,men2024mimocontrollablecharactervideo} commonly employ convolutional networks to extract features from pose sequences, subsequently incorporating these features with Gaussian noise to initialize the denoising process. However, in our multi-character animation framework, where each reference image corresponds to a distinct pose sequence, the conventional approach of naively aggregating pose sequence features proves problematic. This direct feature aggregation often leads to identity ambiguity, as the individual pose sequences lack inherent distinguishing characteristics and become indiscernible when combined.

To establish distinguishable pose sequence features, we implement a cross-attention mechanism that \textbf{binds identity information to the extracted features}. Specifically, as shown in Figure~\ref{fig:pipeline}(b), we encode $j$-th reference images $I^j$ through VAE encoder $\mathcal{E}$ to obtain reference image feature $z^j = \mathcal{E}(I^j)$, where $z^j\in\mathbb{R}^{\frac{H}{8}\times\frac{W}{8}\times4}$ and extract its pose feature $z^j_{pose,1:n}=\mathcal{E}_{pose}(P^j_{1:n})$ from the corresponding pose sequences $P^j_{1:n}$ with convolutional network $\mathcal{E}_{pose}$. Notably, the pose feature of each frame shares the same size with the reference image feature, we consequently introduce a cross-attention layer for identity binding which could formulated as:
\begin{equation}
\begin{gathered}
    f^j_{pose,k} = Conv(z^j_{pose,k}+softmax(\frac{QK^T}{\sqrt{d}})V) \\
    Q=q(z^j_{pose,k}),\quad K=k(z^j),\quad V=v(z^j)
\end{gathered}
\end{equation}
where $f^j_{pose,k}$ denotes the identity-aware pose feature for the $j$-th pose sequence of the $k$-th frame, $Conv$ denotes a $1\times1$ convolution layer. The query is calculated from the pose feature $z^j_{pose,k}$ and the key and value is calculated from the reference image feature $z^j$.

\subsection{Interaction Guider}
\label{sec:maskguider}

While our framework successfully enables multiple character animation through the integration of identity-specific reference net and identity-aware pose encoder, the processing of independent pose sequences within a shared environment presents significant challenges, particularly regarding spatial conflicts and inter-character occlusions. Although 2D skeleton pose representations can be extracted using off-the-shelf networks such as DWPose~\cite{yang2023effective}, DensePose~\cite{Guler2018DensePose}, and OpenPose~\cite{cao2019openposerealtimemultiperson2d}, direct application of these networks fails to preserve crucial occlusion relationships in multi-character scenarios. To address this limitation, we implement an interaction guider module that leverages character-specific mask information produced by segmentation networks (i.e., SAM2~\cite{ravi2024sam2}) to process and refine the extracted 2D skeleton representations, thereby ensuring proper spatial relationships between characters. 

We extend traditional 2D skeleton representation by introducing three distinct states: (i) visible skeletons with high prediction confidence, (ii) masked skeletons with high prediction confidence, and (iii) skeletons with low prediction confidence. The rendering protocol follows corresponding rules: visible skeletons are drawn conventionally, masked skeletons are rendered in white, and low-confidence skeletons are omitted. As shown in Figure~\ref{fig:pipeline}(c), we implement data augmentation through random mask modification and process skeleton representations accordingly, the occluded part of the 2D skeleton representation will be rendered in white. The interaction guider serves as an optional component that enhances the framework's capability to handle complex inter-character interactions. % Specifically, the process of the mask guider can be formulated as follow:

\subsection{Long Video Inference}

To process extended video sequences, we implement a sliding window approach by segmenting the input video into overlapping slices. Each slice consists of 14 frames with a 4 frame overlap between consecutive segments. The model inference is performed on individual slices, and the results in overlapping regions are averaged with their corresponding regions in adjacent slices to ensure temporal consistency. The algorithmic procedure is formalized in the following pseudo code:

\begin{algorithm}
\caption{Sliding window latent fusion for long video generation.}\label{alg:sliding_fusion}
\begin{algorithmic}[1]
\Require $z_i^j$: latent feature of the $j$-th frame in the $i$-th video segment; 
         $N_1$: number of frames in a video segment; 
         $N_2$: number of overlapped frames; 
         $C$: conditions in the generation process; 
         $T$: total number of diffusion steps.
\Ensure $z'$: a long sequence of latent features of video frames.
\State Sample $z \sim \mathcal{N}(0, 1)$
\For{$t = T$ down to $1$}
    \For{each segment index $i = 1, 2, \dots$}
        \State $z_i \gets G(z_i, C, t)$
    \EndFor
    \For{each segment index $i = 1, 2, \dots$}
        \For{$j = 1$ to $N_1$}
            \If{$i > 1$ \textbf{and} $j \leq N_2$}
                \State $z_i^j \gets \frac{1}{2}\bigl(z_i^j + z_{i-1}^{\,N_1 - N_2 + j}\bigr)$
                \Comment{Latent fusion with the previous sliding window.}
            \ElsIf{$j > N_1 - N_2$}
                \State $z_i^j \gets \frac{1}{2}\bigl(z_i^j + z_{i+1}^{\,N_2 - N_1 + j}\bigr)$
                \Comment{Latent fusion with the next sliding window.}
            \EndIf
        \EndFor
    \EndFor
\EndFor
\State Concatenate all segments $\{z_i^j\}$ to obtain $z'$
\end{algorithmic}
\end{algorithm}

\section{Experiment}
\label{sec:exp}

\subsection{Dataset Construction}

Our data collection process began with sourcing videos from public online platforms that feature multi-character interactions. Following data collection, we implement a filtering protocol to mitigate the impact of noise on the training process. Initially, we employ YOLO-X~\cite{yolox2021} for character detection and bounding box extraction from the video sequences in the dataset. Characters whose bounding box dimensions constitute less than 10\% of the frame size are excluded from subsequent processing procedures, including segmentation and pose extraction, to maintain data quality and computational efficiency.

After detecting the bounding boxes of the characters, we utilize the extracted bounding boxes as prompts for SAM2~\cite{ravi2024sam2} to generate segmentation masks for all eligible characters within the video sequences. To maintain consistent character identification throughout sequences containing complex inter-character interactions, particularly position interchanges, we implement a position-based labeling protocol. This protocol establishes character indices based on their initial horizontal spatial distribution, where numerical identifiers (0,1,2,...) are assigned according to the sorted coordinates of bounding box centers in the first frame.

\begin{figure}[h]
  \centering
  \includegraphics[width=5cm]{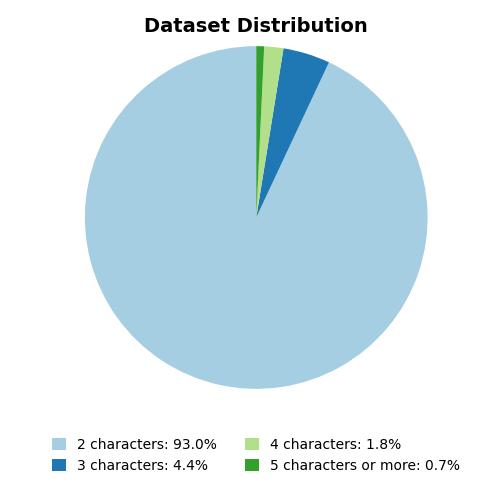}
  \caption{The distribution of the percentage of the number of characters in our dataset.}
  \label{fig:dis}
\end{figure}

Furthermore, we implement an occlusion-based filtering criterion to exclude videos containing extensively occluded frames, as such occlusions can significantly compromise the quality of reference feature extraction and subsequent reconstruction. The filtering process evaluates the occlusion metrics for each character throughout the video sequence. Specifically, we compute the frame-wise occlusion ratio between each character's bounding box and those of other characters. Videos are excluded from the dataset if any character's bounding box exhibits occlusion rates exceeding 65\% in more than half of the total frames, as such severe occlusions would impair the extraction of reliable reference features. Figure~\ref{fig:dis} illustrates the distribution of the number of characters per video in our dataset. The dataset comprises over 10,000 videos featuring multi-character interactions across diverse action categories. From this collection, we randomly select a 5\% subset to serve as the test set.

% \subsubsection{Dataset Distribution.} 

\subsection{Experimental Setups}

\subsubsection{Implementation Details.} Our training implements a self-supervised strategy with careful reference image selection criteria: characters are excluded if they occupy minimal image space or experience substantial occlusion throughout the video sequence. We employ state-of-the-art SAM2~\cite{ravi2024sam2} for character-specific mask segmentation and DWPose~\cite{yang2023effective} for extracting corresponding 2D skeleton representations of eligible characters.

Our model training procedure consists of two stages on 8 NVIDIA A100 GPUs. The initial stage excludes the motion module and utilizes a batch size of 16. In the subsequent stage, we freeze all previously trained parameters and exclusively train the motion module using video sequences of 14 frames. During training, we implement center-cropping guided by pose sequences to optimize character centralization within each frame.

\subsubsection{Evaluation Setups.} We perform the following evaluations on our multi-character benchmark. For all compared methods, we use the official pre-trained models. We note that comparisons with recent multiple character animation methods~\cite{xue2025multiplecharacterimageanimation,
wang2025multiidentityhumanimageanimation} are not feasible due to their unavailability as open-source implementations. Given that the compared methods do not support multiple character animation inherently, we adapted their pipelines through a standardized extension: utilizing reference image that contains all target characters and employing the original methods' pose extraction networks (e.g., DWPose~\cite{yang2023effective} or DensePose~\cite{Guler2018DensePose}) to generate consolidated pose representations. These pose representations encompass all character poses within a single image for each frame of the video sequence. Moreover, to facilitate fair comparison with baseline methods, we select those images that contain all the characters to be animated as reference images.

\subsection{Qualitative Comparison}
\label{sec:quali}

\begin{figure*}[tb]
    \begin{center}
    \setlength{\tabcolsep}{0.5pt}
    \begin{tabular}{m{1.8cm}<{\centering}m{1.8cm}<{\centering}m{1.8cm}<{\centering}m{1.8cm}<{\centering}m{1.8cm}<{\centering}m{1.8cm}<{\centering}m{1.8cm}<{\centering}}

    \fontsize{5pt}{6.5pt}\selectfont{MimicMotion\cite{mimicmotion2024}} & \fontsize{5pt}{6.5pt}\selectfont{Moore-Animate\cite{hu2023animateanyone}} & \fontsize{5pt}{6.5pt}\selectfont{UniAnimate\cite{wang2024unianimate}} & \fontsize{5pt}{6.5pt}\selectfont{MagicAnimate\cite{xu2023magicanimate}} &\fontsize{5pt}{6.5pt}\selectfont{MagicPose\cite{chang2023magicdance}} & \fontsize{5pt}{6.5pt}\selectfont{Ours} & \fontsize{5pt}{6.5pt}\selectfont{Ground Truth}\\
    \includegraphics[width=1.75cm]{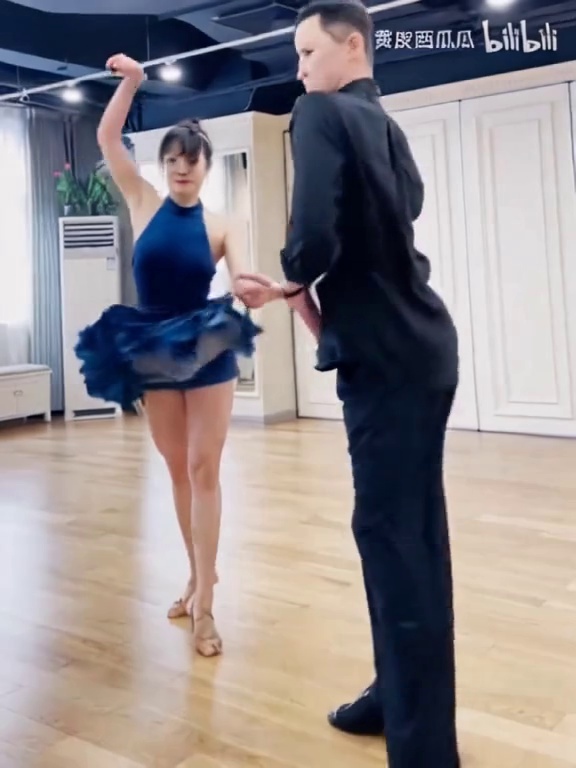}
    &\includegraphics[width=1.75cm]{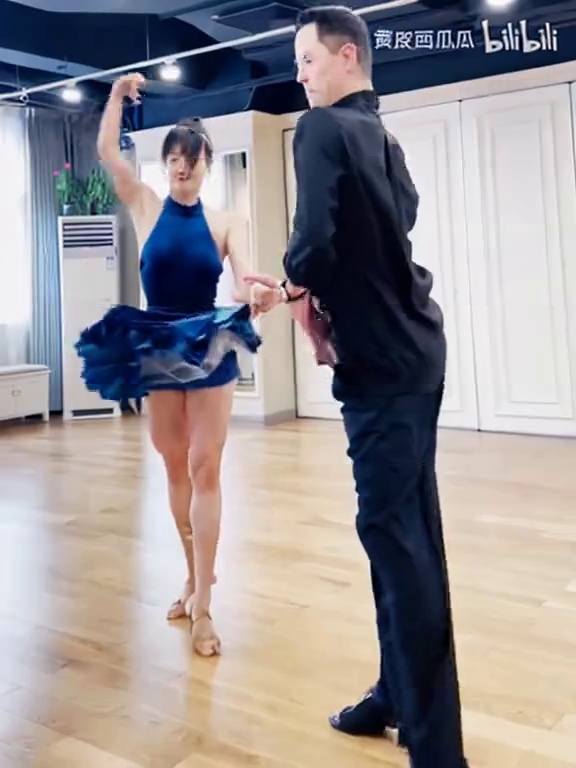}
    &\includegraphics[width=1.75cm]{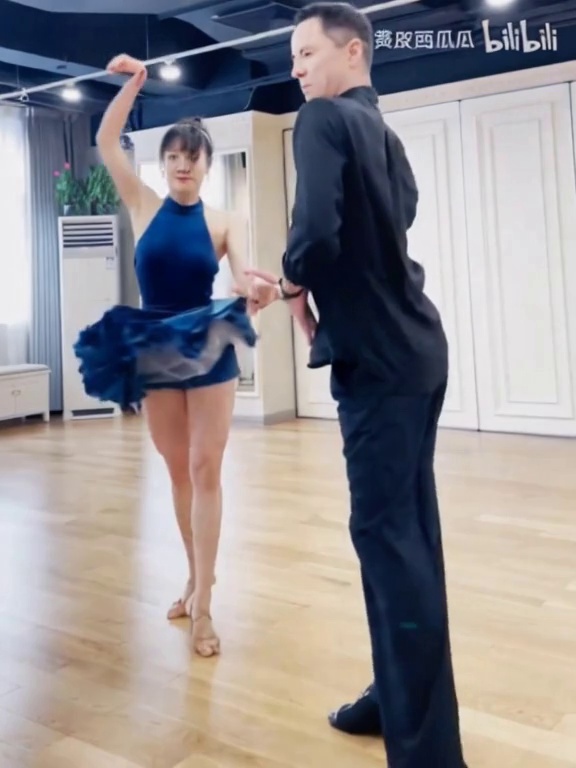}
    &\includegraphics[width=1.75cm]{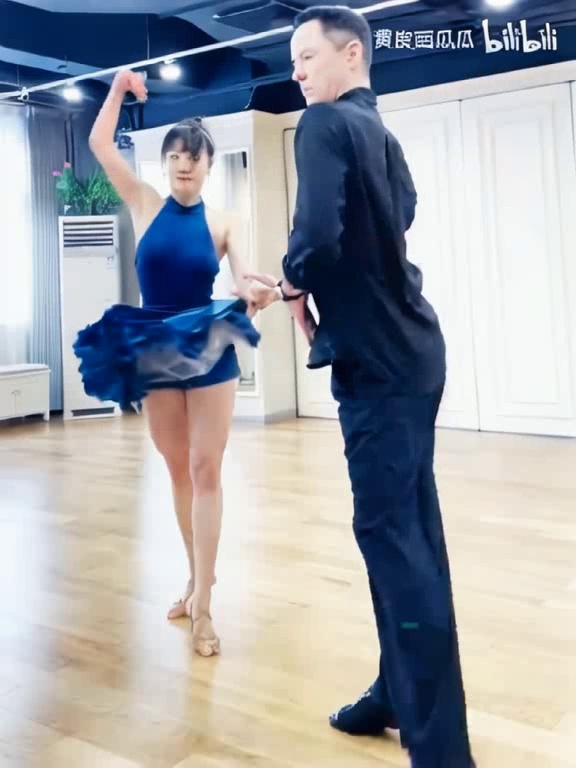}
    &\includegraphics[width=1.75cm]{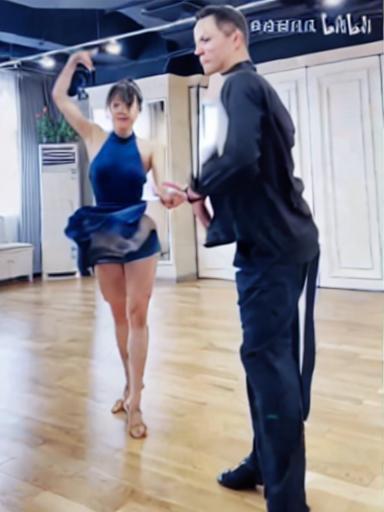}
    &\includegraphics[width=1.75cm]{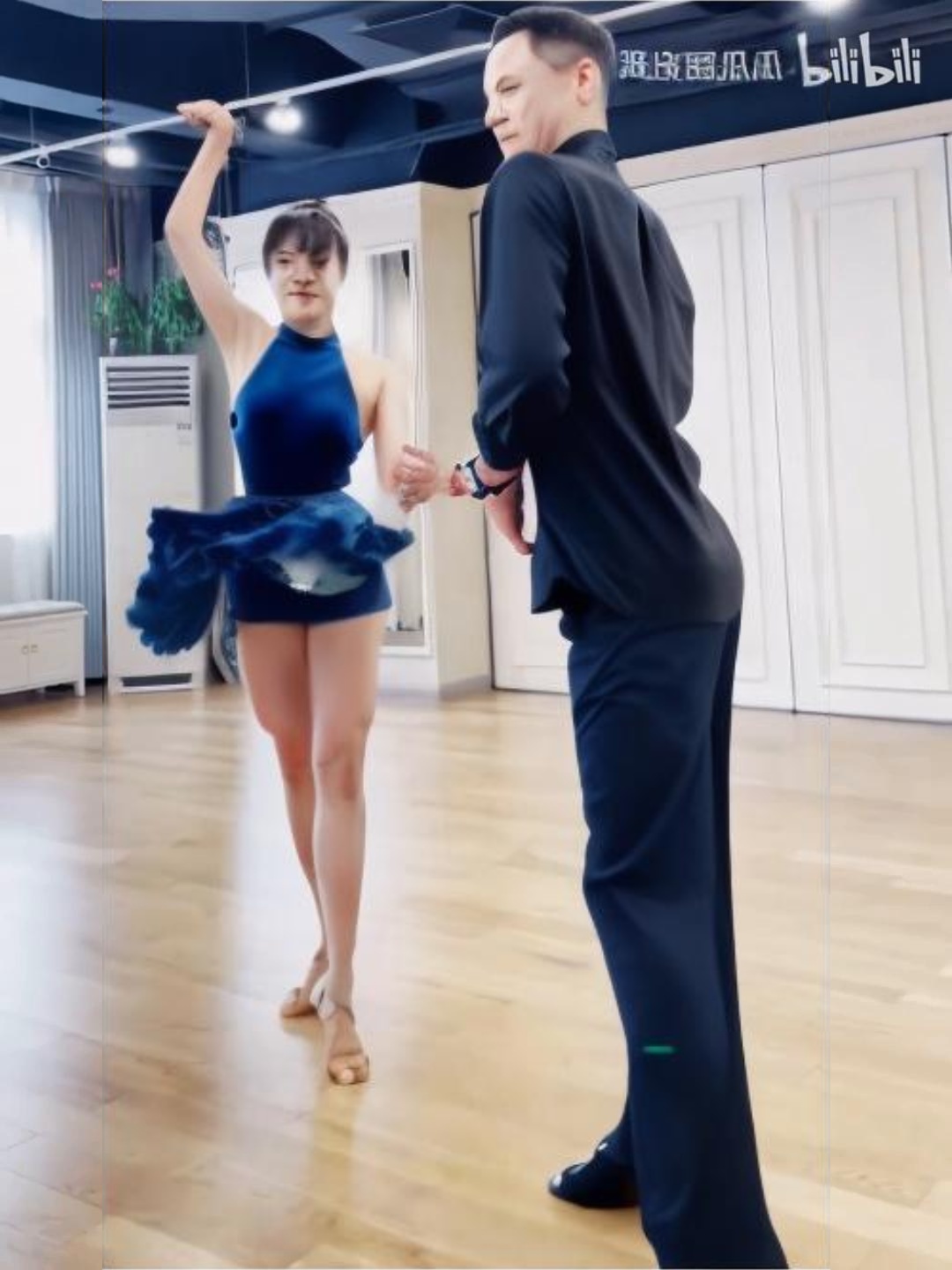}
    &\includegraphics[width=1.75cm]{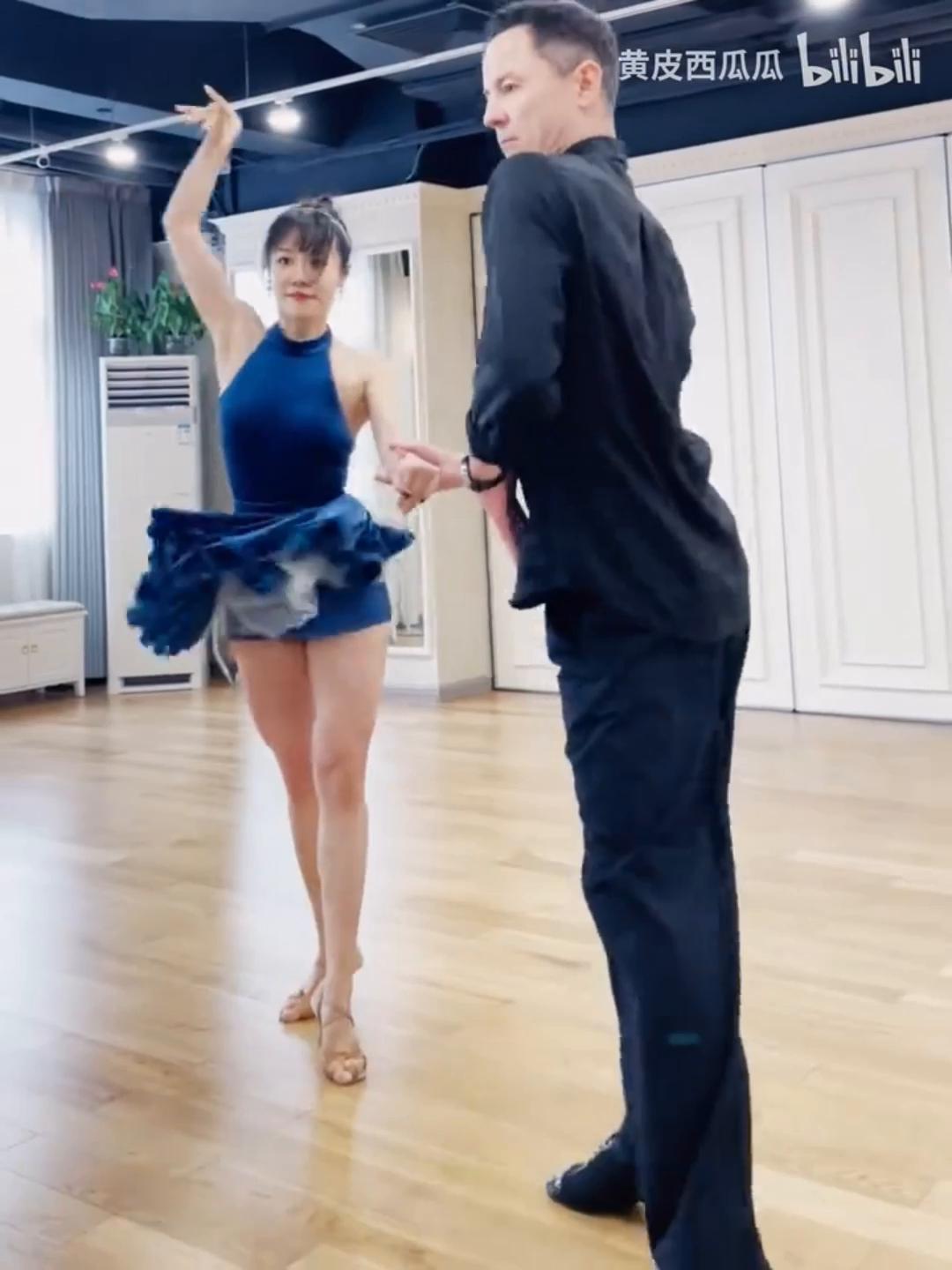} \\

    \includegraphics[width=1.75cm]{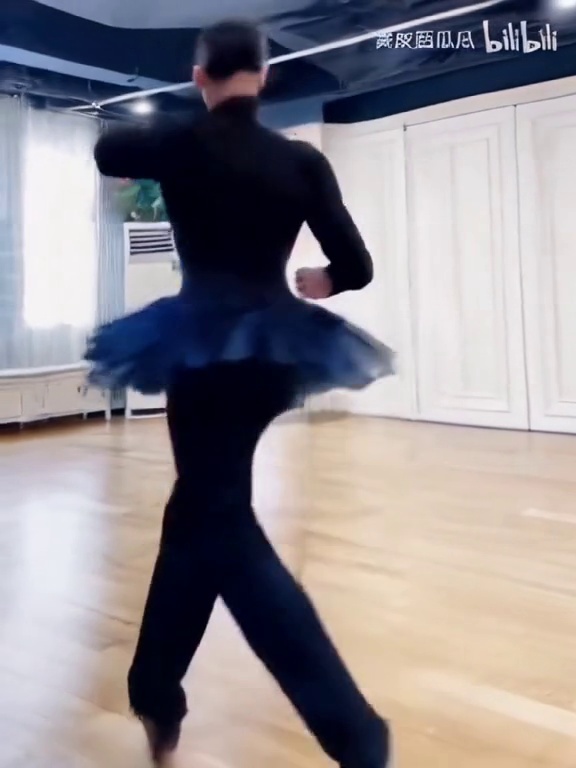}
    &\includegraphics[width=1.75cm]{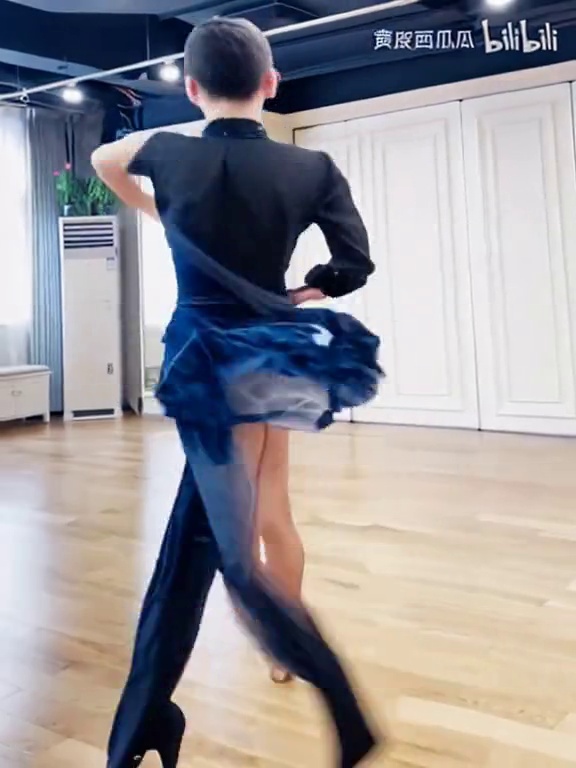}
    &\includegraphics[width=1.75cm]{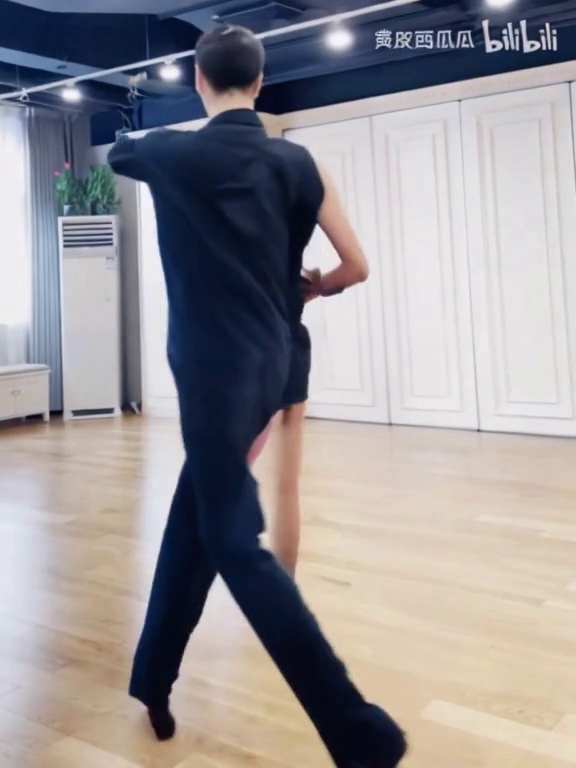}
    &\includegraphics[width=1.75cm]{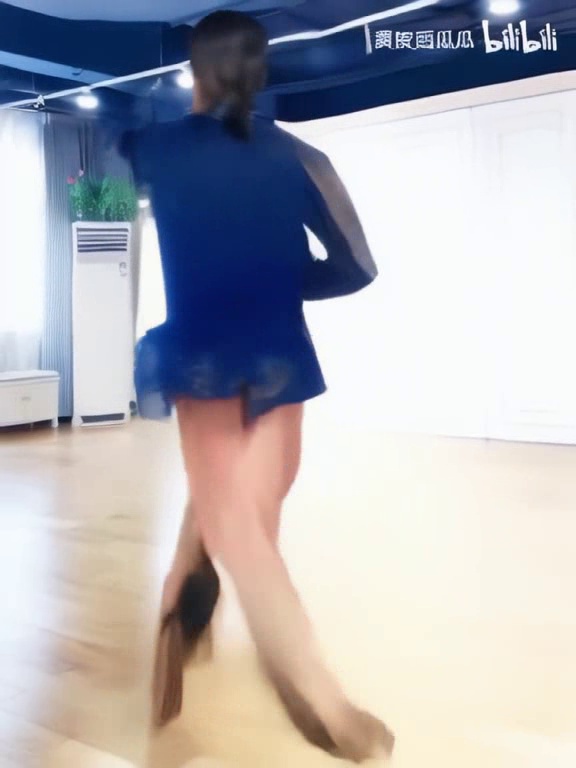}
    &\includegraphics[width=1.75cm]{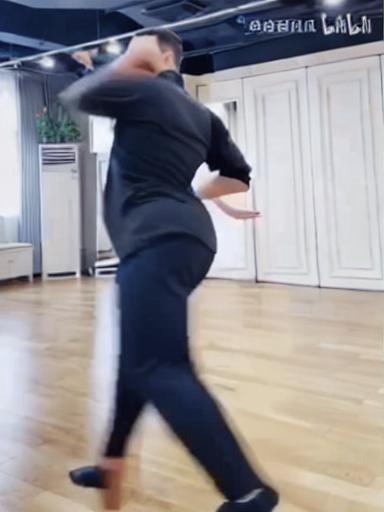}
    &\includegraphics[width=1.75cm]{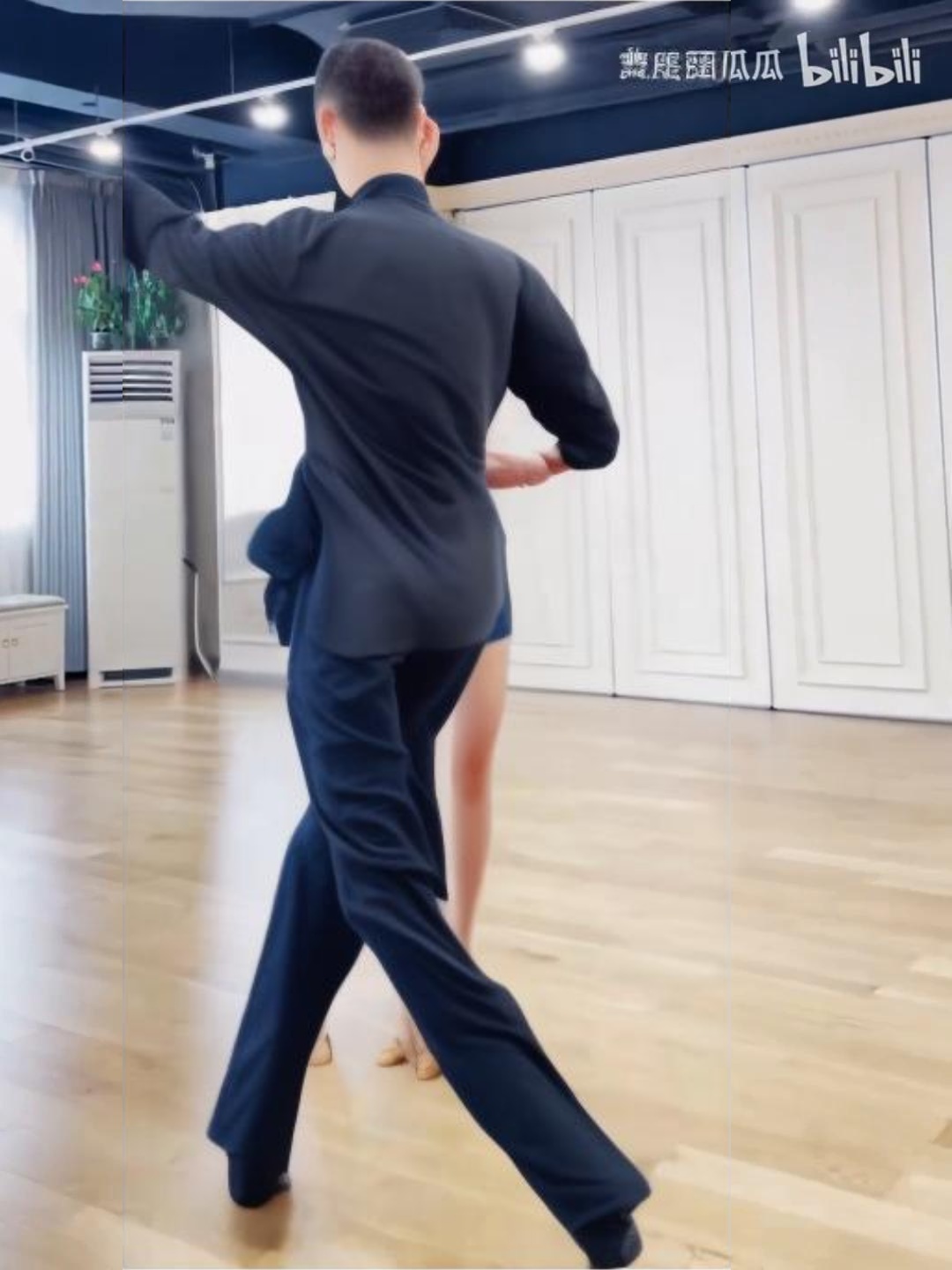}
    &\includegraphics[width=1.75cm]{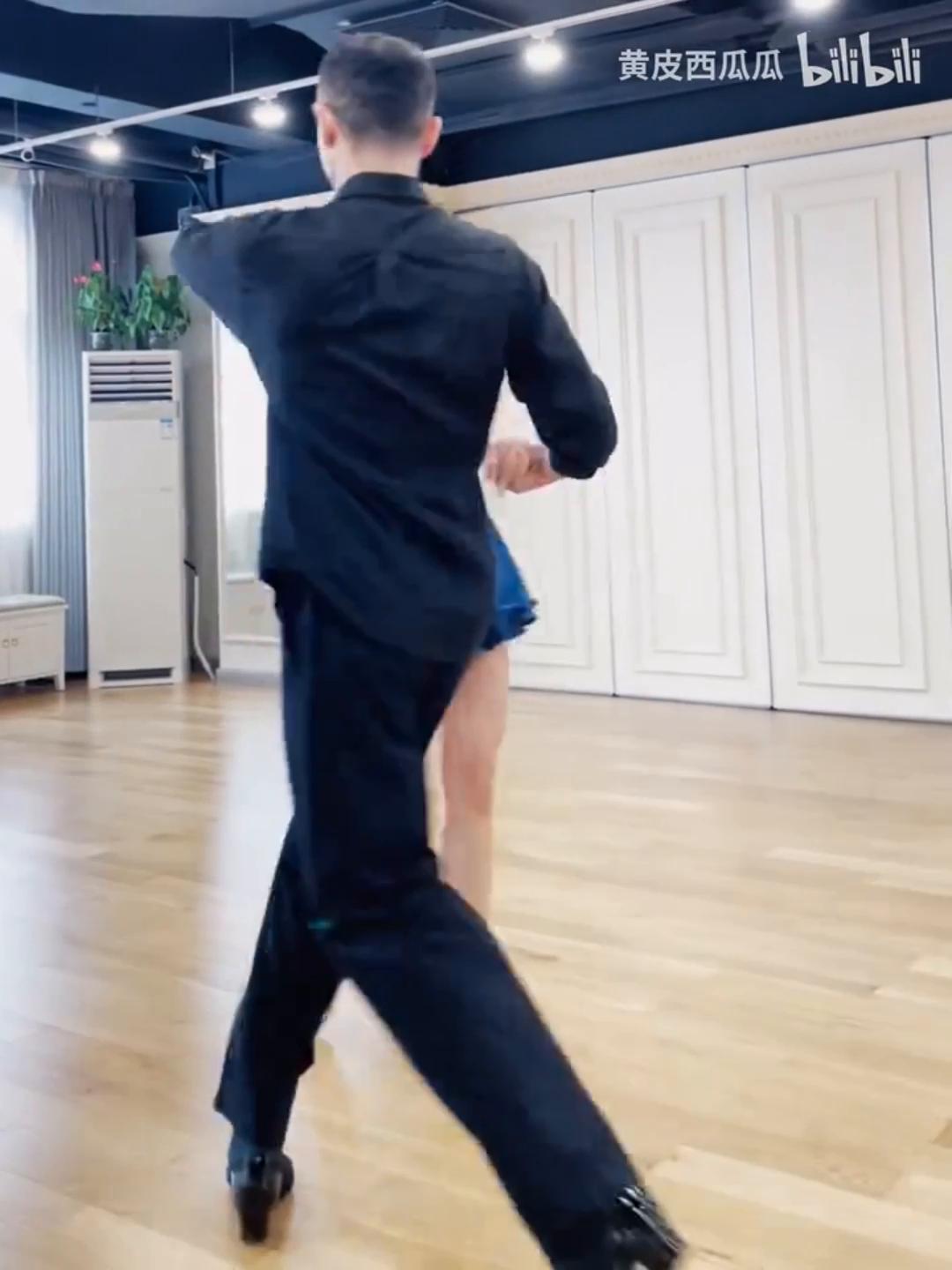} \\

    \includegraphics[width=1.75cm]{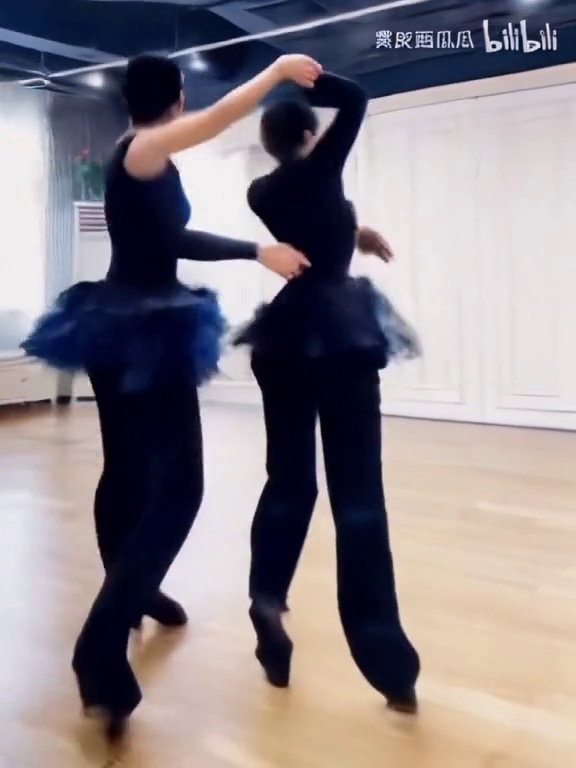}
    &\includegraphics[width=1.75cm]{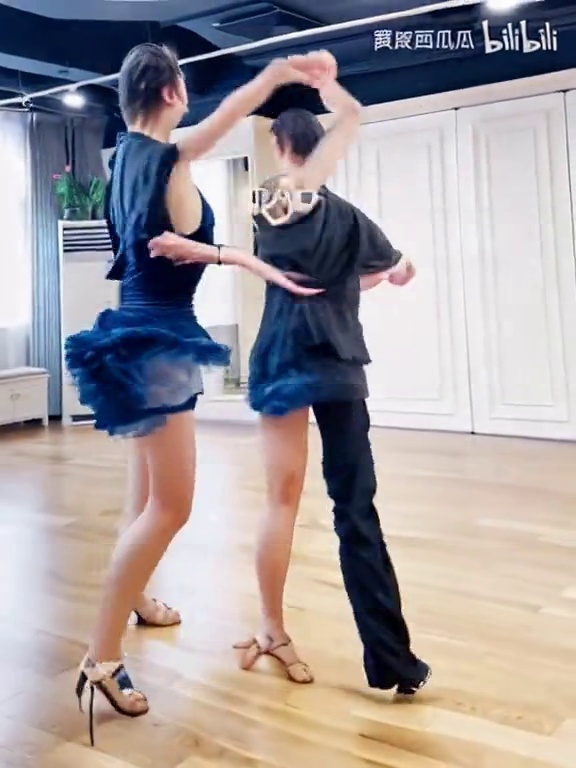}
    &\includegraphics[width=1.75cm]{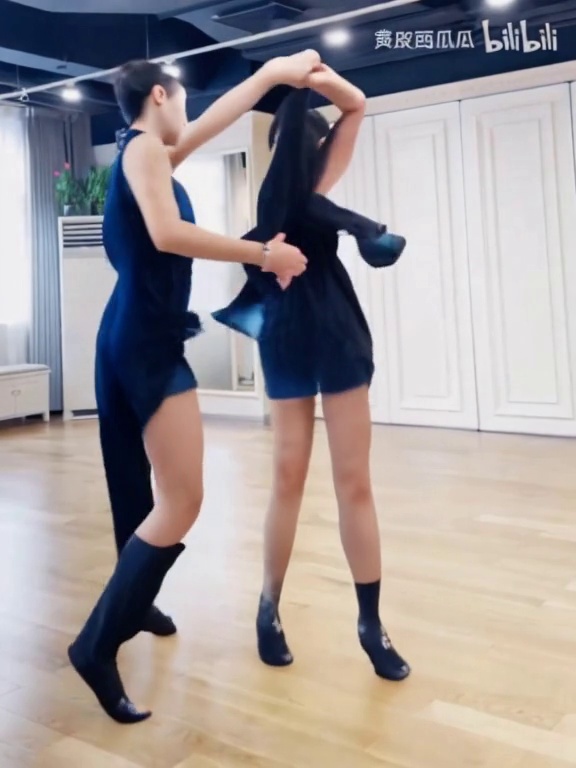}
    &\includegraphics[width=1.75cm]{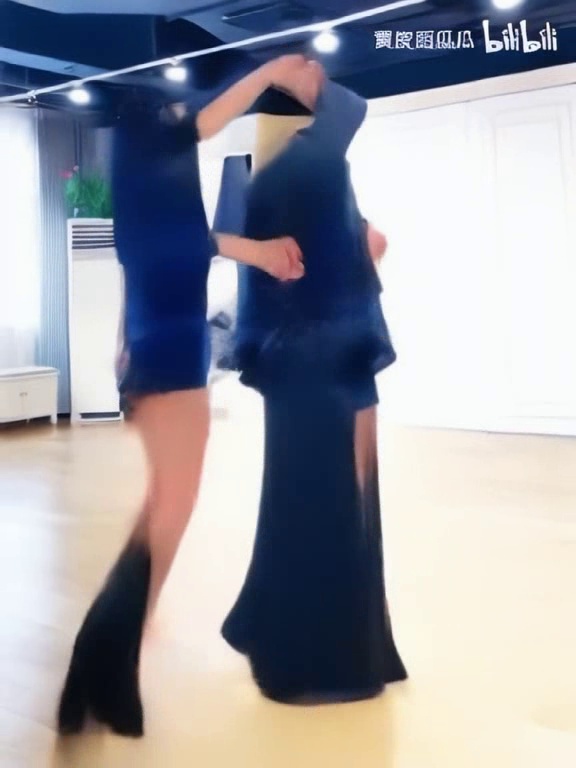}
    &\includegraphics[width=1.75cm]{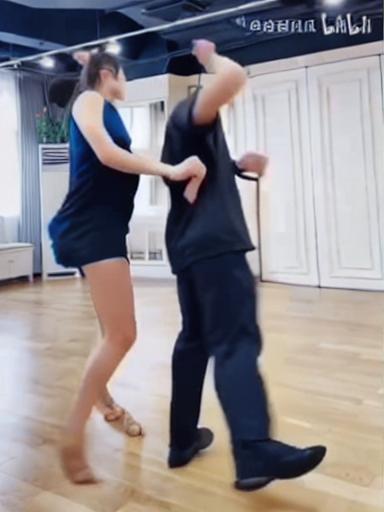}
    &\includegraphics[width=1.75cm]{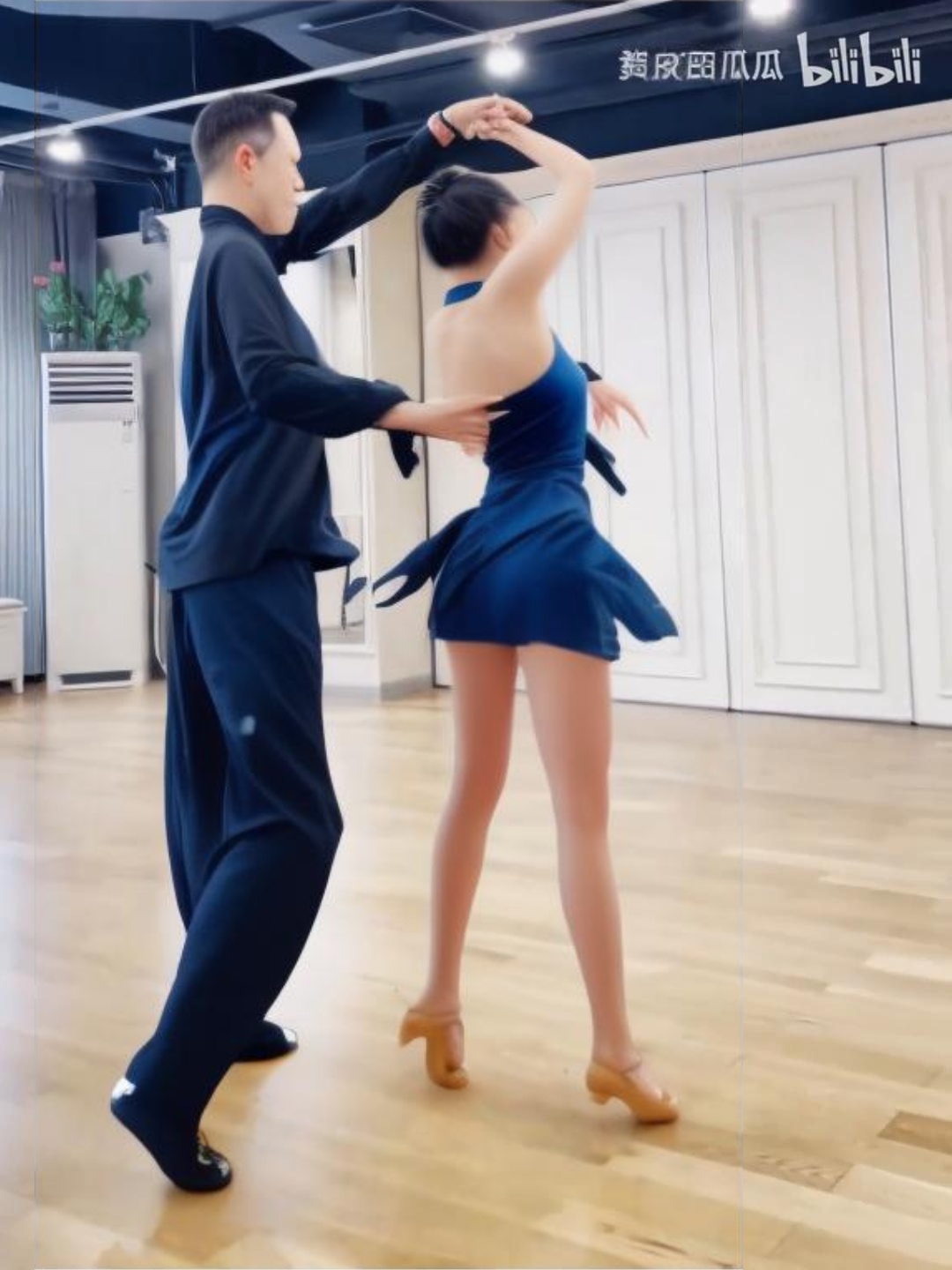}
    &\includegraphics[width=1.75cm]{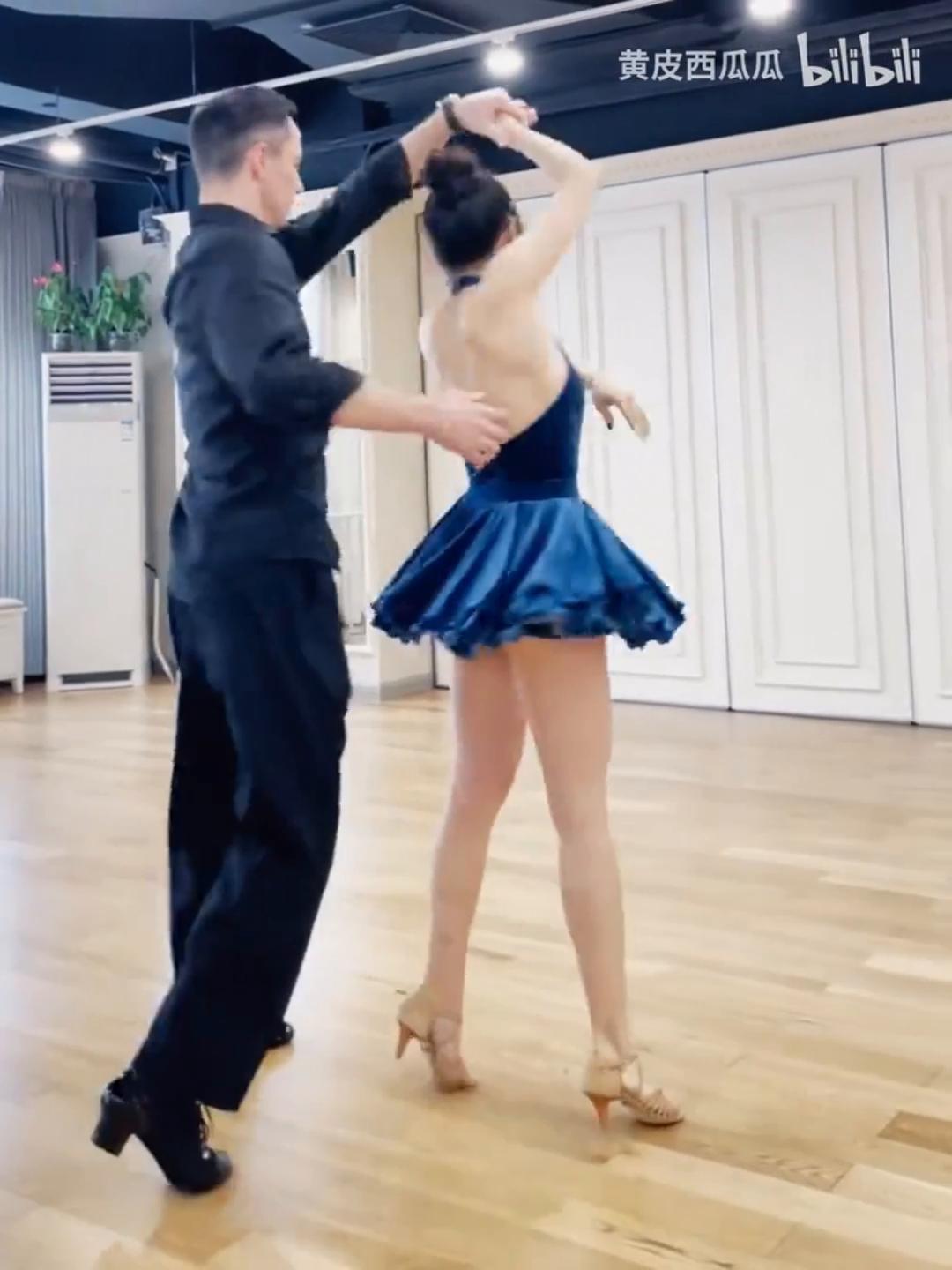} \\

    \includegraphics[width=1.75cm]{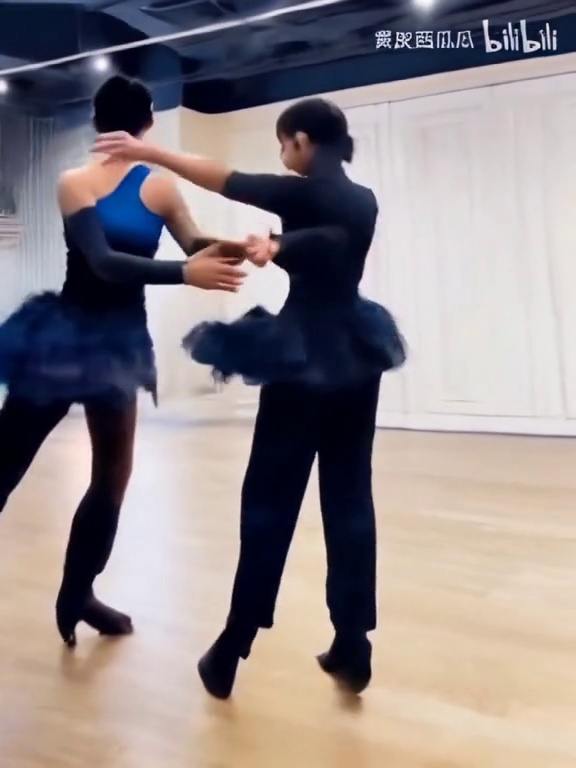}
    &\includegraphics[width=1.75cm]{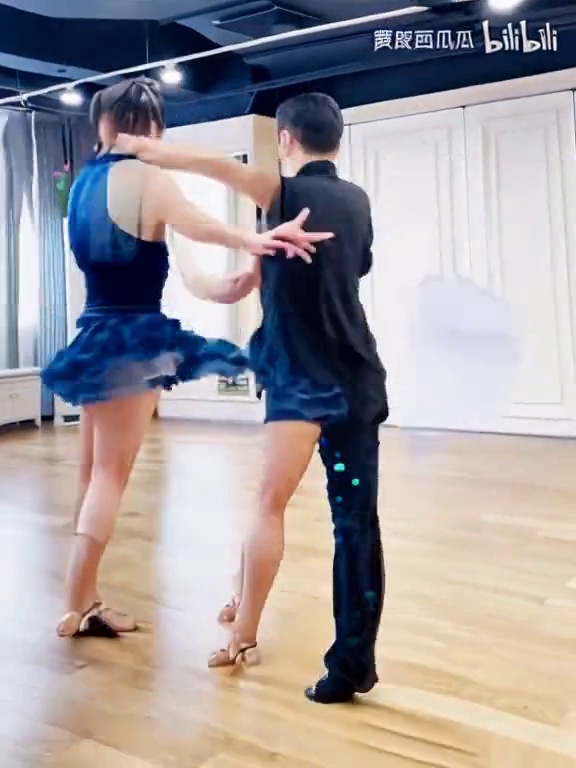}
    &\includegraphics[width=1.75cm]{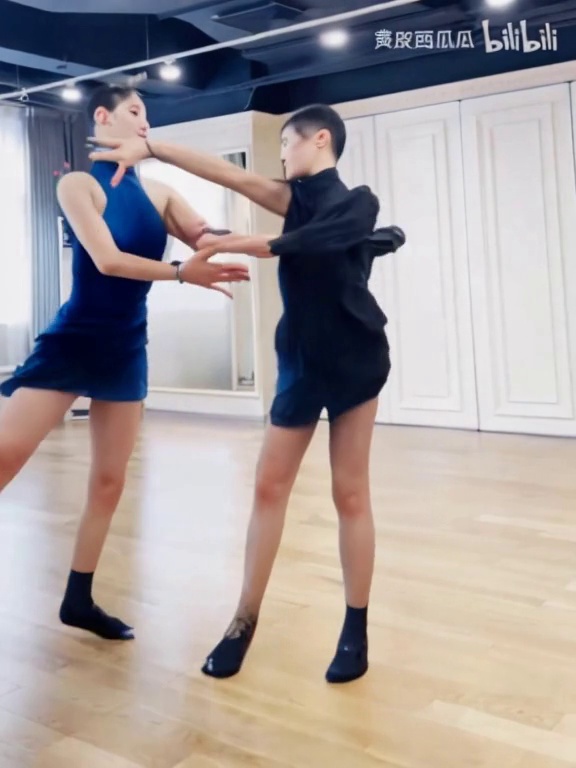}
    &\includegraphics[width=1.75cm]{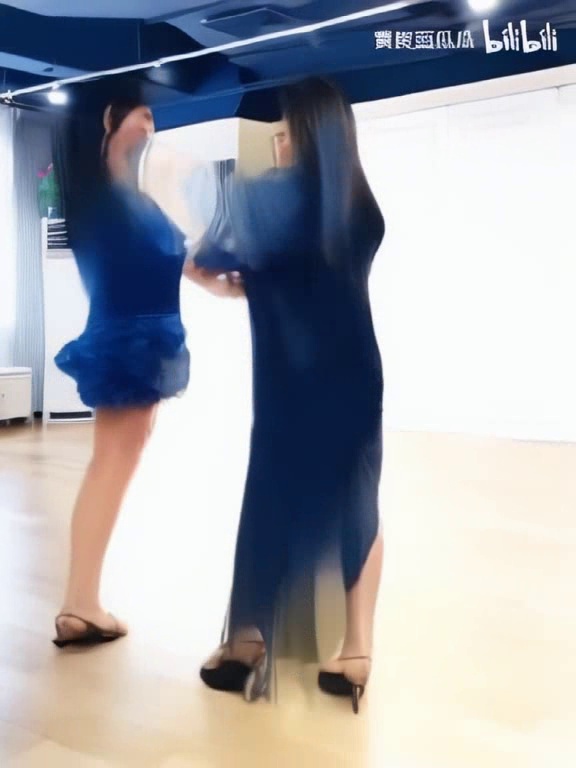}
    &\includegraphics[width=1.75cm]{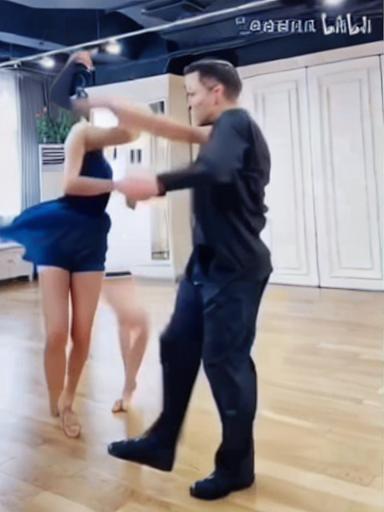}
    &\includegraphics[width=1.75cm]{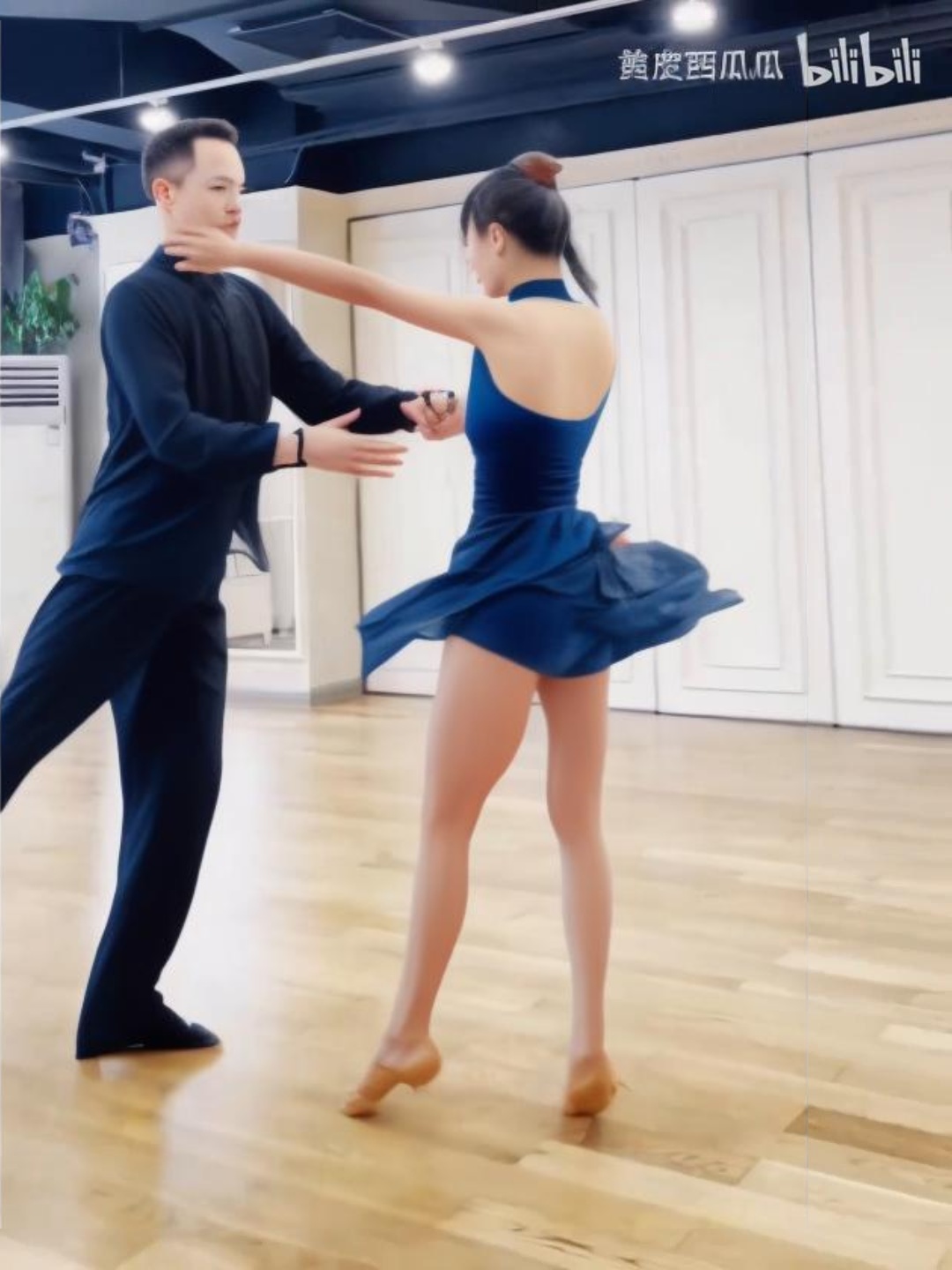}
    &\includegraphics[width=1.75cm]{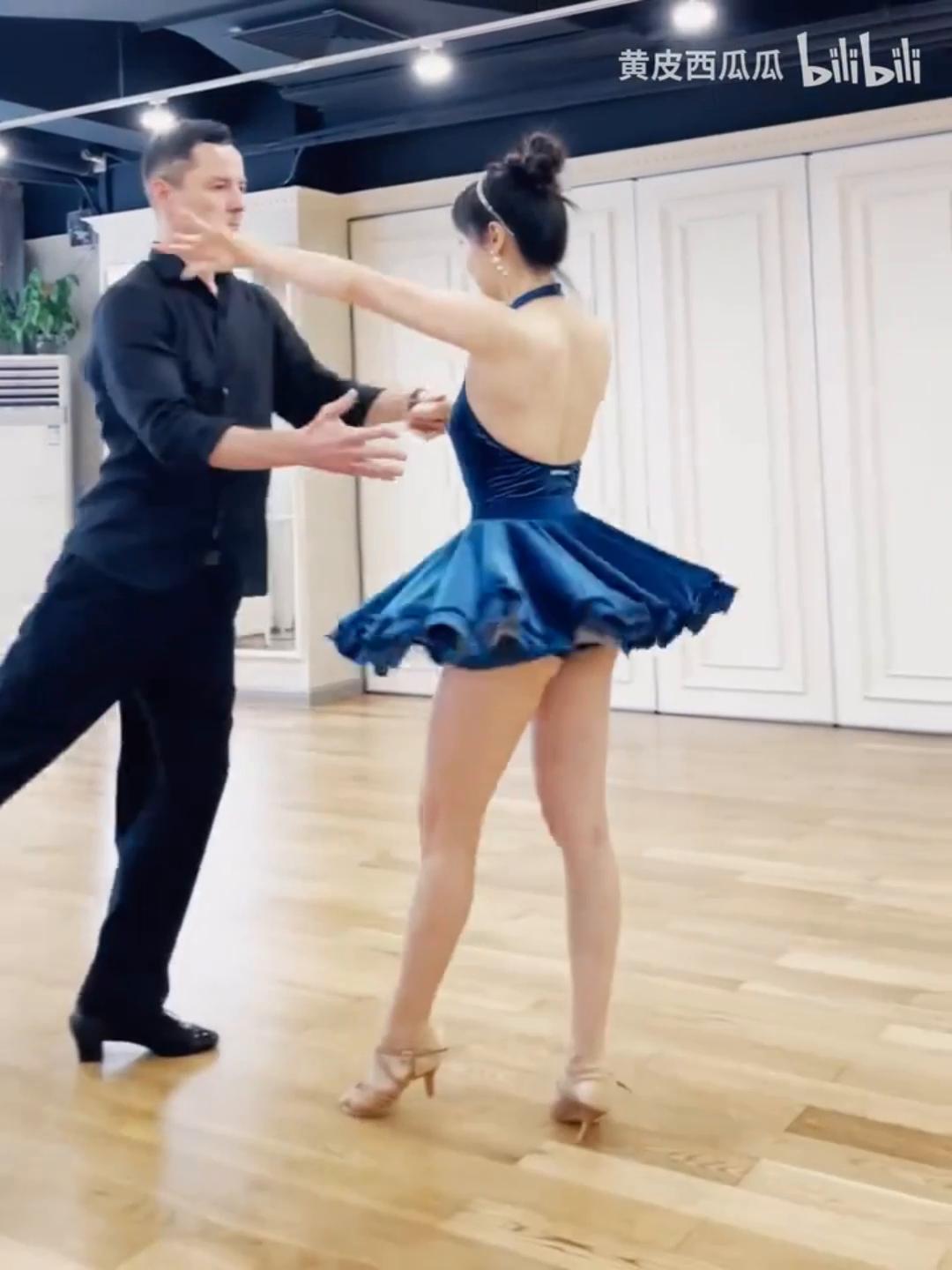} \\

    \includegraphics[width=1.75cm]{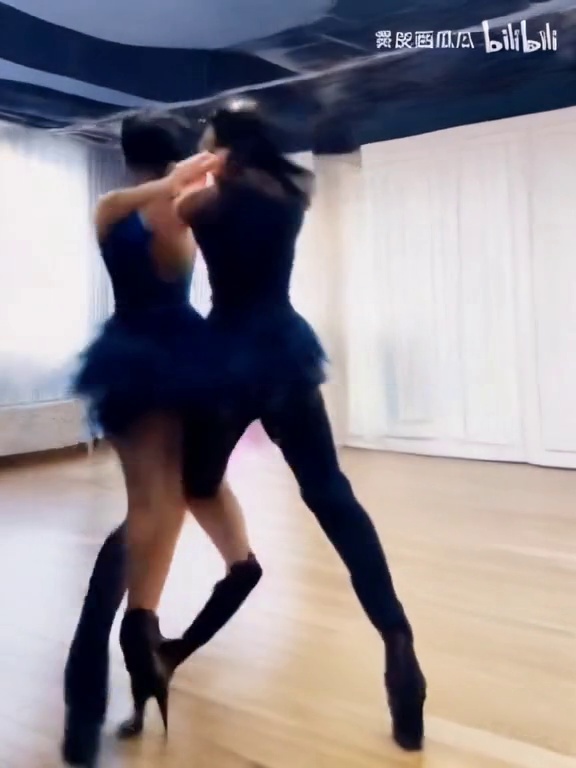}
    &\includegraphics[width=1.75cm]{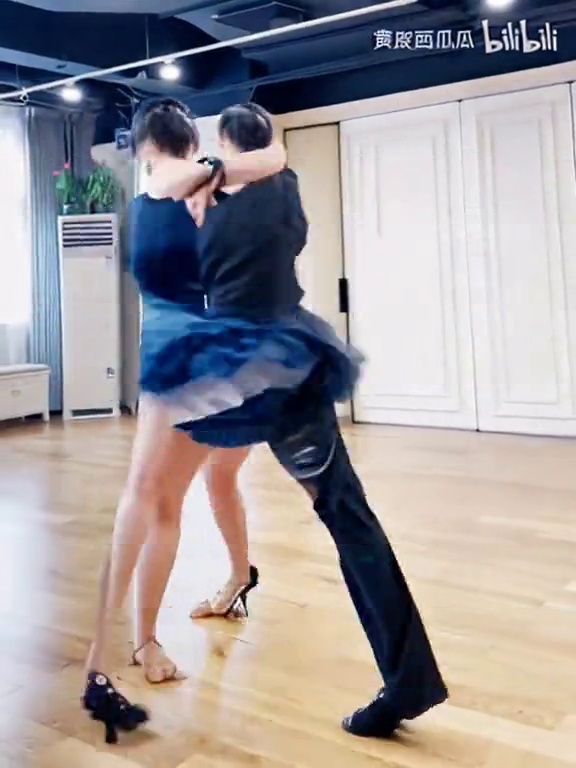}
    &\includegraphics[width=1.75cm]{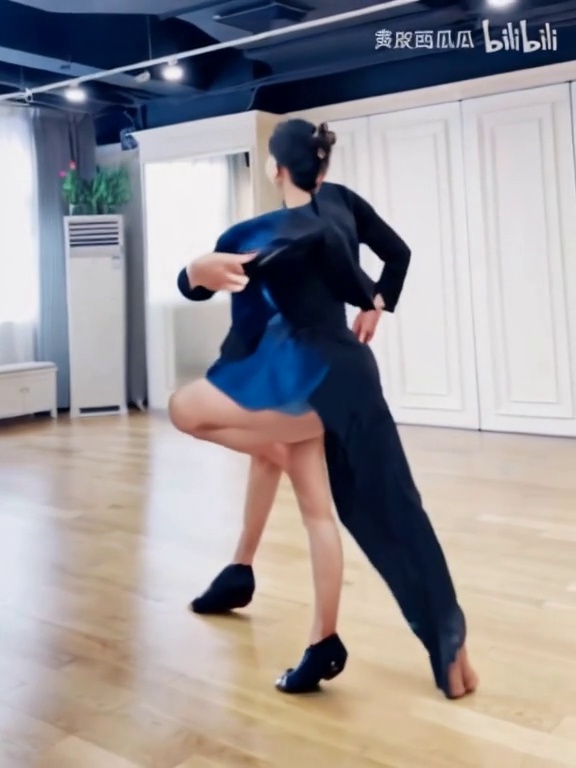}
    &\includegraphics[width=1.75cm]{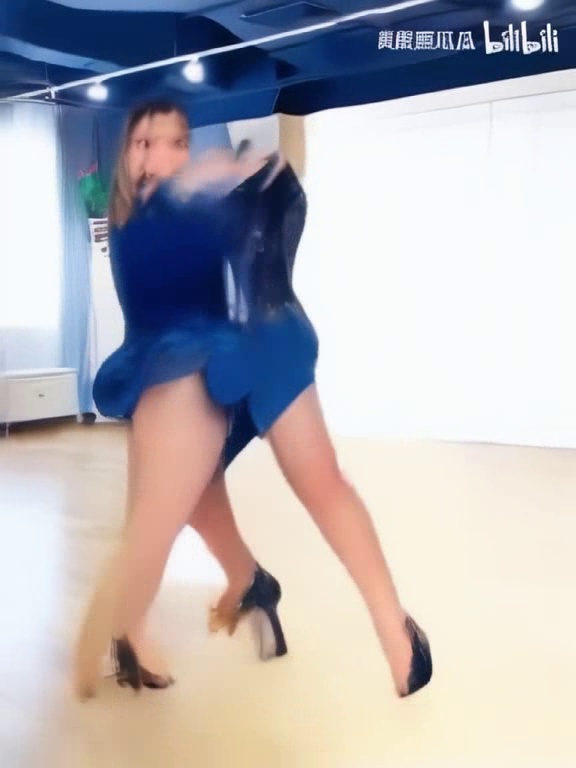}
    &\includegraphics[width=1.75cm]{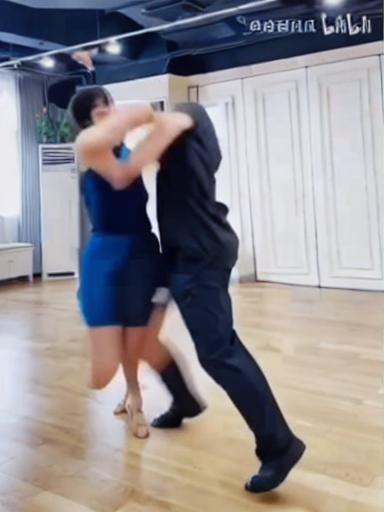}
    &\includegraphics[width=1.75cm]{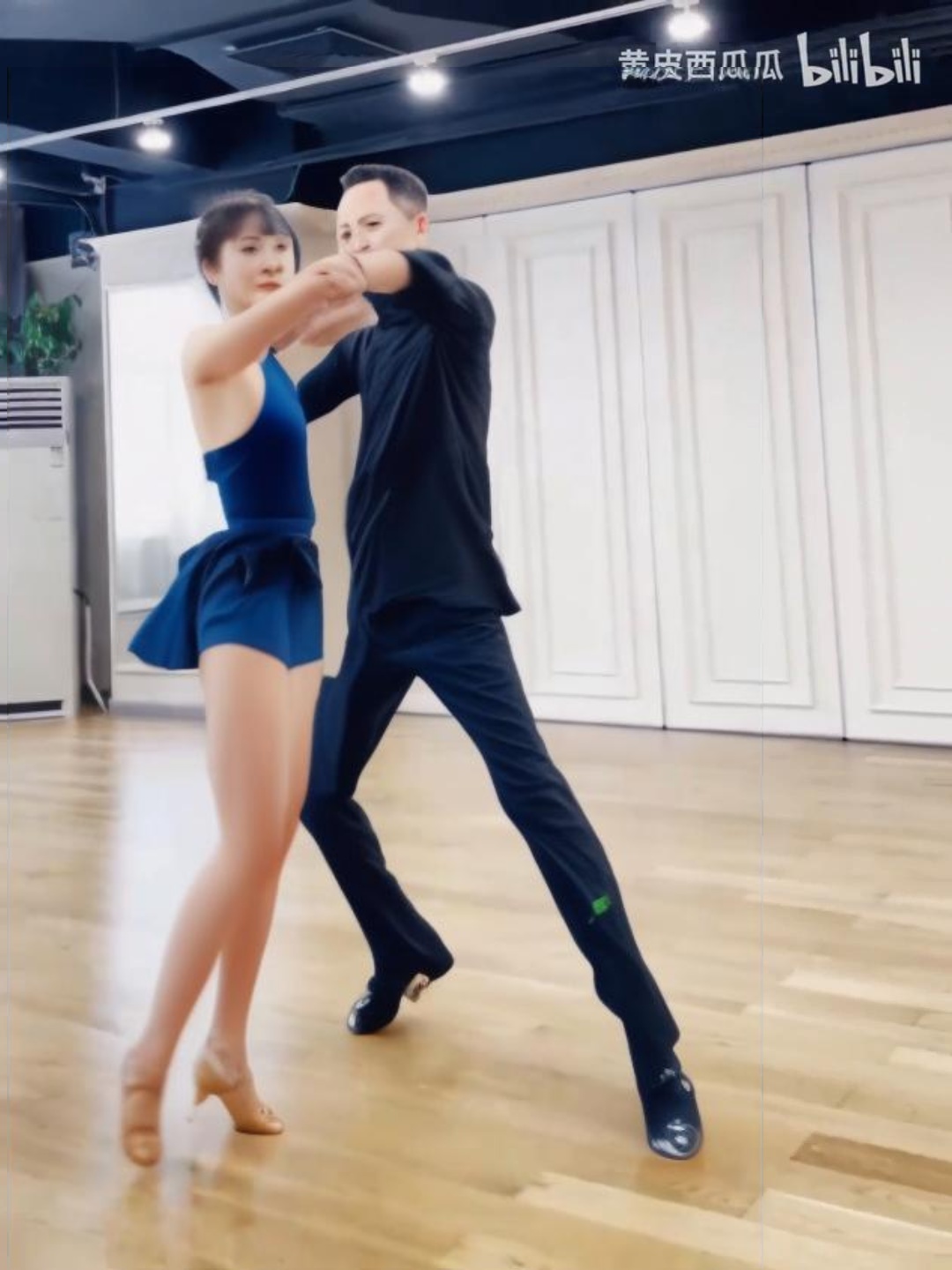}
    &\includegraphics[width=1.75cm]{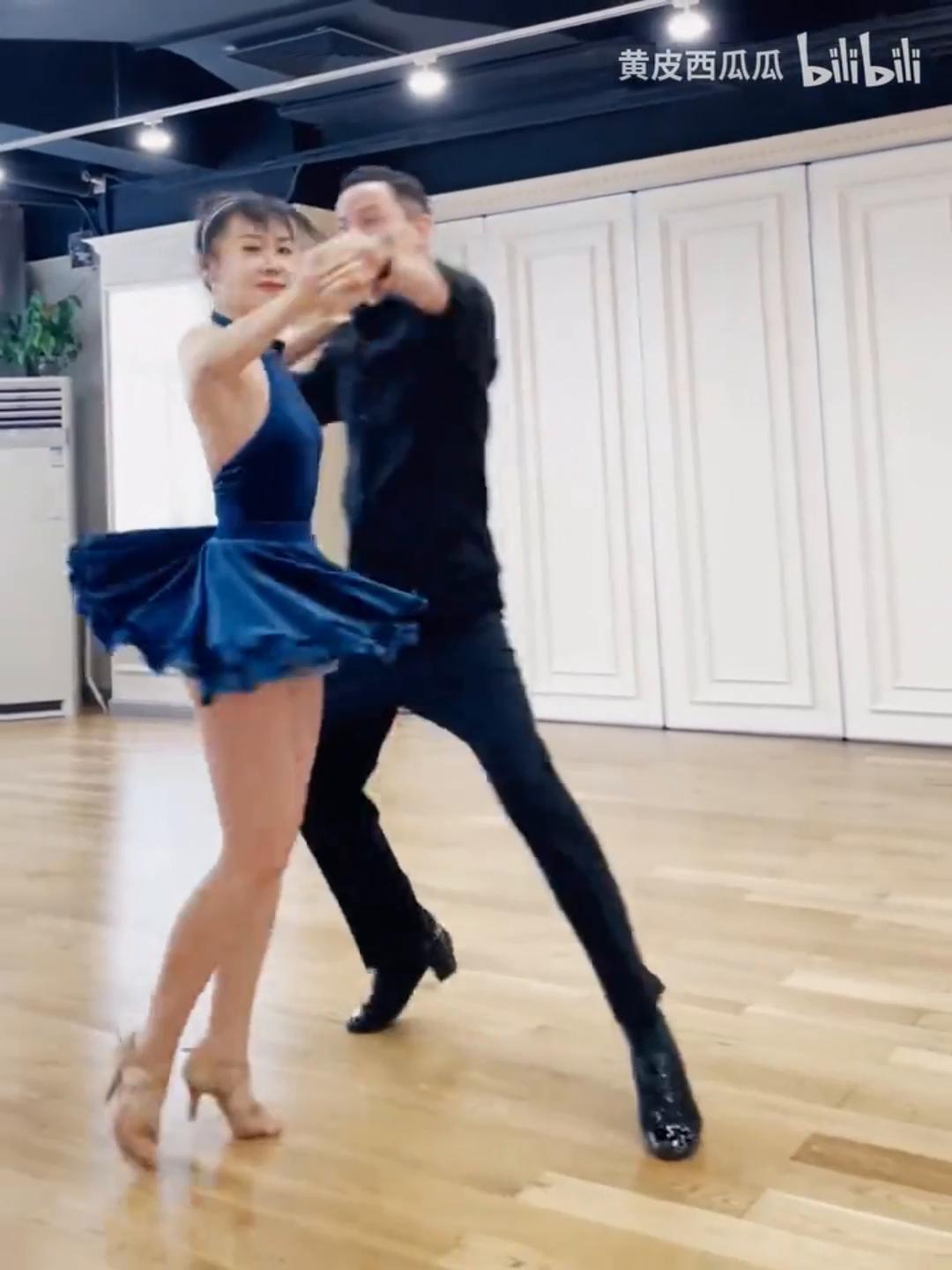} \\
    \end{tabular}
    \end{center}
    \caption{Qualitative Comparison of multiple character animation. Our method successfully animate multiple characters while others exist clear artifacts.} 
    \label{fig:qualitative}
    
\end{figure*}

\begin{figure*}[h]
    \begin{center}
    \setlength{\tabcolsep}{0.5pt}
    \begin{tabular}{m{1.25cm}<{\centering}m{1.25cm}<{\centering}m{1.25cm}<{\centering}m{1.25cm}<{\centering}m{1.25cm}<{\centering}m{0.2cm}<{\centering}m{1.25cm}<{\centering}m{1.25cm}<{\centering}m{1.25cm}<{\centering}m{1.25cm}<{\centering}m{1.25cm}<{\centering}}

    \includegraphics[width=1.225cm]{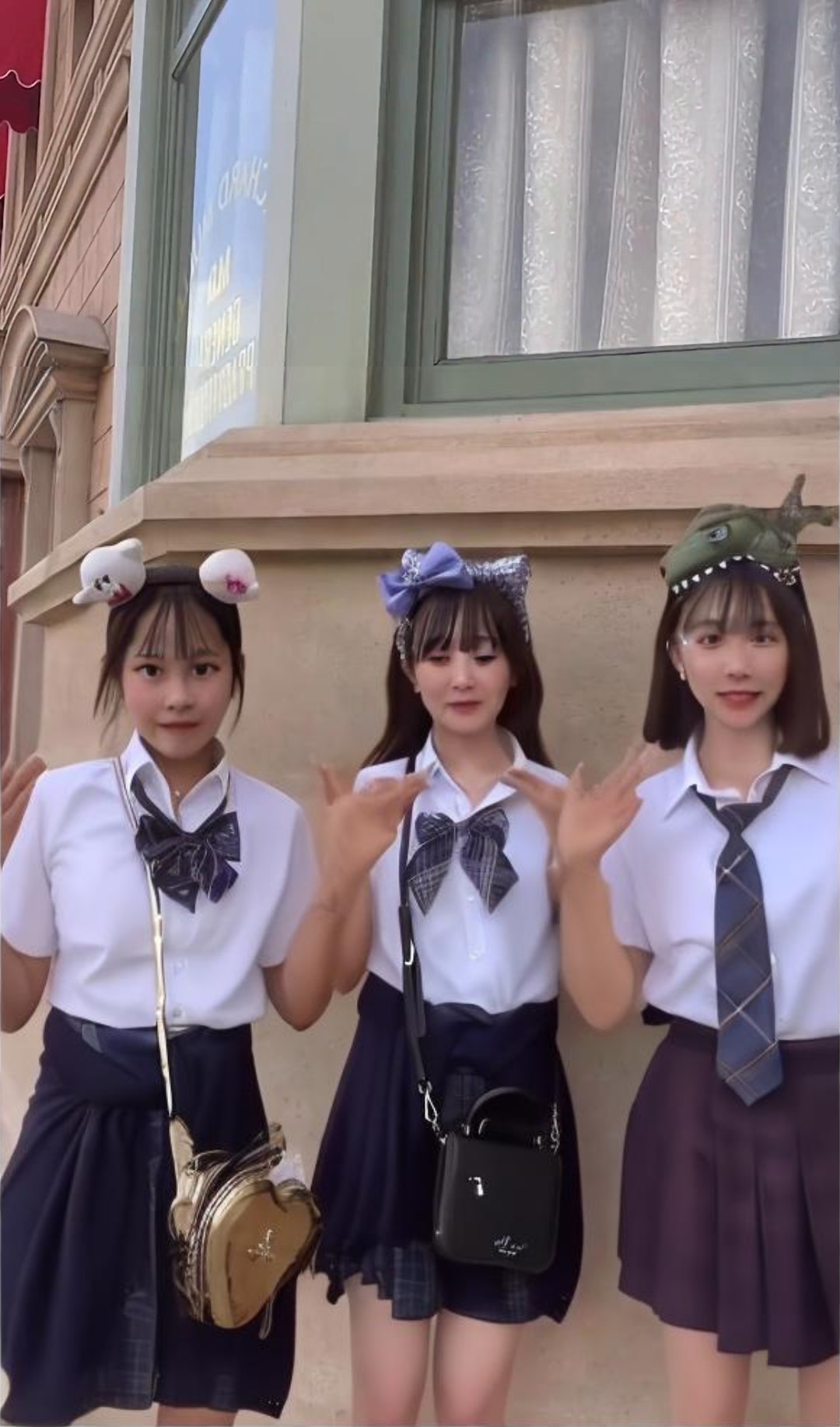}
    &\includegraphics[width=1.225cm]{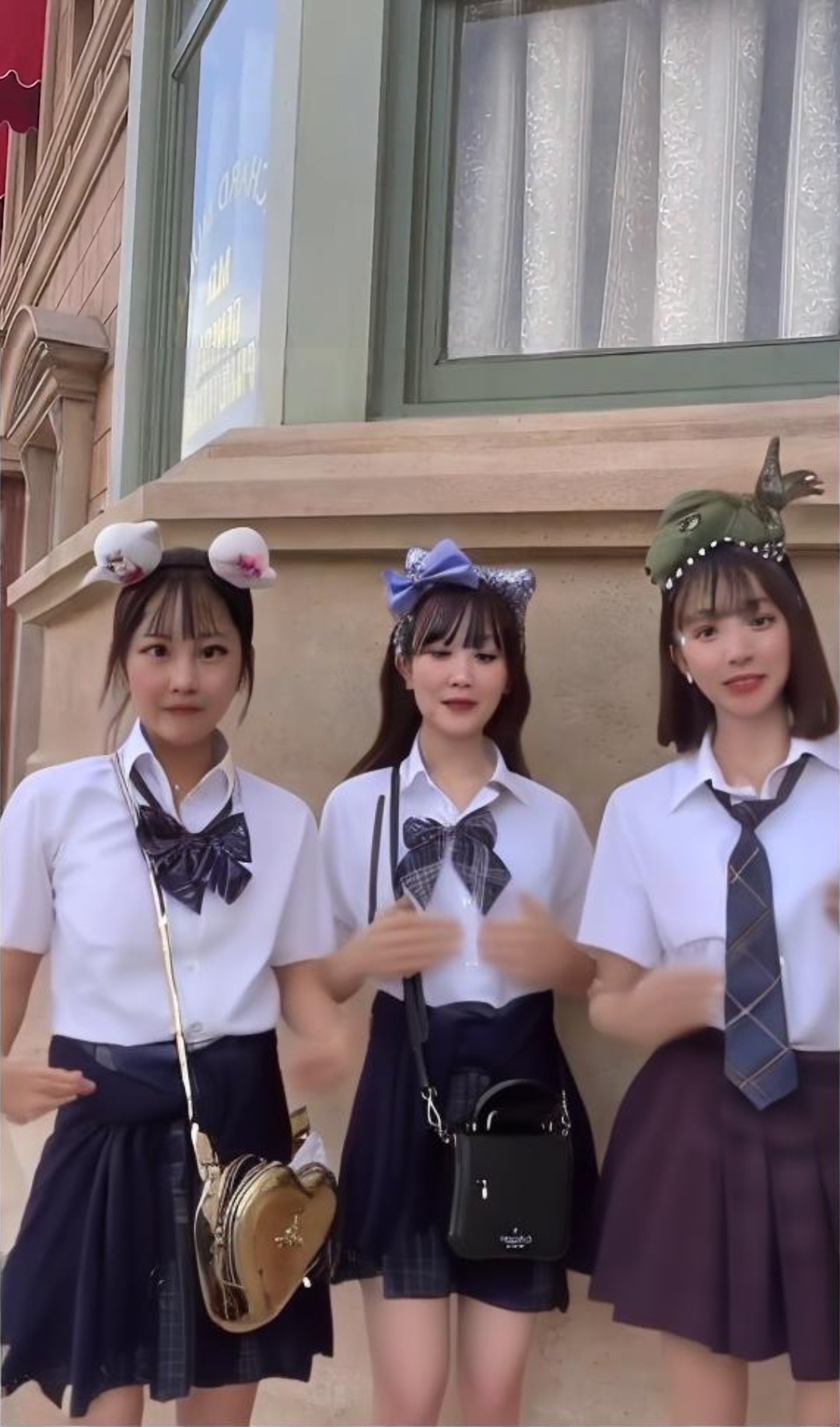}
    &\includegraphics[width=1.225cm]{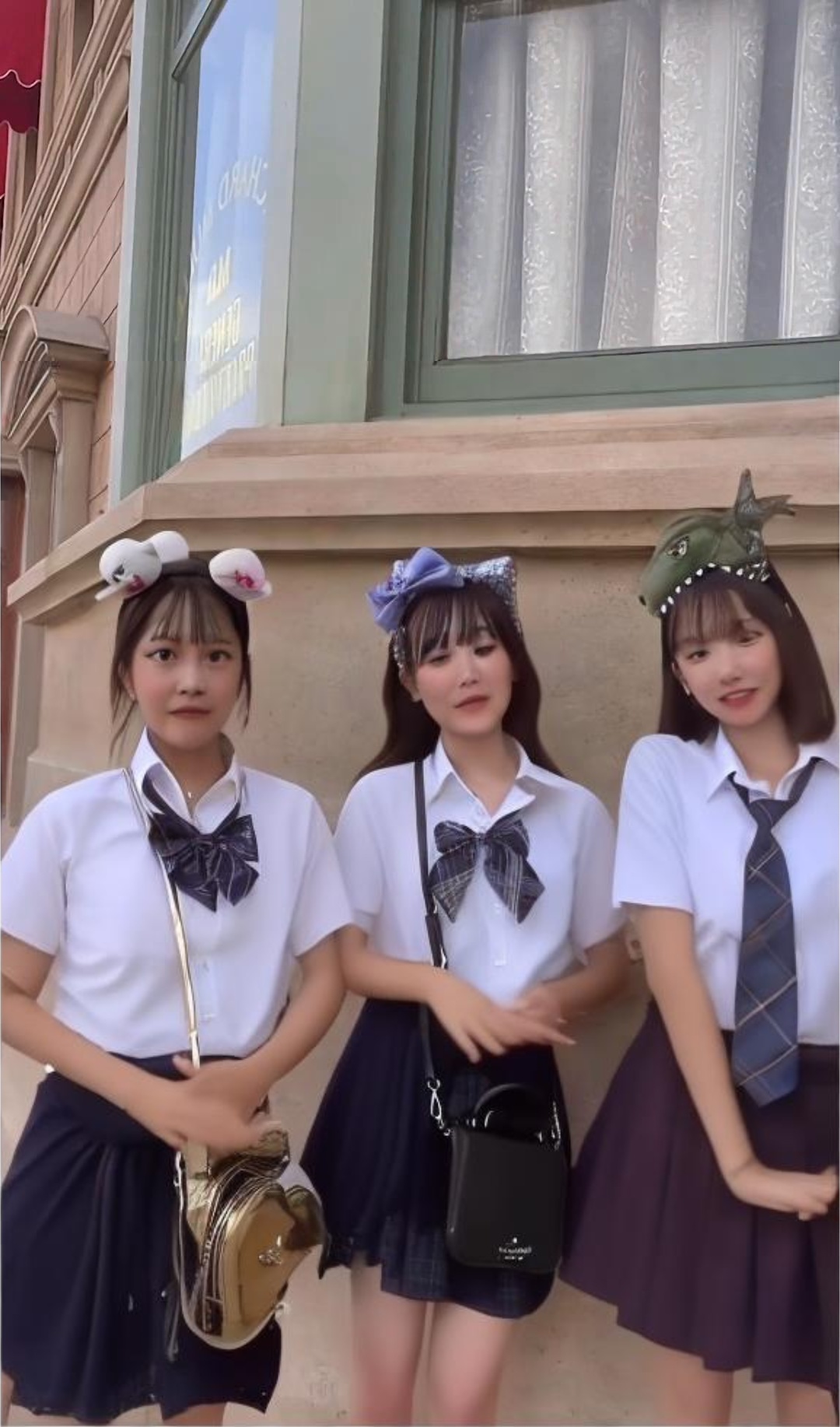}
    &\includegraphics[width=1.225cm]{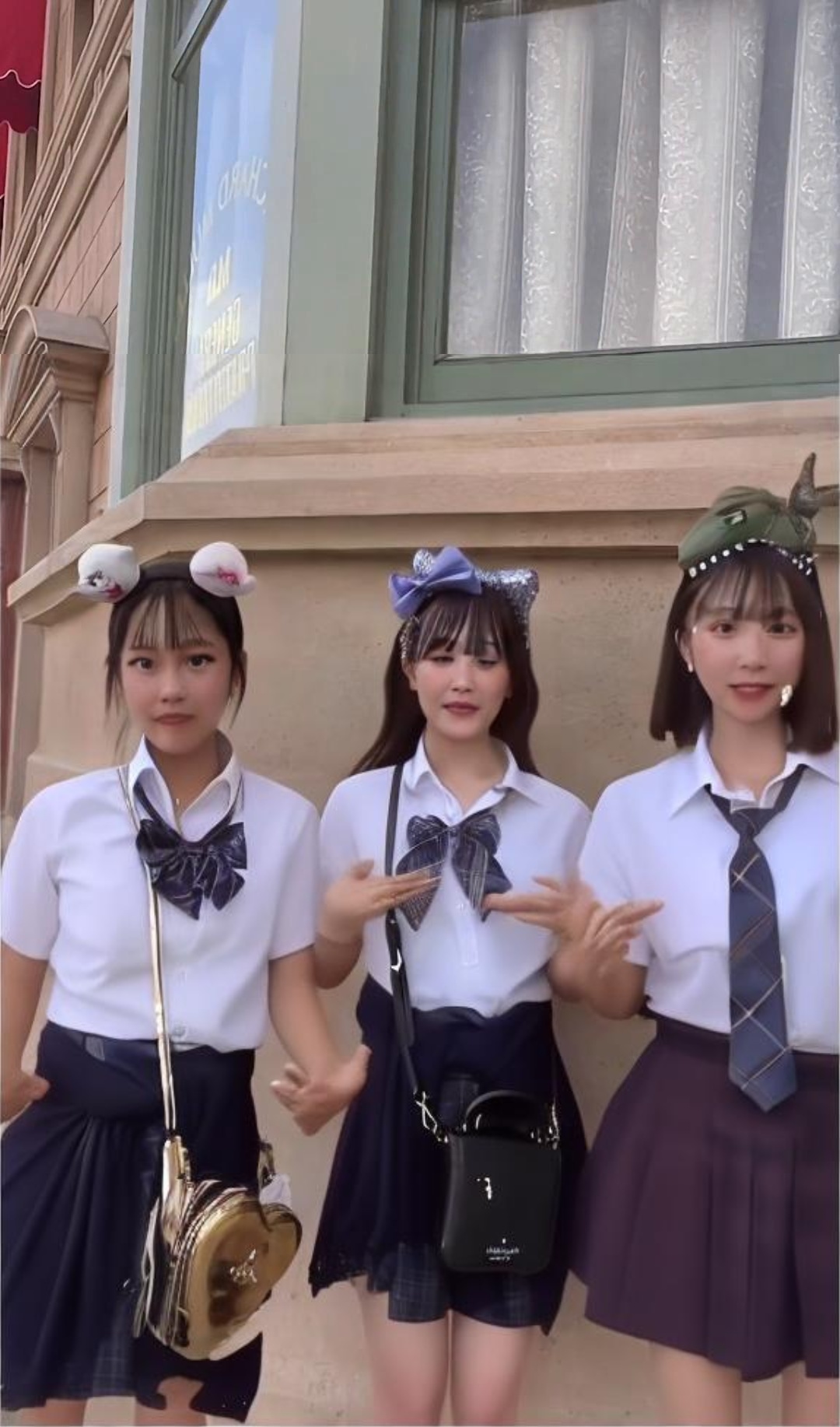}
    &\includegraphics[width=1.225cm]{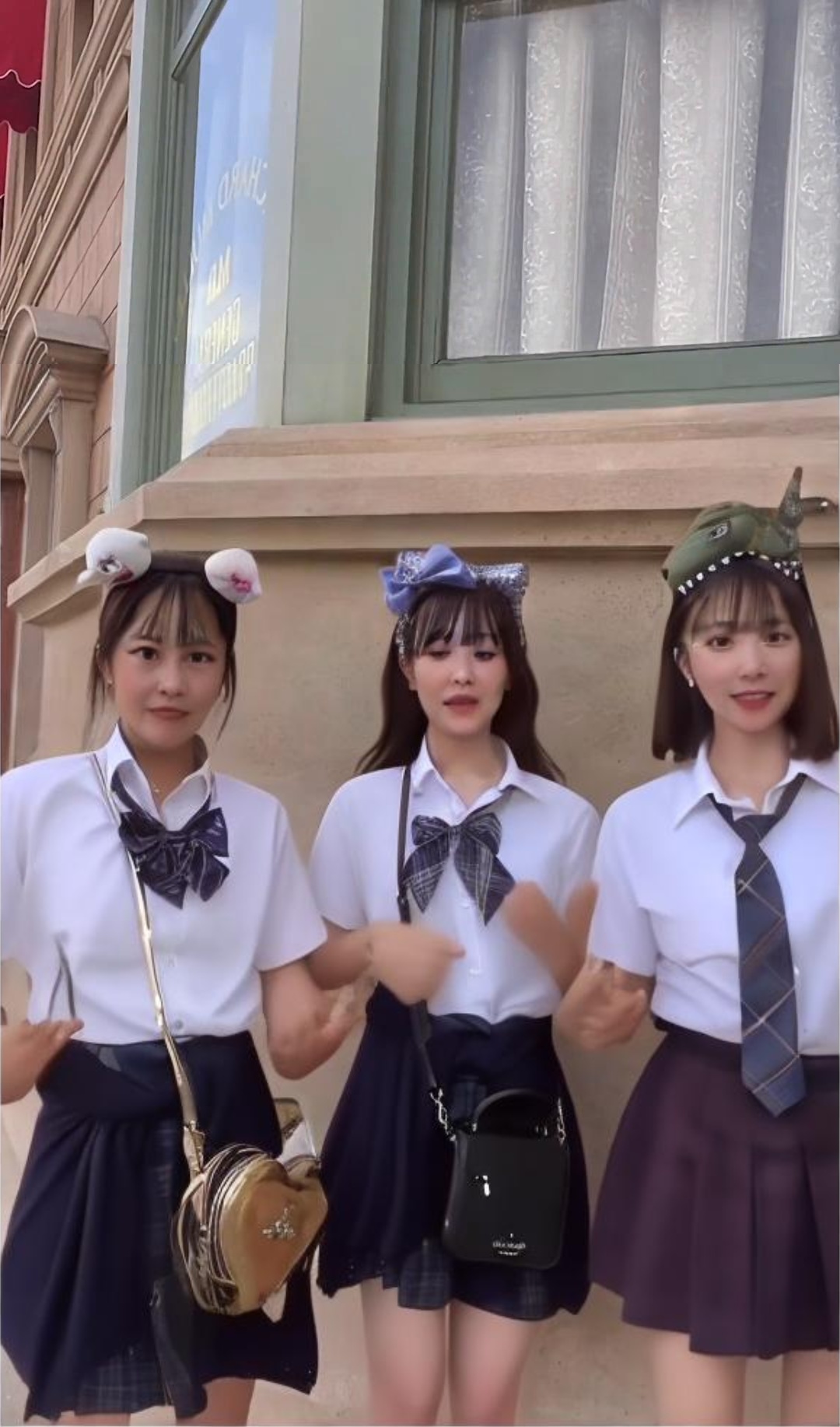}
    &
    &\includegraphics[width=1.225cm]{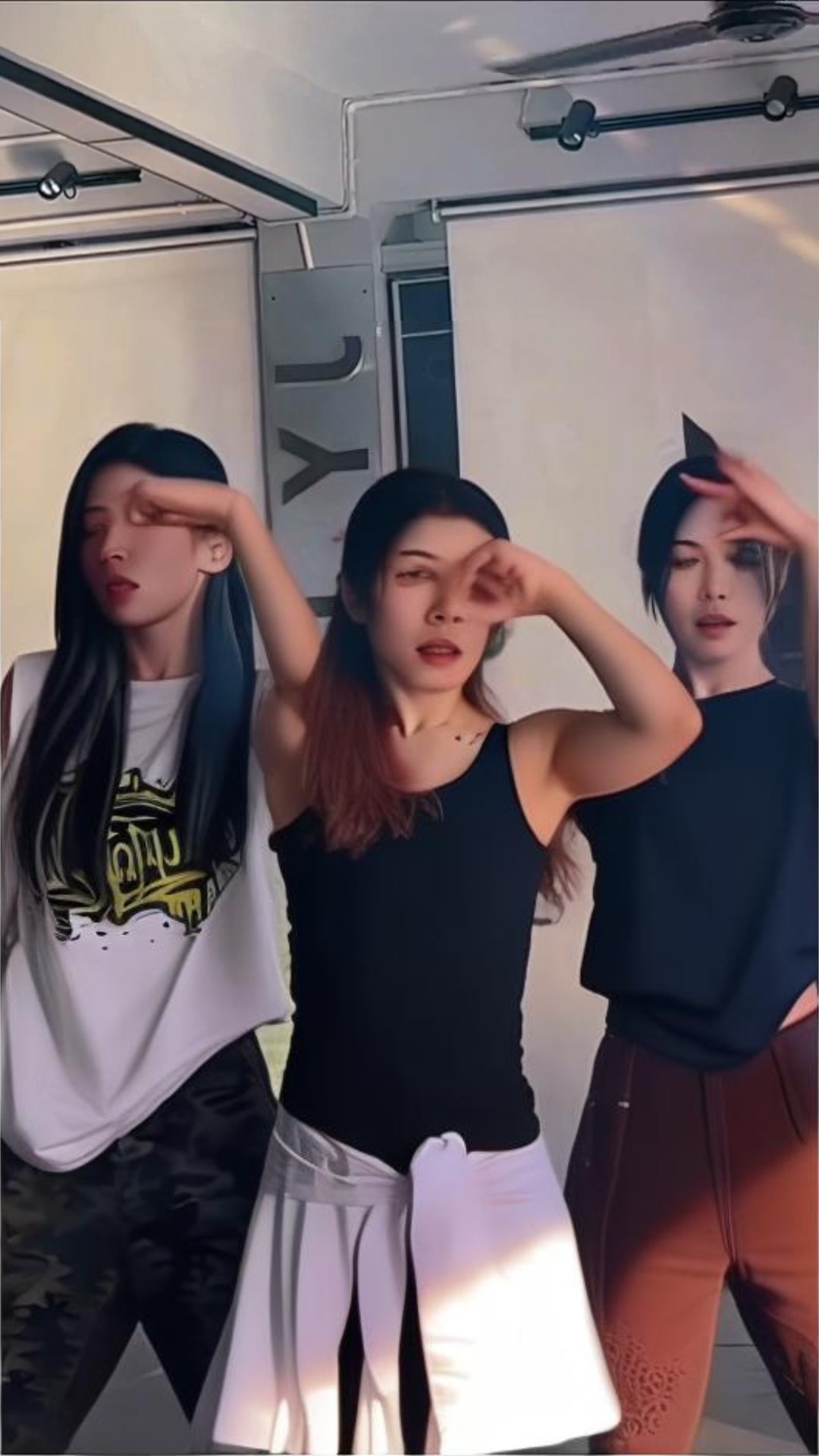}
    &\includegraphics[width=1.225cm]{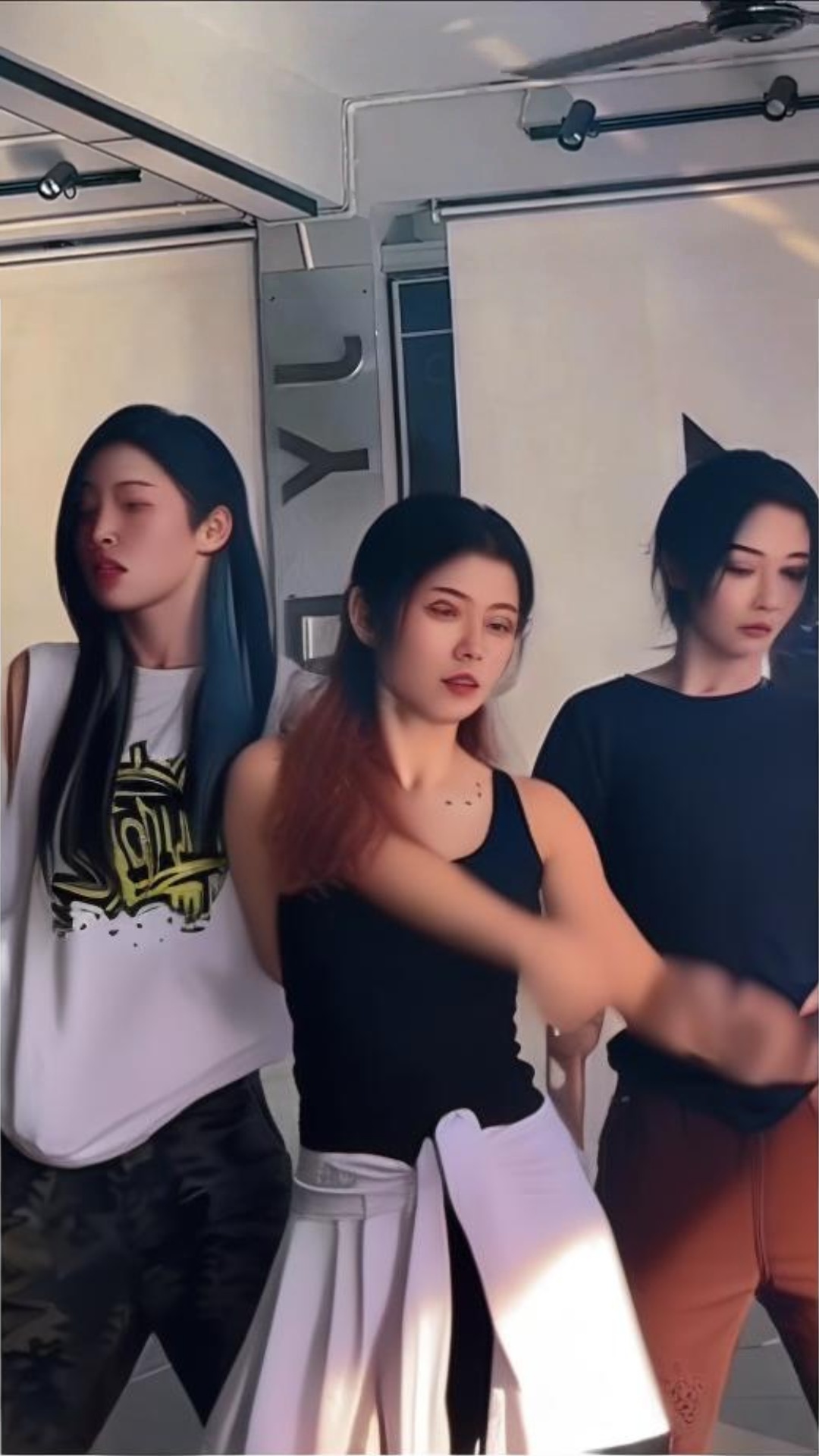}
    &\includegraphics[width=1.225cm]{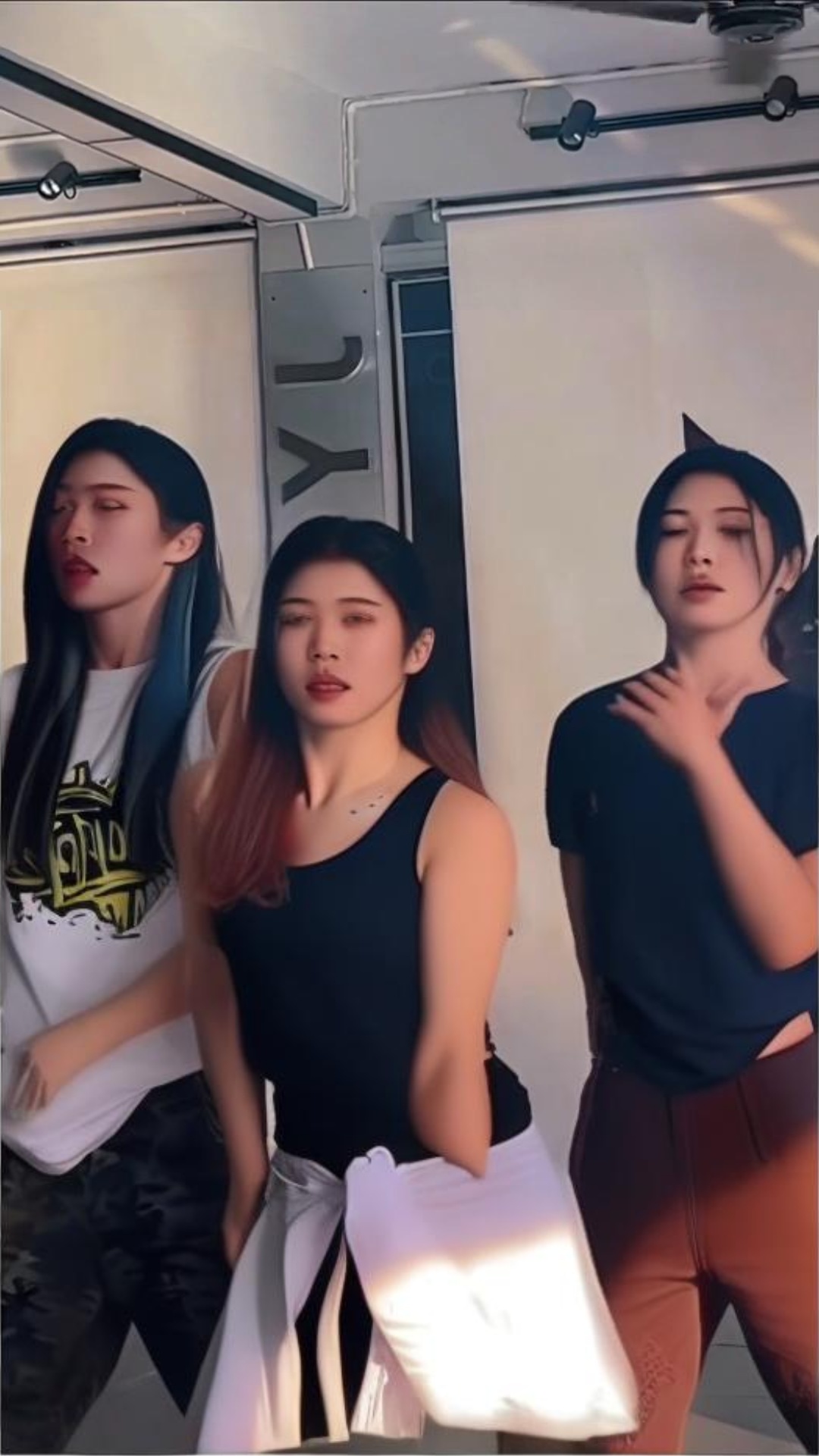}
    &\includegraphics[width=1.225cm]{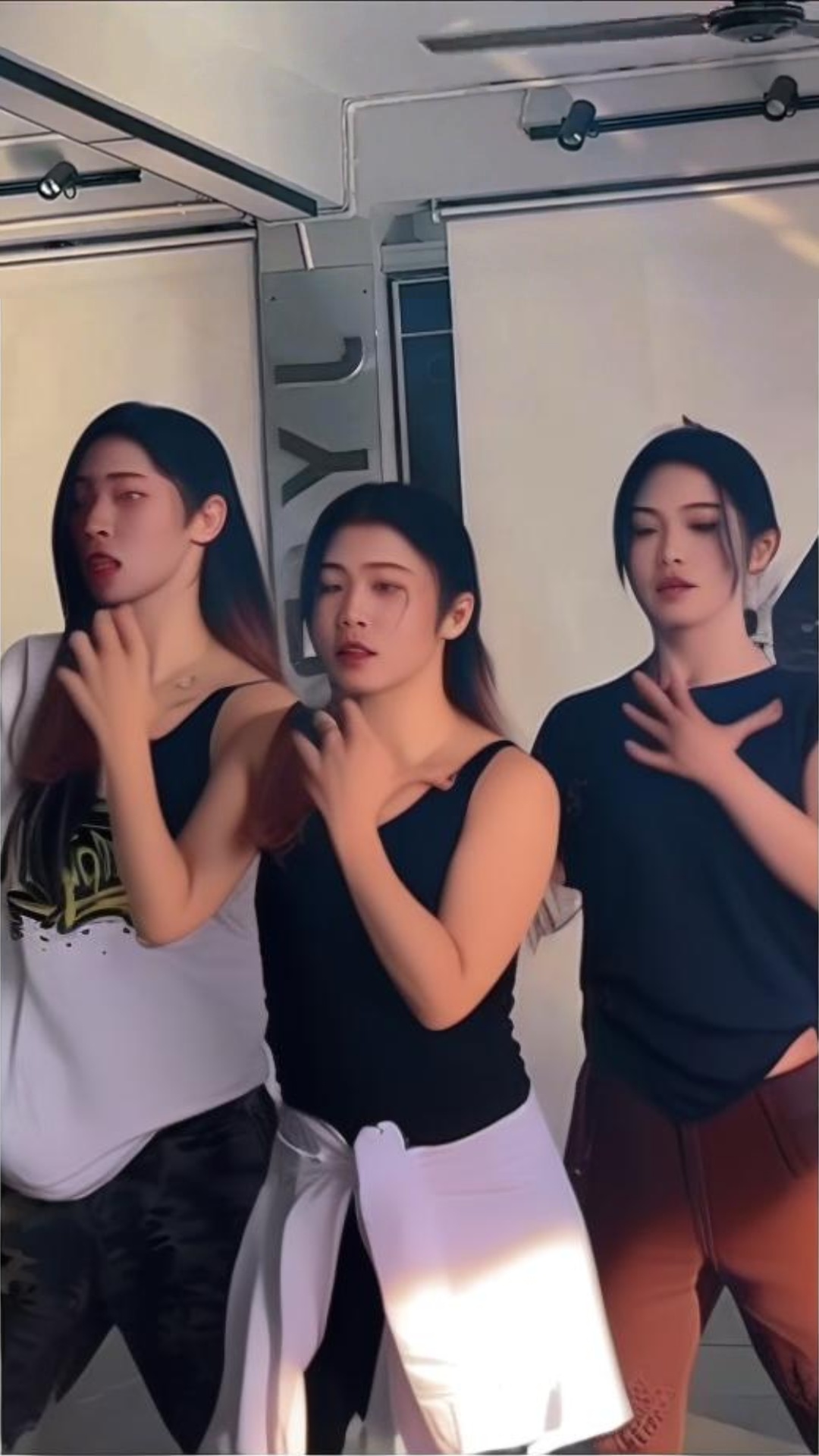}
    &\includegraphics[width=1.225cm]{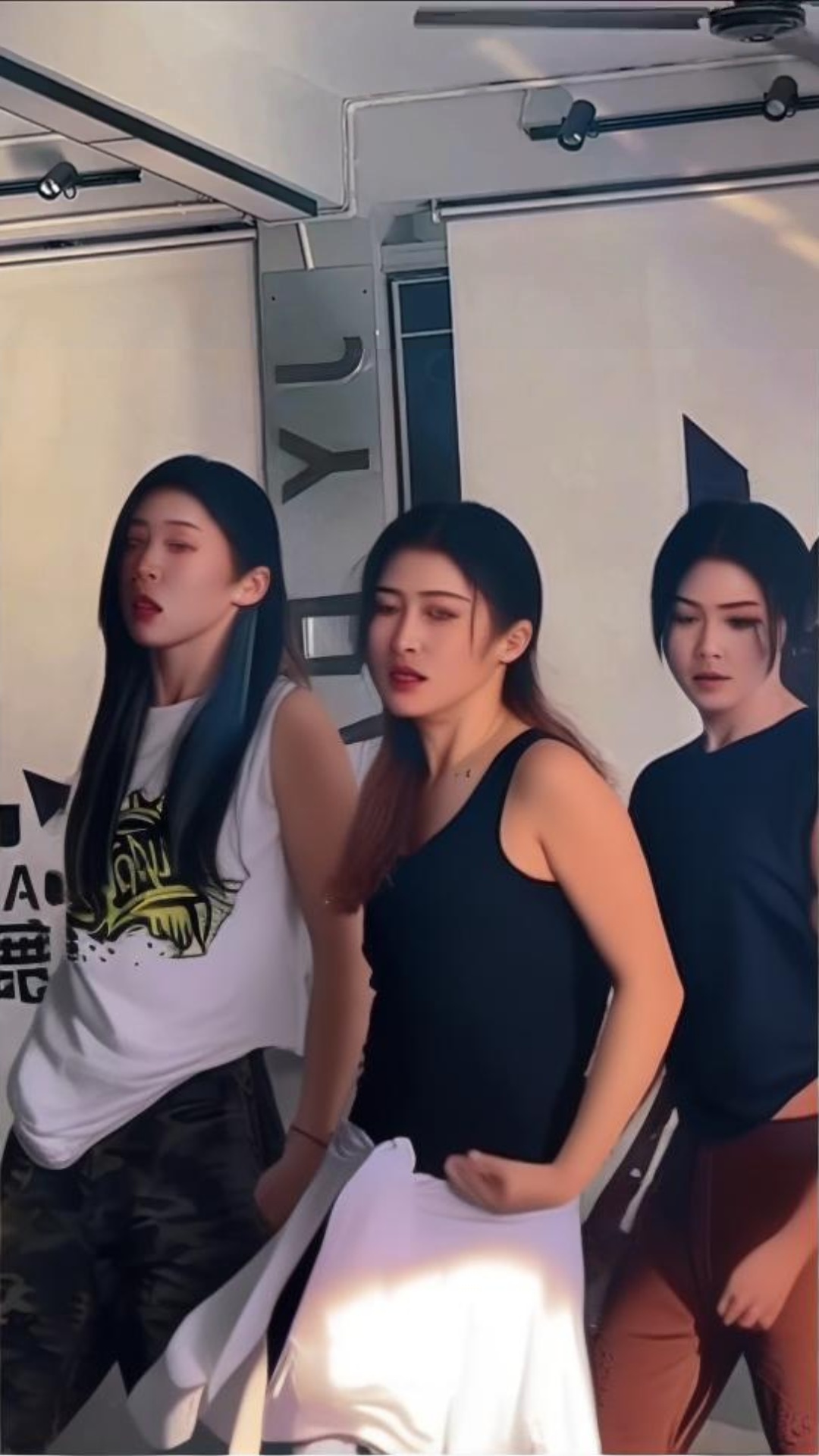}\\

    \includegraphics[width=1.225cm]{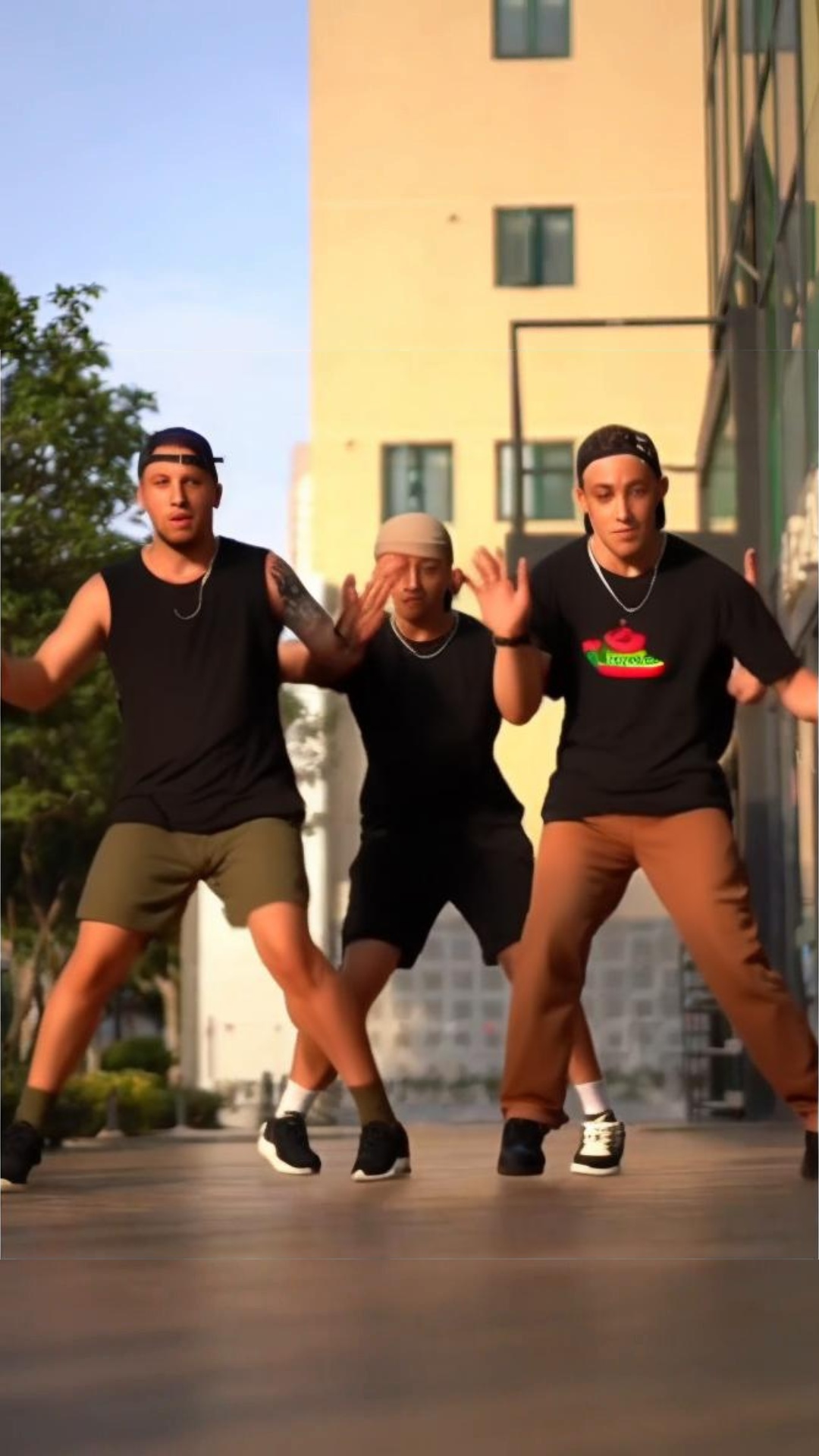}
    &\includegraphics[width=1.225cm]{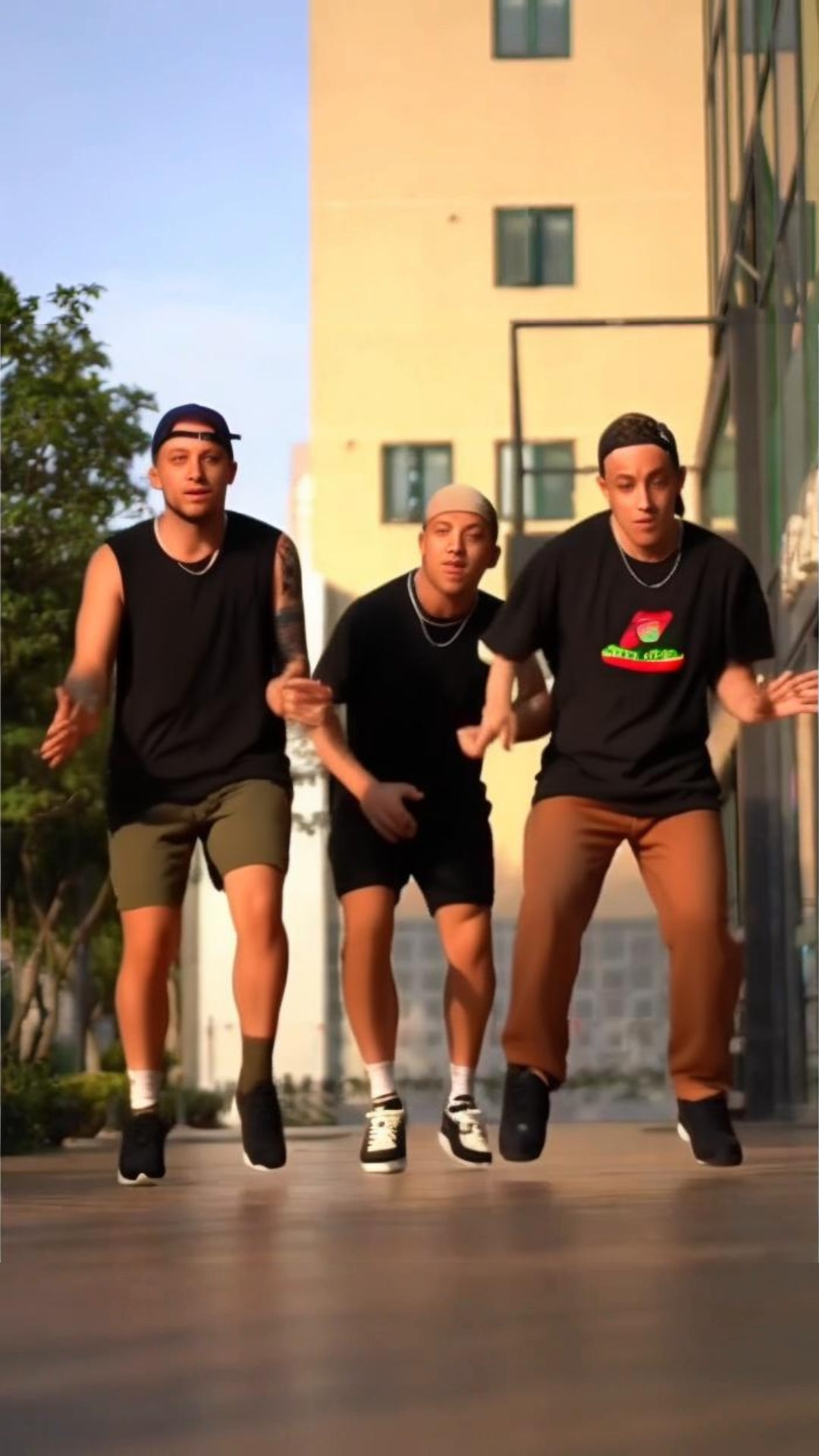}
    &\includegraphics[width=1.225cm]{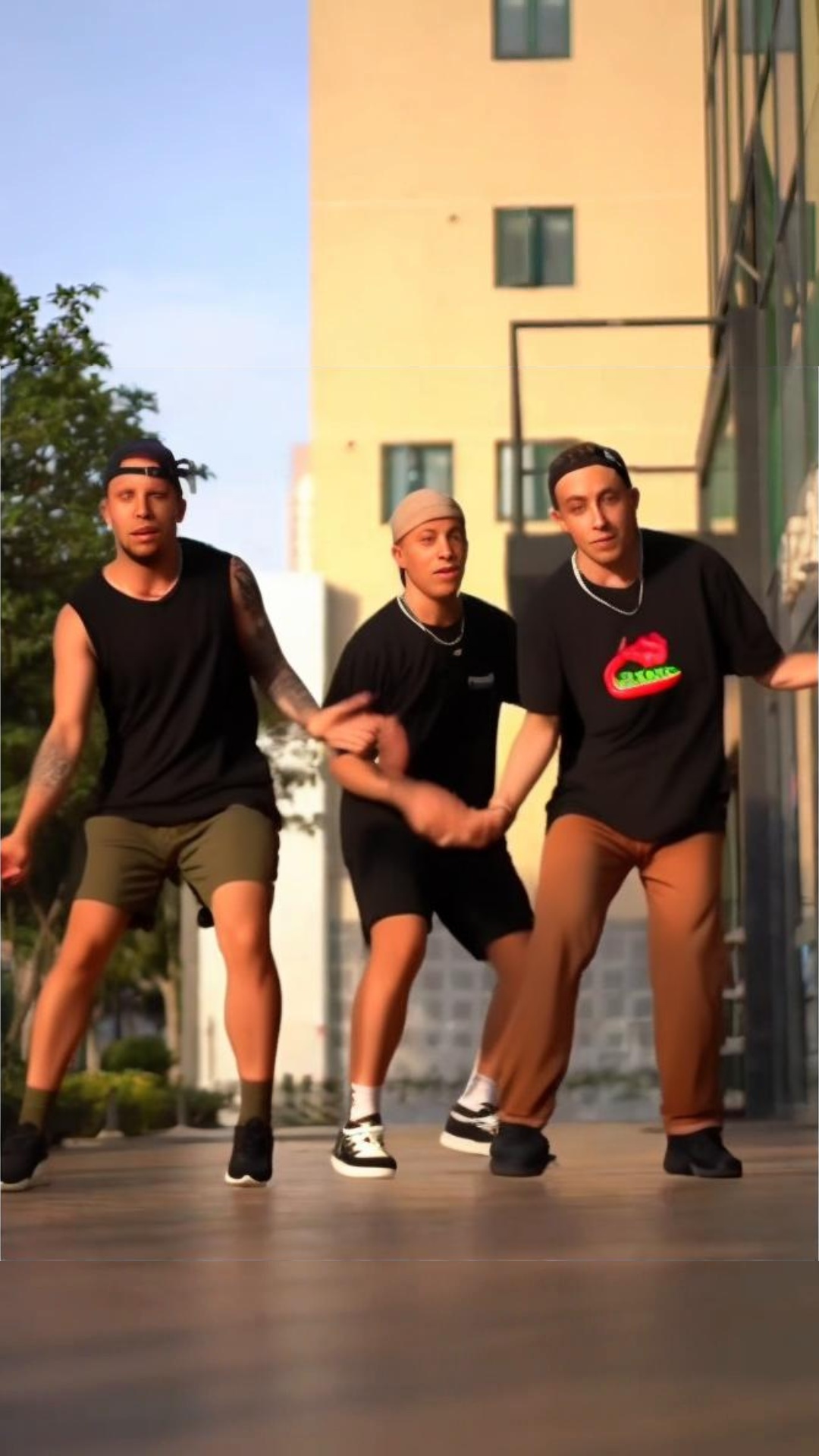}
    &\includegraphics[width=1.225cm]{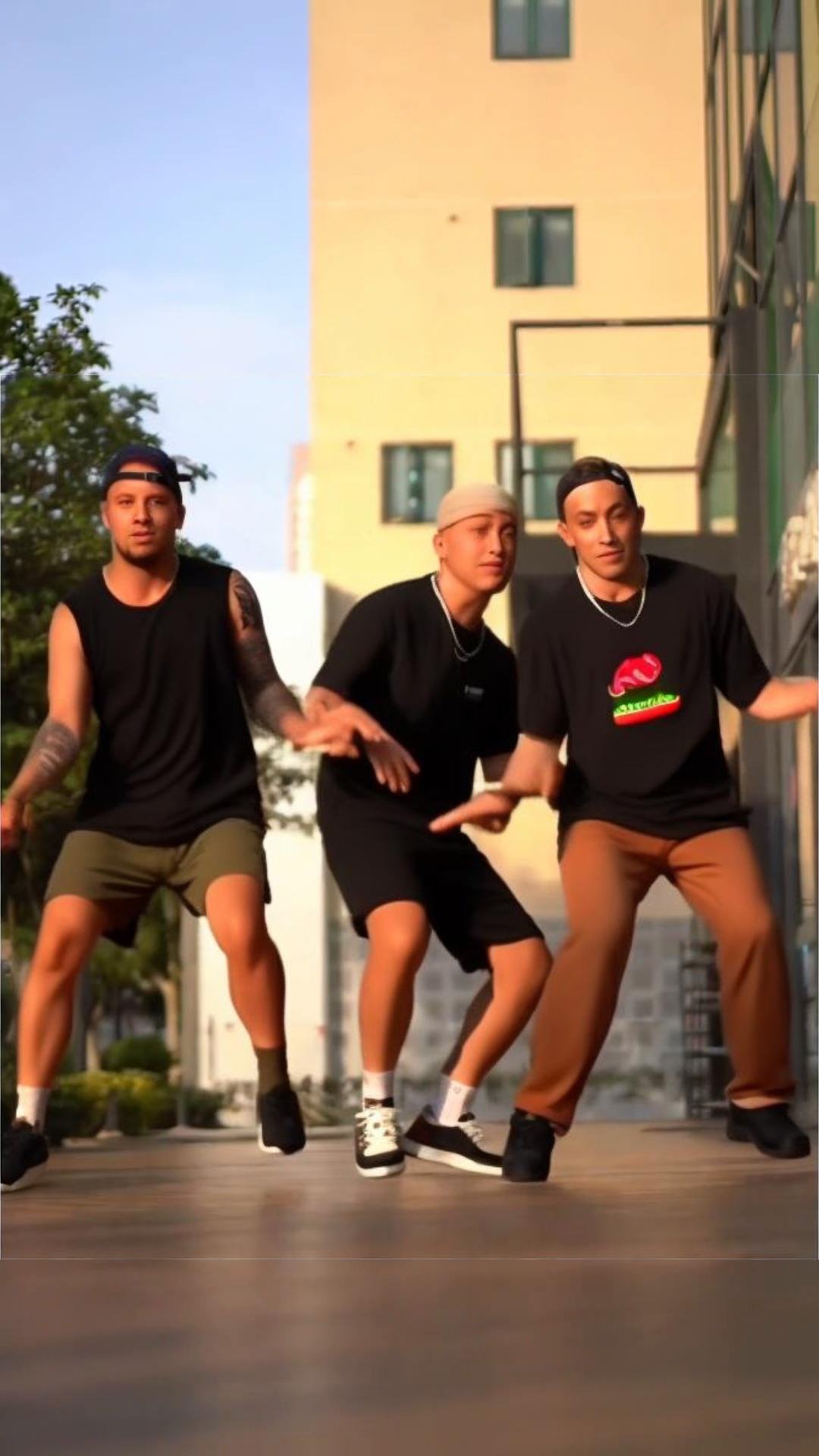}
    &\includegraphics[width=1.225cm]{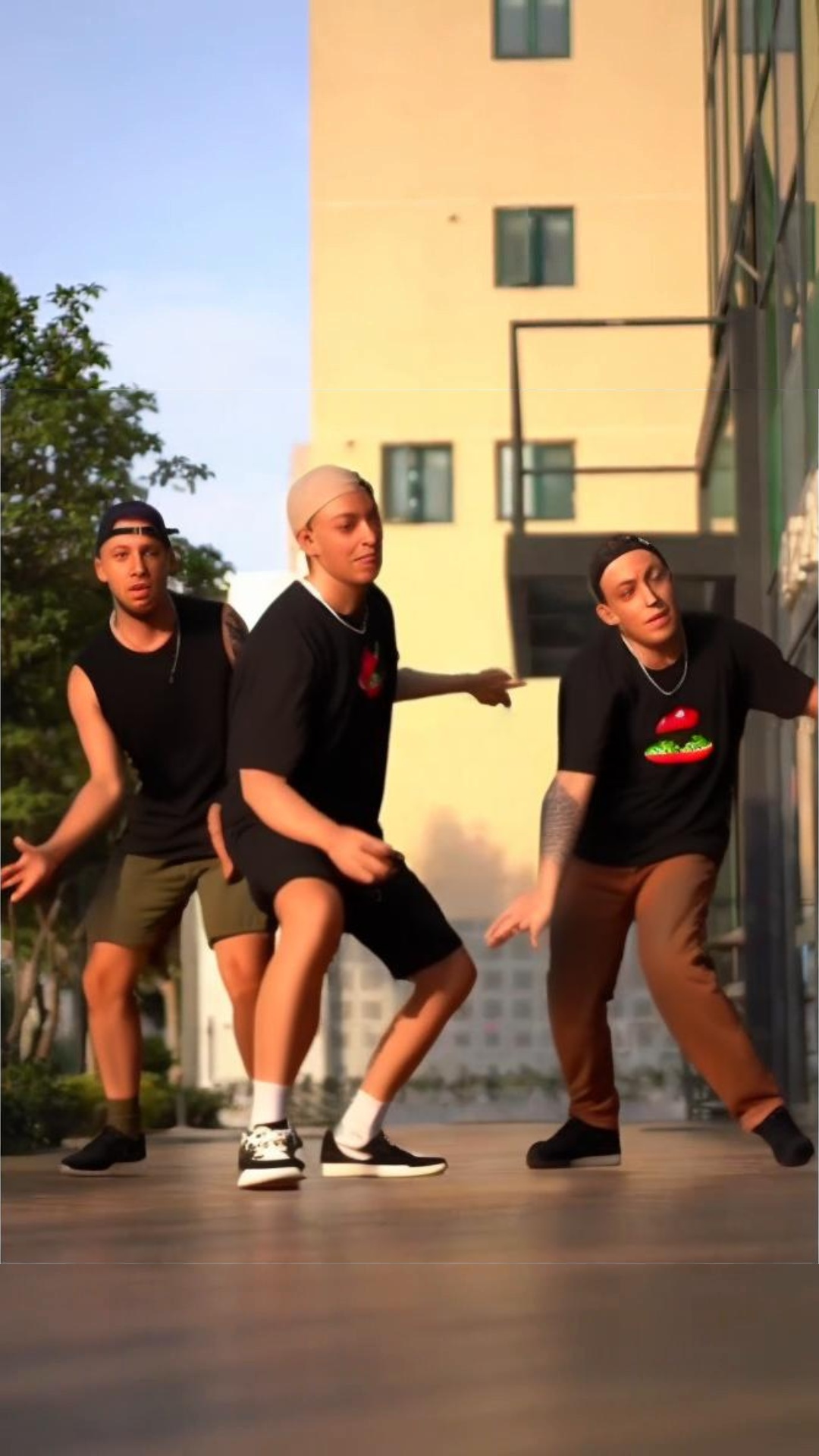}
    &
    &\includegraphics[width=1.225cm]{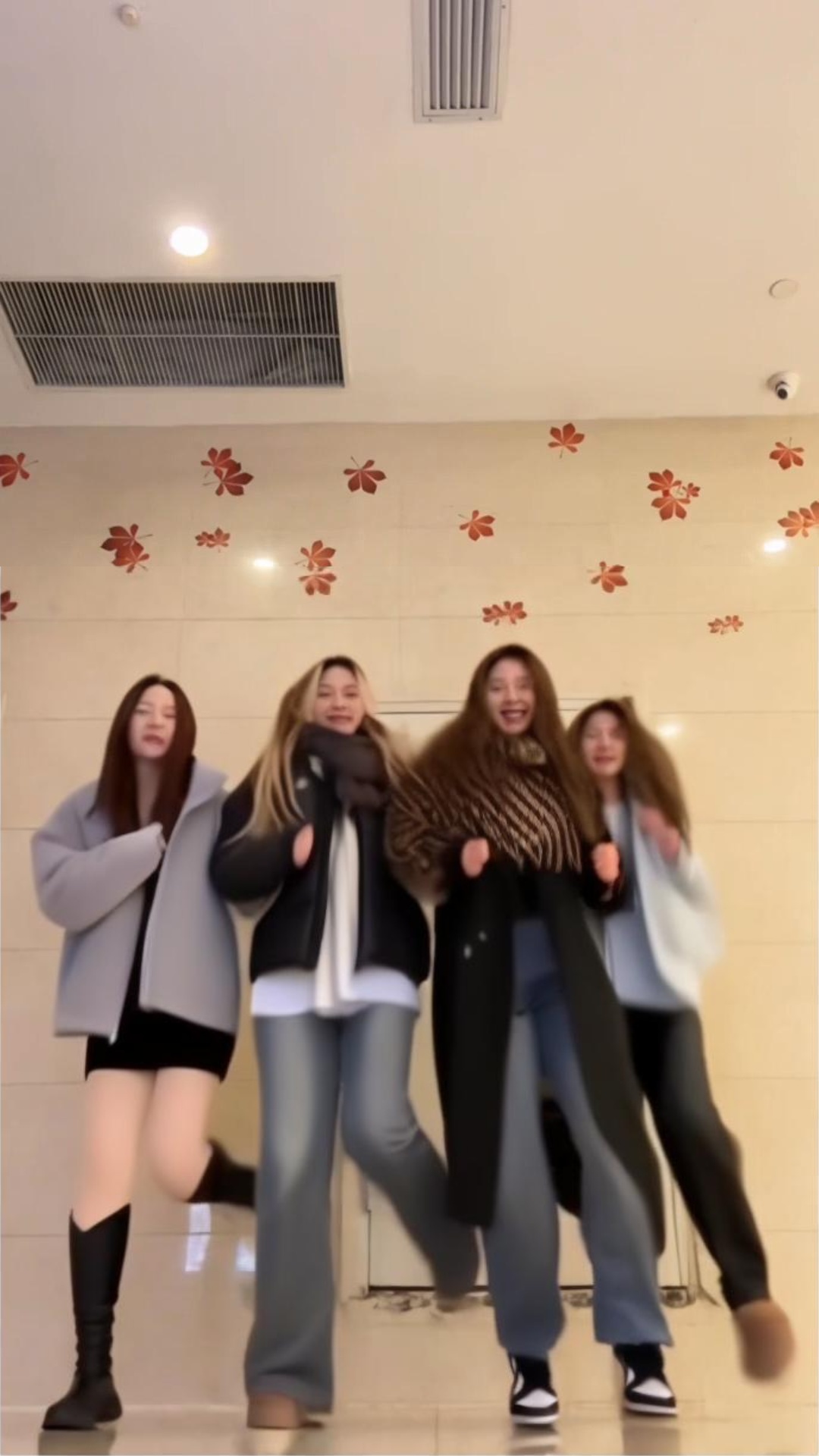}
    &\includegraphics[width=1.225cm]{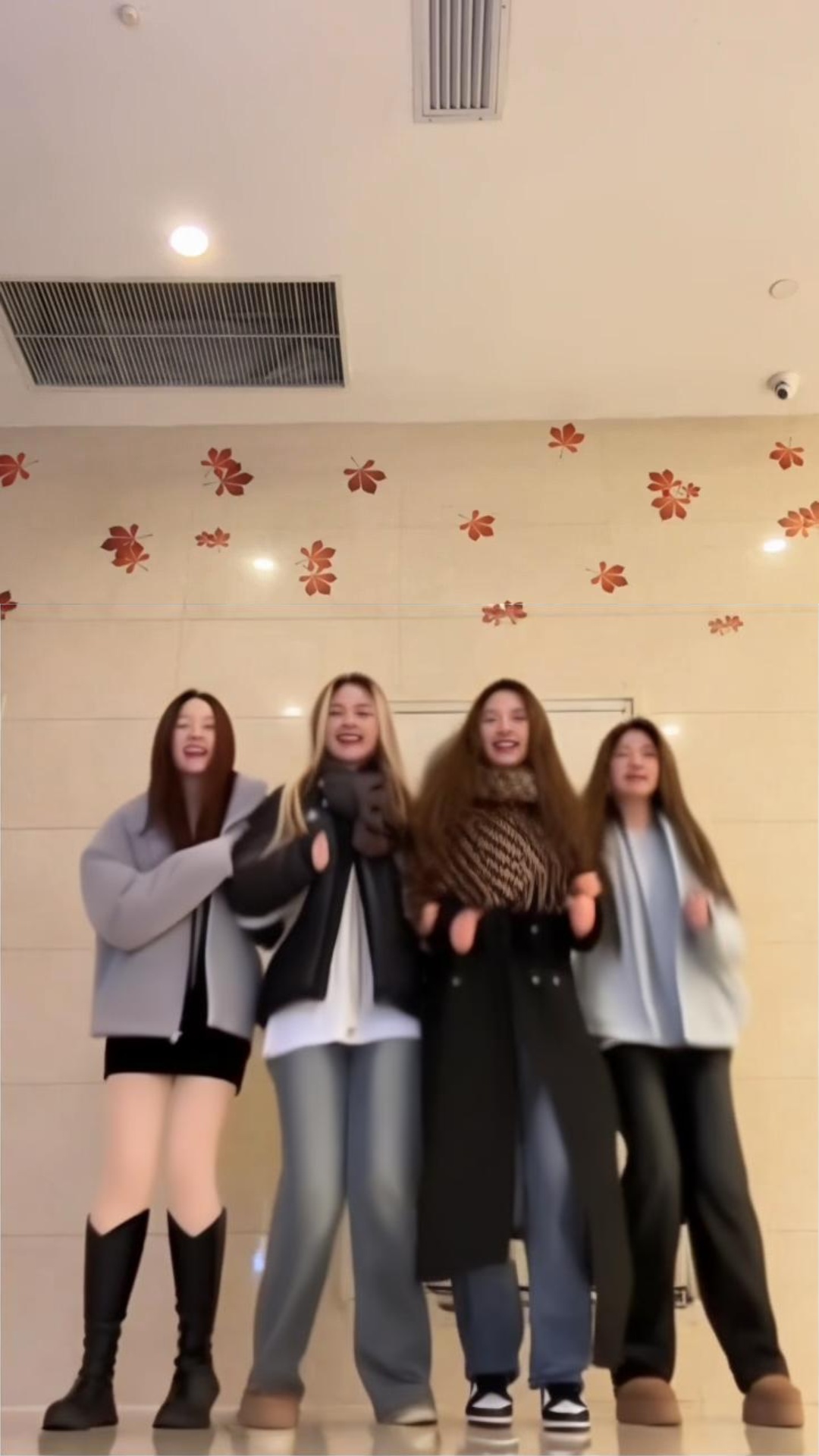}
    &\includegraphics[width=1.225cm]{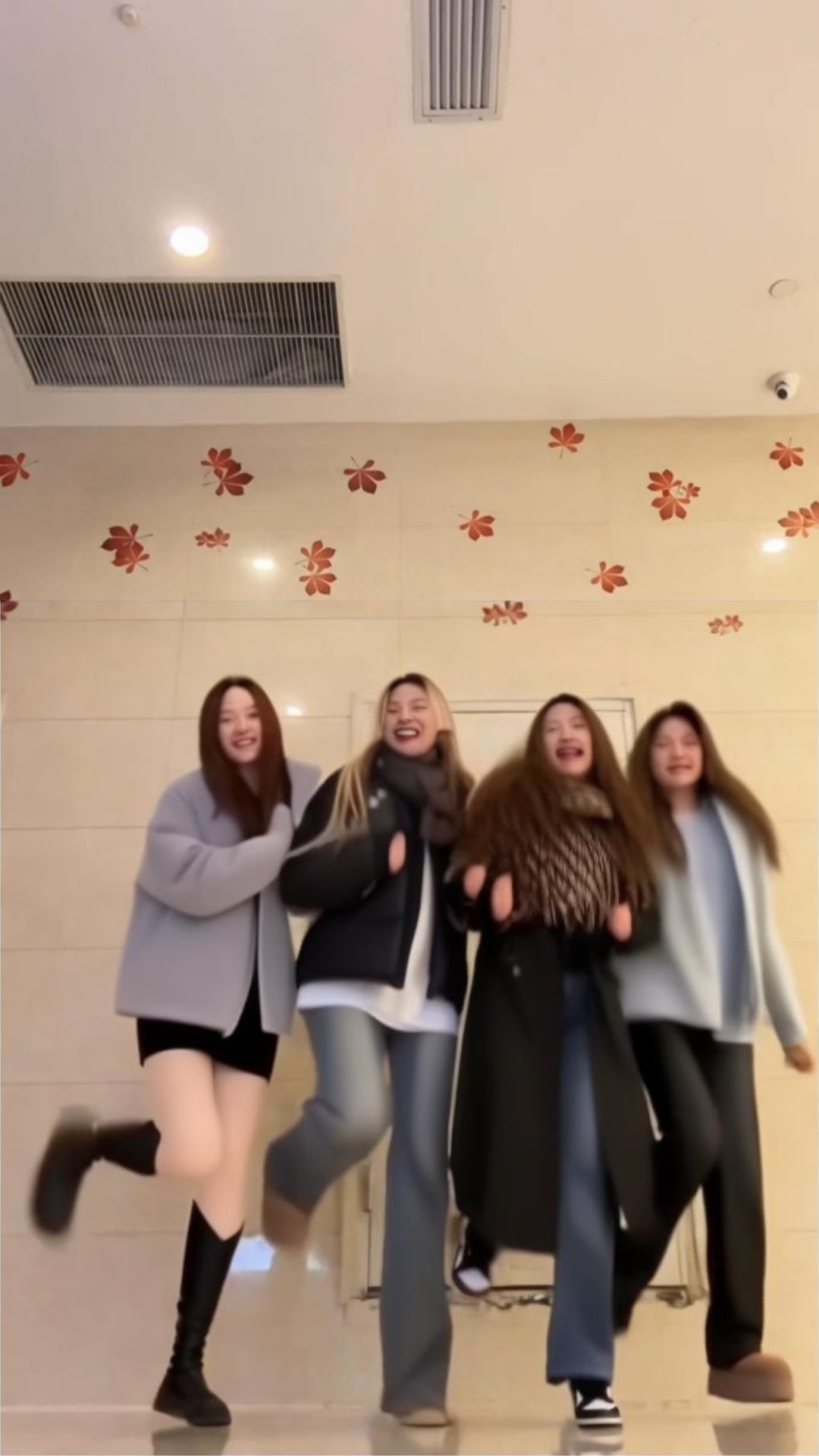}
    &\includegraphics[width=1.225cm]{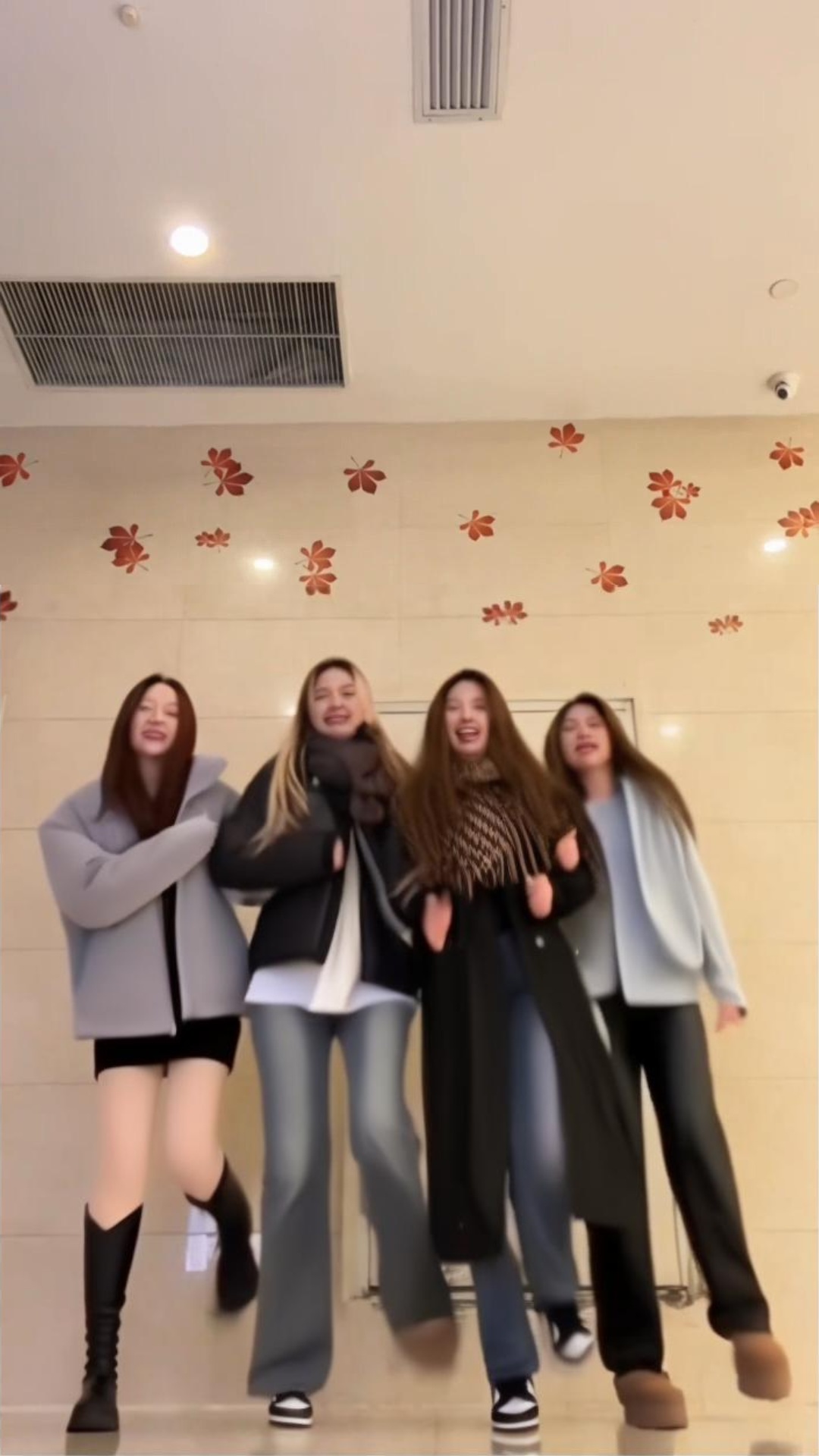}
    &\includegraphics[width=1.225cm]{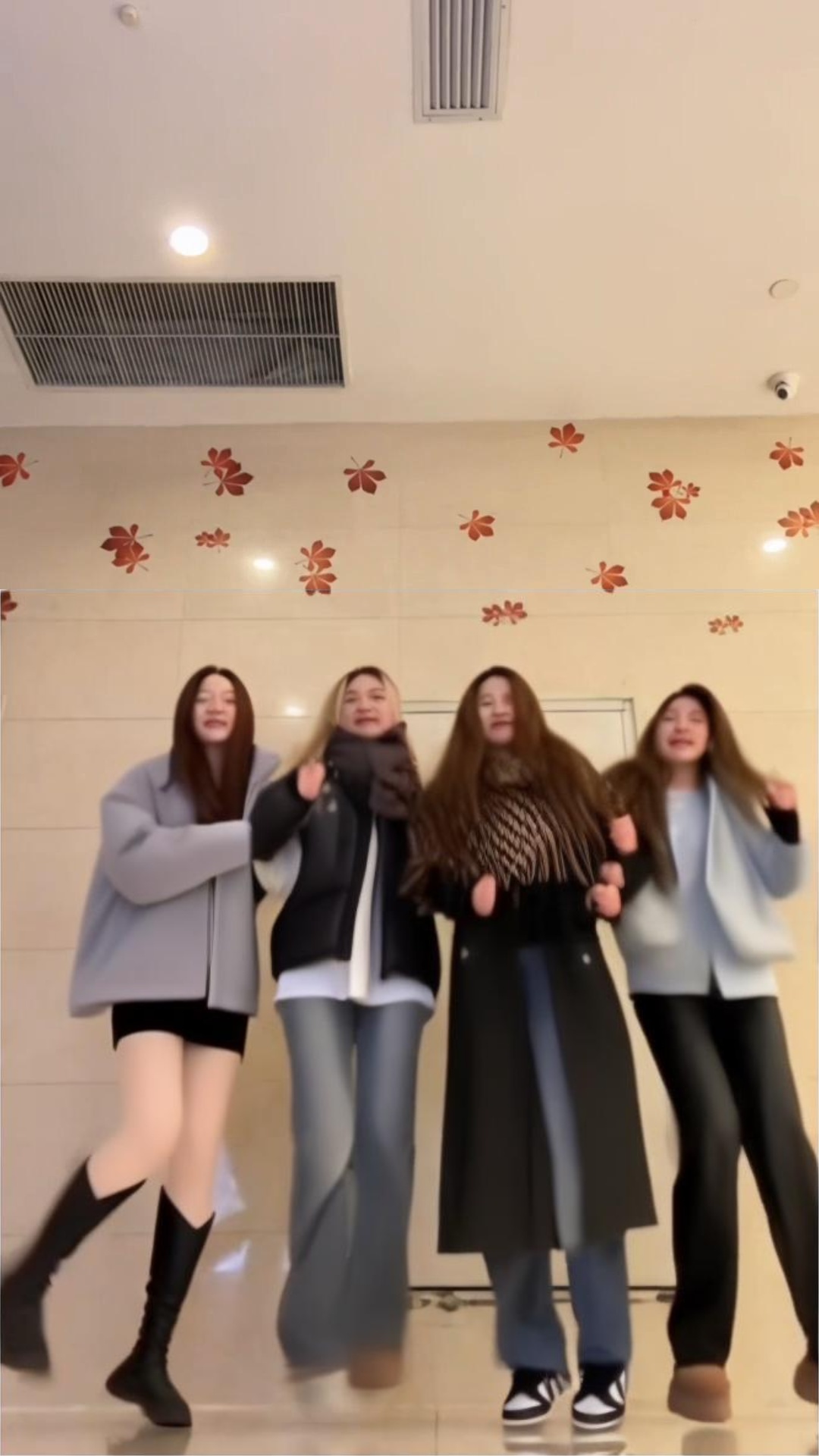}\\

    \end{tabular}
    \end{center}
    \caption{Qualitative results of more characters. MultiAnimate can generate animated results for multiple characters while preserving each character's identity and appearance.} 
    \label{fig:qual-more}
\end{figure*}

Qualitative results of multiple character animation can be seen in Figure~\ref{fig:qualitative}. While some existing methods maintain appearance consistency in initial frames, they exhibit significant degradation during complex scenarios, particularly when handling occlusions (as shown in the second row) and position interchange operations (demonstrated in the third row). In contrast, our method, benefiting from its architectural design, successfully maintains both appearance consistency and spatial relationships throughout these challenging scenarios. Moreover, in Figure~\ref{fig:qual-more}, we present qualitative results of MultiAnimate for animating multiple characters, demonstrating that MultiAnimate can handle animation tasks involving more than two characters while preserving the identity and appearance of each individual character.

\subsection{Quantitative Comparison}
\label{sec:quanti}

We evaluate performance across multiple methods using both image-level and video-level metrics. For image-level assessment, we employ SSIM~\cite{1284395}, PSNR~\cite{5596999}, LPIPS~\cite{zhang2018unreasonableeffectivenessdeepfeatures}, MSE, and FID~\cite{heusel2018ganstrainedtimescaleupdate} metrics, while utilizing FVD~\cite{unterthiner2019accurategenerativemodelsvideo} for video-level evaluation. As demonstrated in Table~\ref{tab:quan-compare}, our method achieves superior performance across all metrics compared to existing approaches.

\begin{table*}
  \caption{Quantitative comparison on multiple character animation, best results are in \textbf{bold}. Our method demonstrates significant performance advantages over comparative approaches. The MSE results are presented after being multiplied by $10^5$.}
  \label{tab:quan-compare}
  \centering
  \begin{tabular}{ccccccc}
  \hline
  & SSIM $\uparrow$ & PSNR $\uparrow$ & LPIPS $\downarrow$ & MSE $\downarrow$ & FID $\downarrow$ & FVD $\downarrow$ \\
  \hline
  Moore-Animate~\cite{hu2023animateanyone} & 0.633 & 15.24 & 0.340 & 6.02 & 80.37 & 636.95 \\
  MimicMotion~\cite{mimicmotion2024} & 0.686 & 17.26 & 0.303 & 5.92 & 110.96 & 551.78 \\
  MagicPose~\cite{chang2023magicdance} & 0.627 & 14.93 & 0.436 & 6.51 & 77.53 & 522.62 \\
  MagicAnimate~\cite{xu2023magicanimate} & 0.668 & 13.69 & 0.387 & 8.63 & 94.36 & 731.85 \\
  UniAnimate~\cite{wang2024unianimate} & 0.705 & 18.27 & 0.263 & 4.60 & 63.26 & 346.71 \\
  Ours & \textbf{0.801} & \textbf{21.71} & \textbf{0.192} & \textbf{1.58} & \textbf{32.09} & \textbf{229.06} \\
  \hline
  \end{tabular}
\end{table*}

\subsection{User Study}

To better show the superiority of our method, we conducted a comprehensive user study through a structured evaluation protocol. We recruited participants with backgrounds in computer vision research, offering compensation for their participation. The evaluation instrument comprised 20 questions addressing multiple quality metrics: visual fidelity, appearance and identity preservation relative to reference images, and temporal coherence. Participants were tasked with selecting optimal video outputs from randomized presentations of results from different methods, methodological sources were presented in randomized order to minimize selection bias. The results of the user study are shown in Table~\ref{tab:userstudy} and the results indicate that our method is significantly favored.

\begin{table}
  \caption{Result of our user study.}
  \label{tab:userstudy}
  \centering
  \begin{tabular}{cccc}
  \hline
  & Video Quality (\%) & Appearance (\%) & Temporal Consistency (\%)\\
  \hline
  Moore-Animate~\cite{hu2023animateanyone} & 0.1 & 0.1 & 0.2 \\
  MimicMotion~\cite{mimicmotion2024} & 1.6 & 1.5 & 2.0 \\
  MagicPose~\cite{chang2023magicdance} & 0.8 & 0.5 & 0.4 \\
  MagicAnimate~\cite{xu2023magicanimate} & 0.2 & 0.6 & 0.4\\
  UniAnimate~\cite{wang2024unianimate} & 1.7 & 1.8 & 2.2 \\
  Ours & \textbf{95.6} & \textbf{95.5} & \textbf{94.8} \\
  \hline
  \end{tabular}
\end{table}

\subsection{Ablation Studies}
\label{sec:ablation}

\label{sec:ab-pe}
\begin{figure}[]
    \begin{center}
    \setlength{\tabcolsep}{0.5pt}
    \begin{tabular}{m{1.75cm}<{\centering}m{1.75cm}<{\centering}m{1.75cm}<{\centering}m{1.75cm}<{\centering}m{1.75cm}<{\centering}}

    \scriptsize{Reference 1} & \scriptsize{Reference 2} & \scriptsize{Setting (a)} & \scriptsize{Setting (b)} &\scriptsize{Setting (c)} \\
    \includegraphics[width=1.7cm]{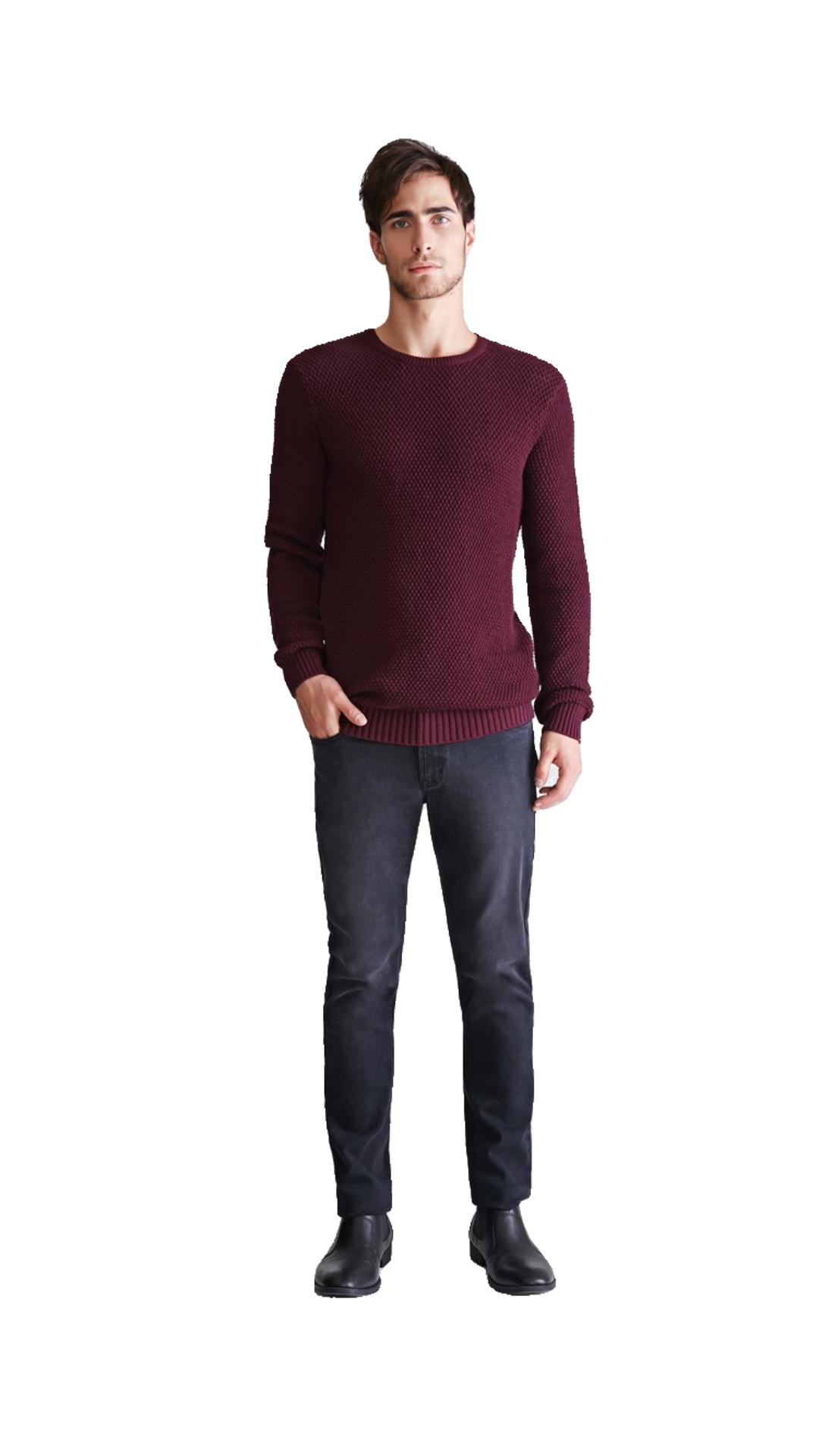}
    % &\includegraphics[width=1.5cm]{images/ablation/position/pose_0-res-1.jpg}
    &\includegraphics[width=1.7cm]{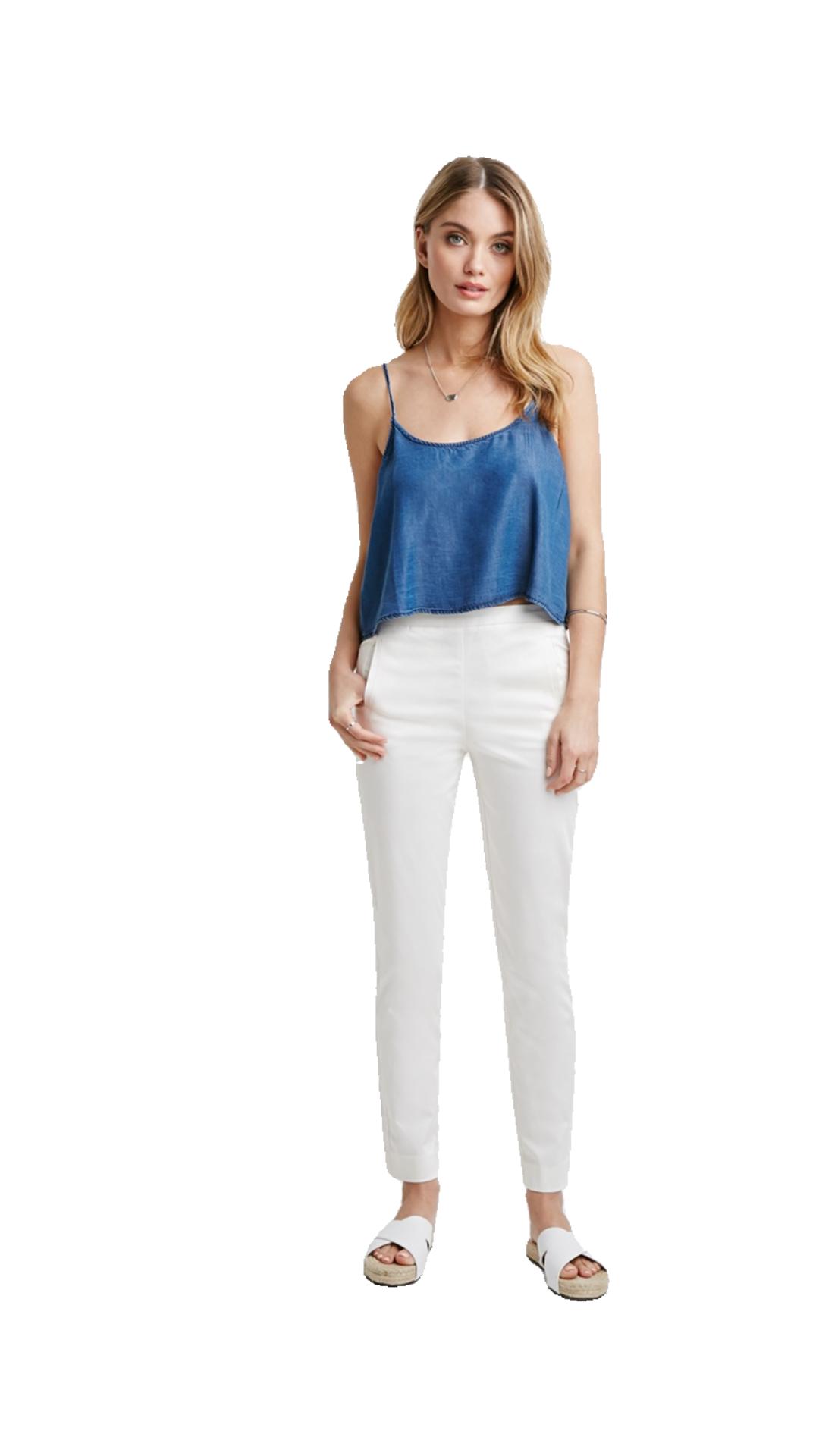}
    % &\includegraphics[width=1.5cm]{images/ablation/position/pose_1-res-1.jpg}
    &\includegraphics[width=1.7cm]{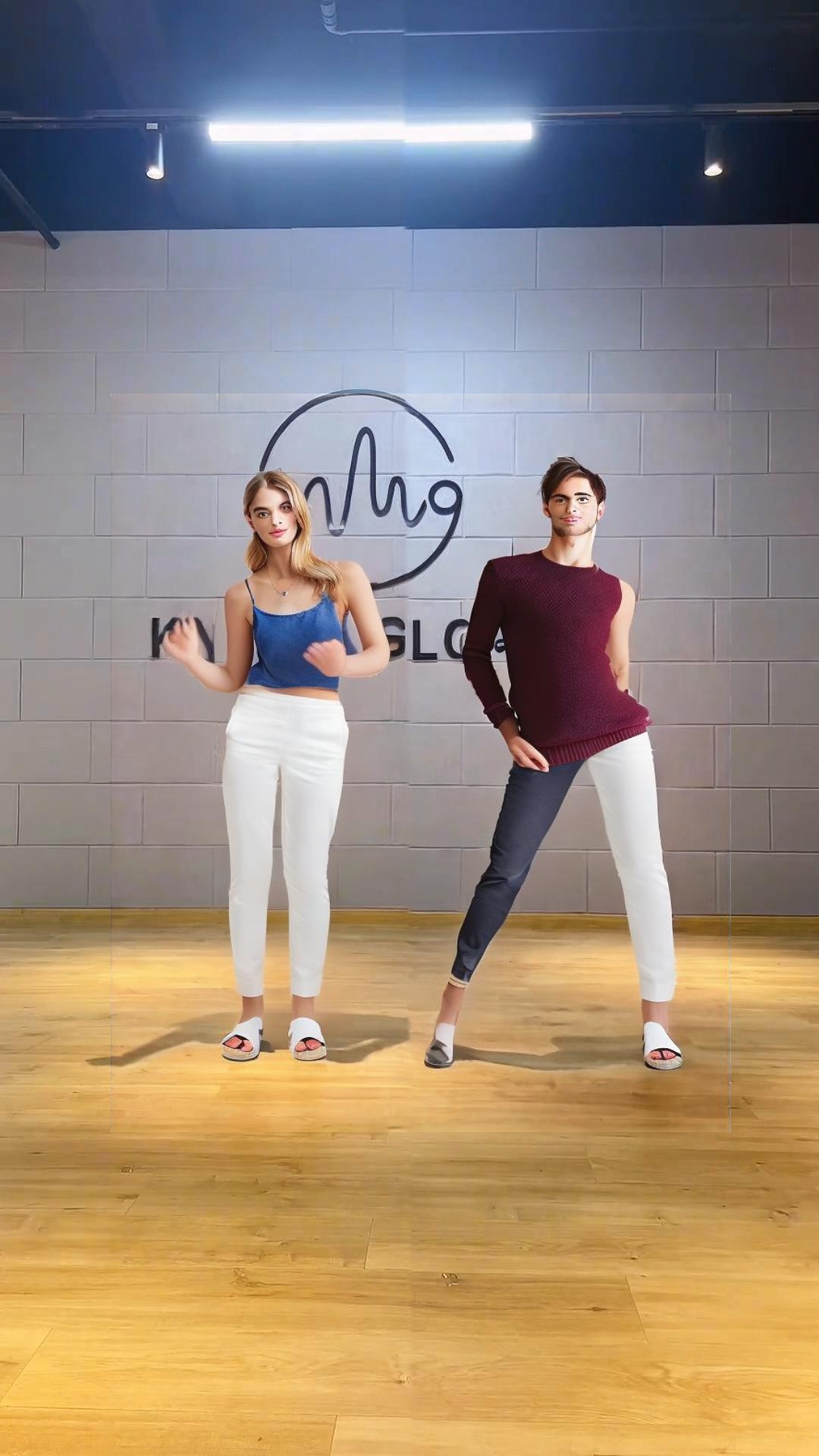}
    &\includegraphics[width=1.7cm]{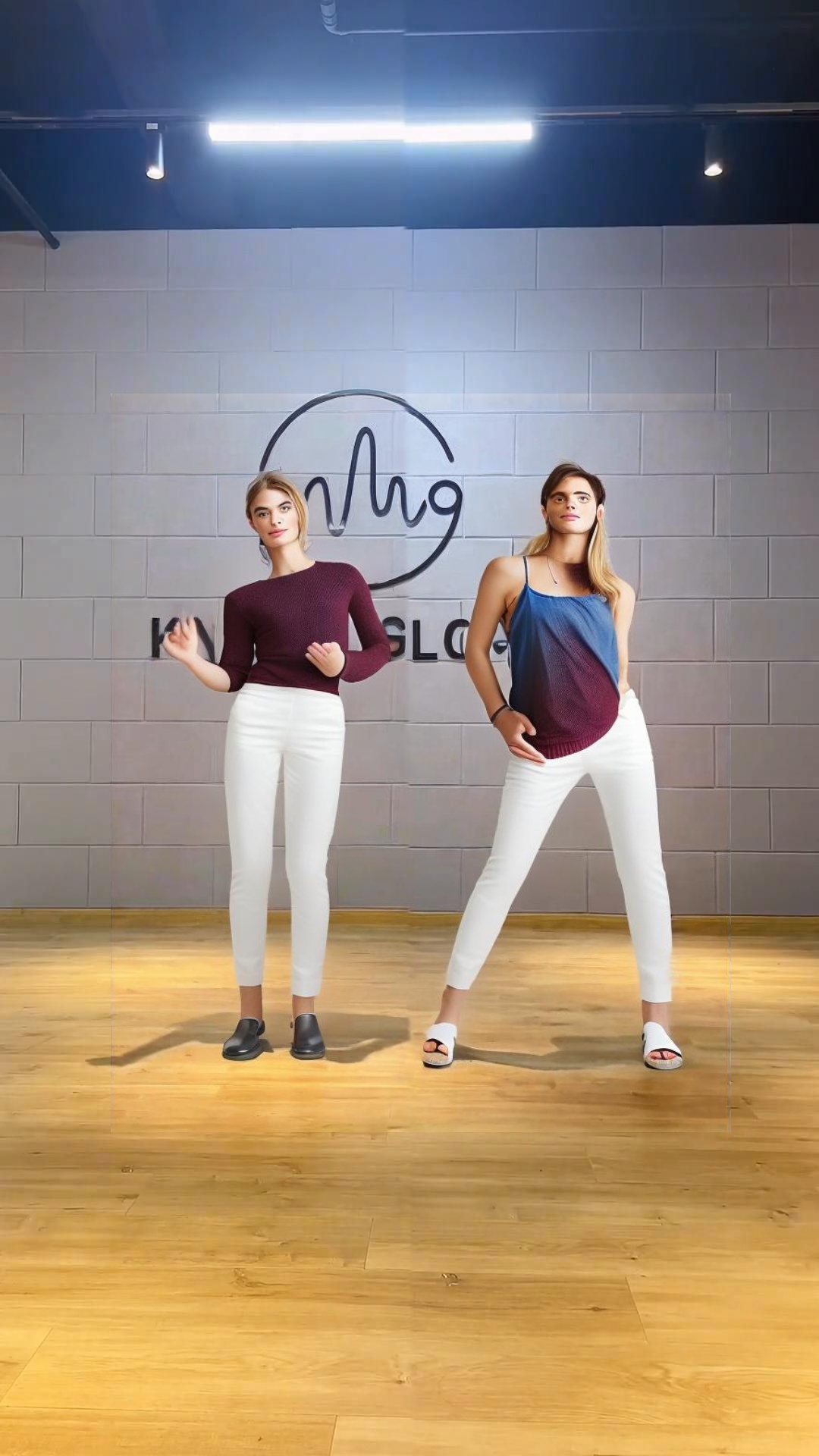}
    &\includegraphics[width=1.7cm]{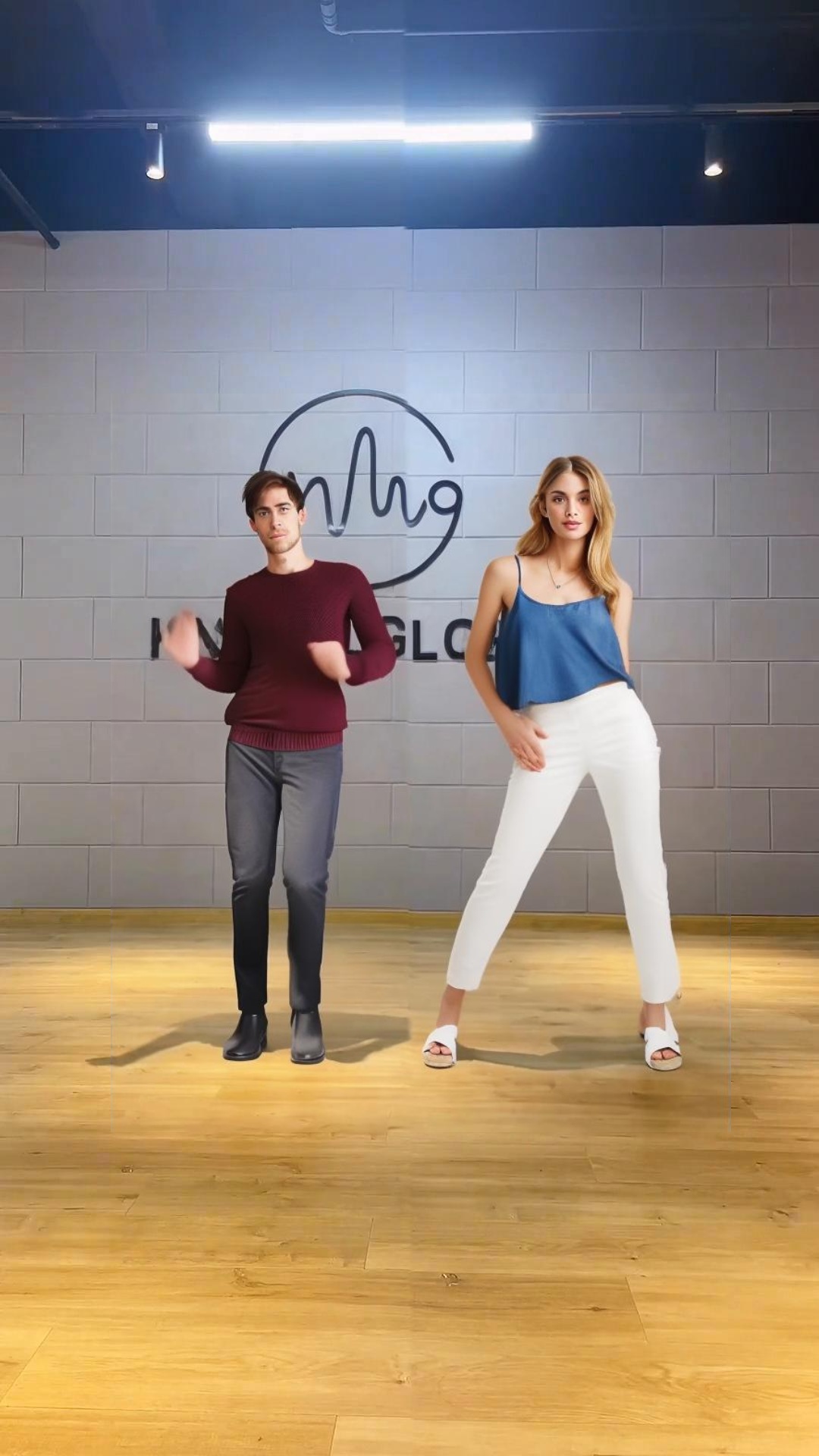} \\

    \includegraphics[width=1.7cm]{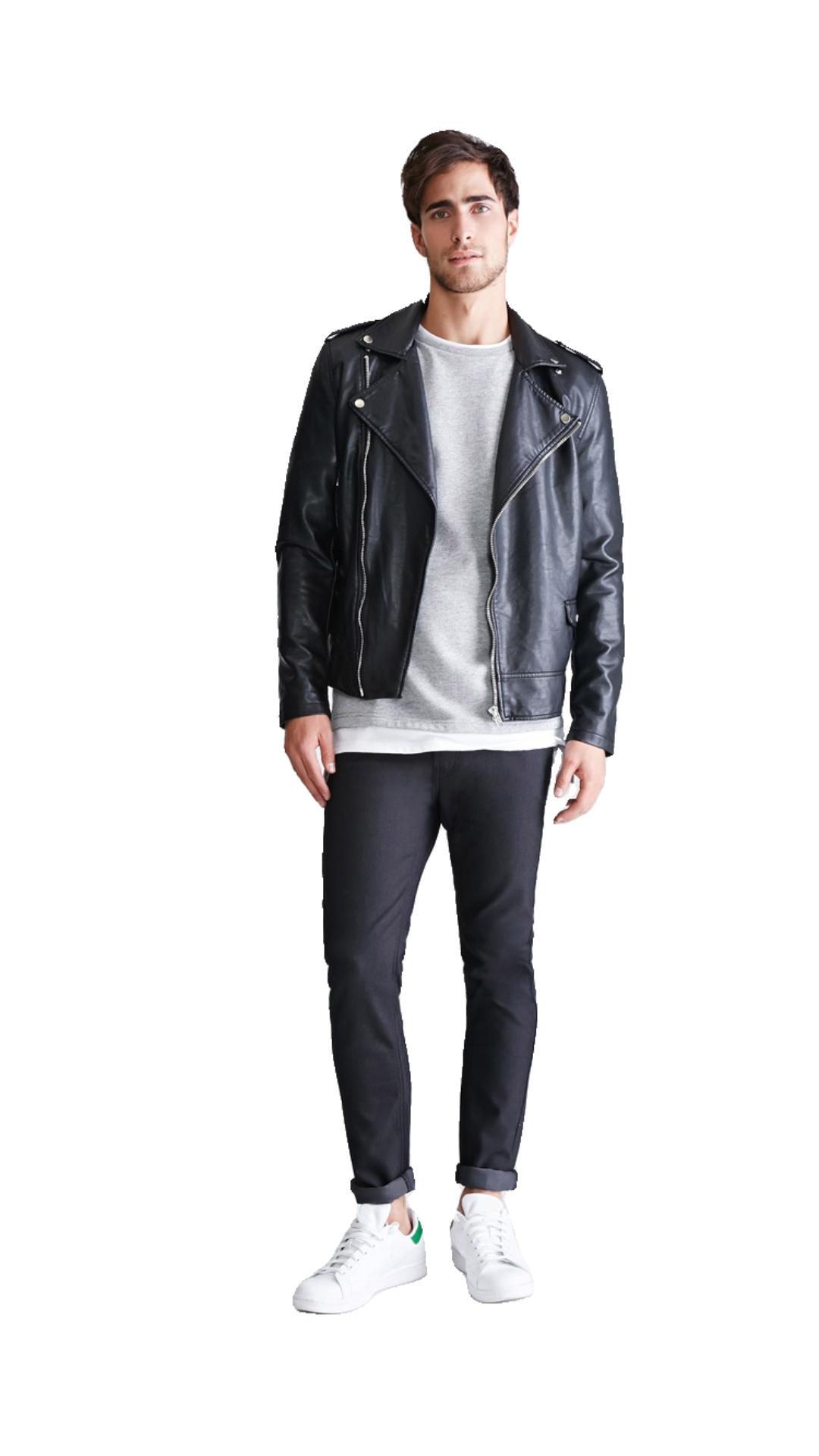}
    % &\includegraphics[width=1.5cm]{images/ablation/position/pose_0-res-2.jpg}
    &\includegraphics[width=1.7cm]{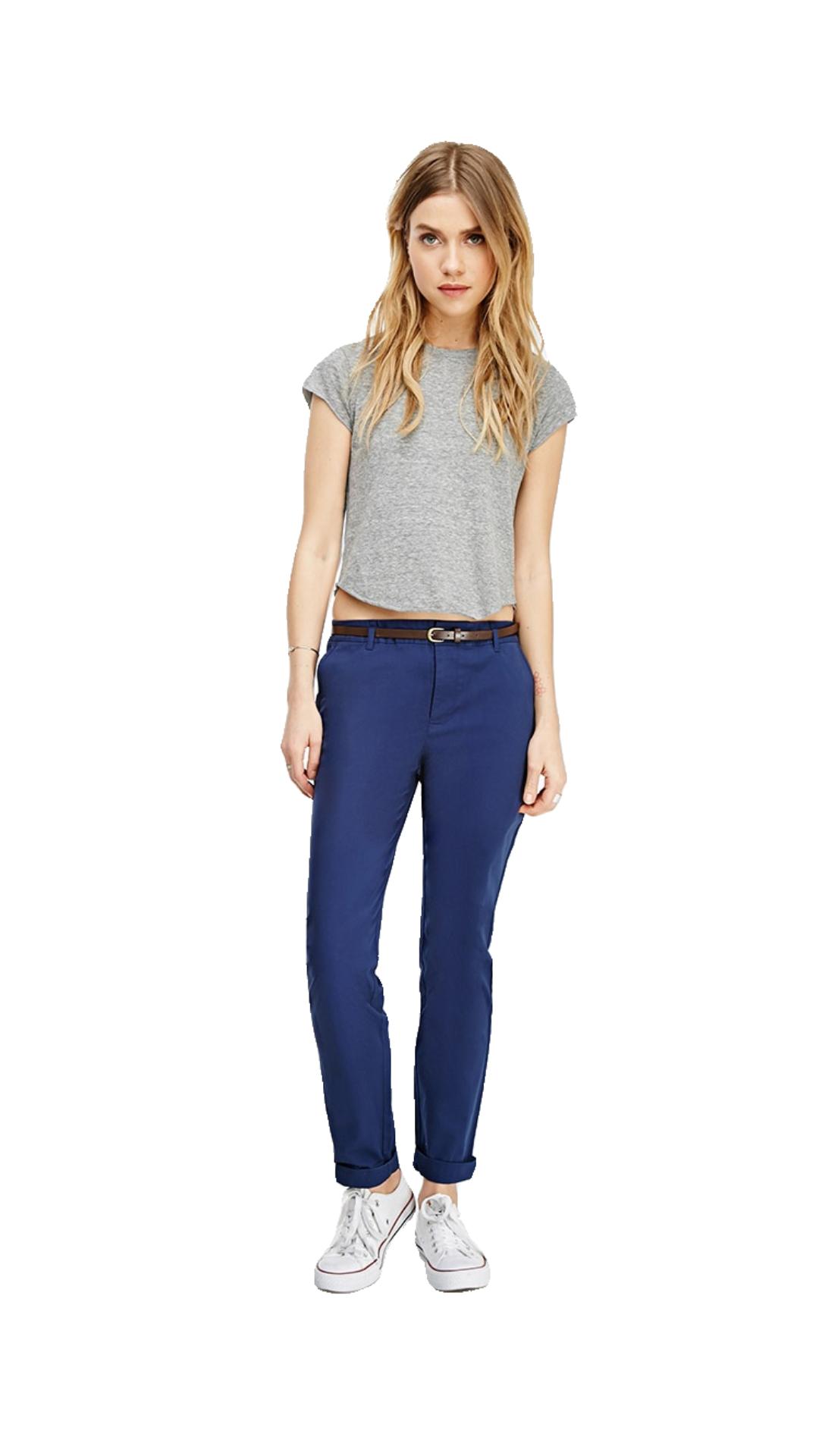}
    %&\includegraphics[width=1.5cm]{images/ablation/position/pose_1-res-2.jpg}
    &\includegraphics[width=1.7cm]{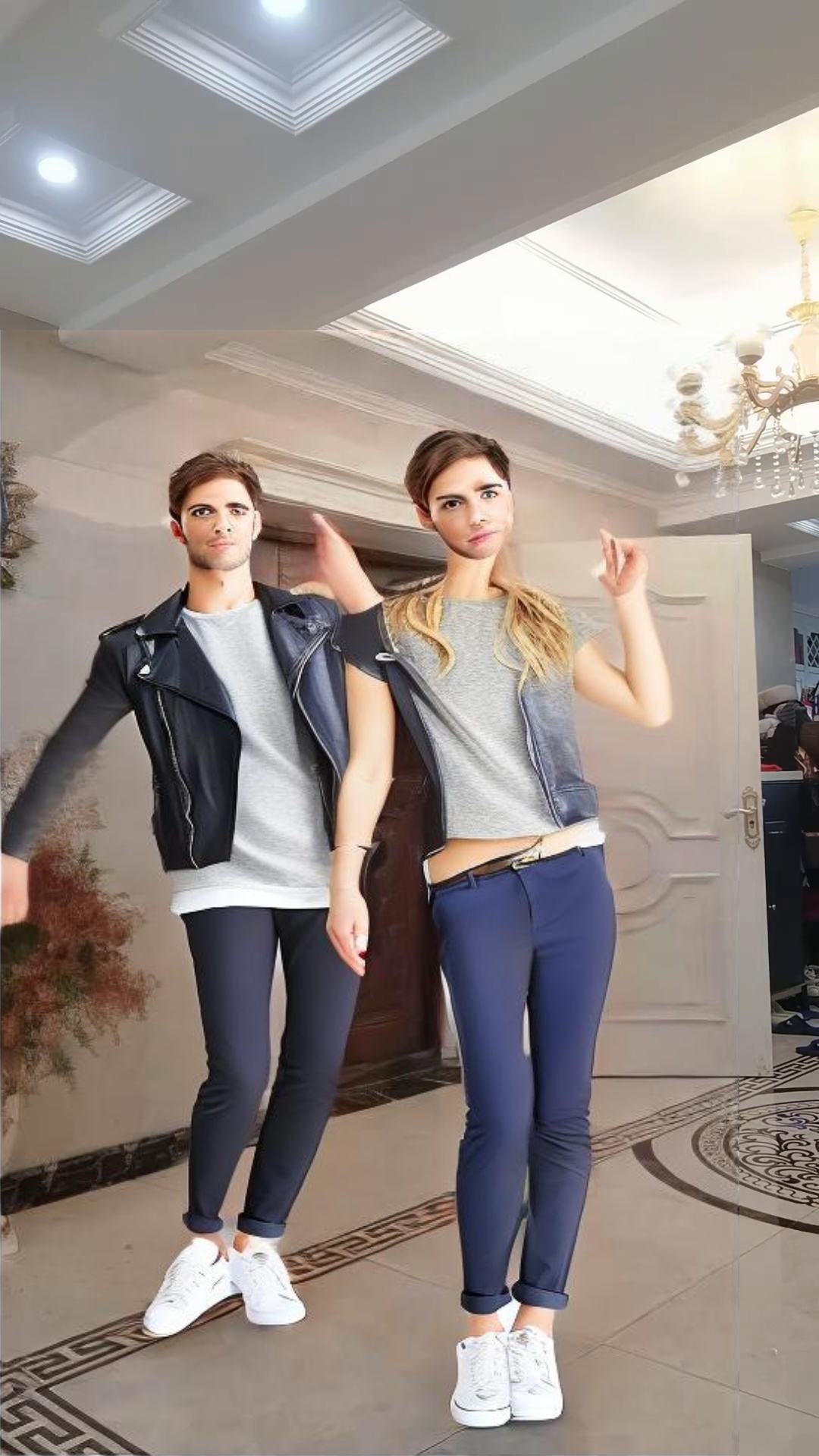}
    &\includegraphics[width=1.7cm]{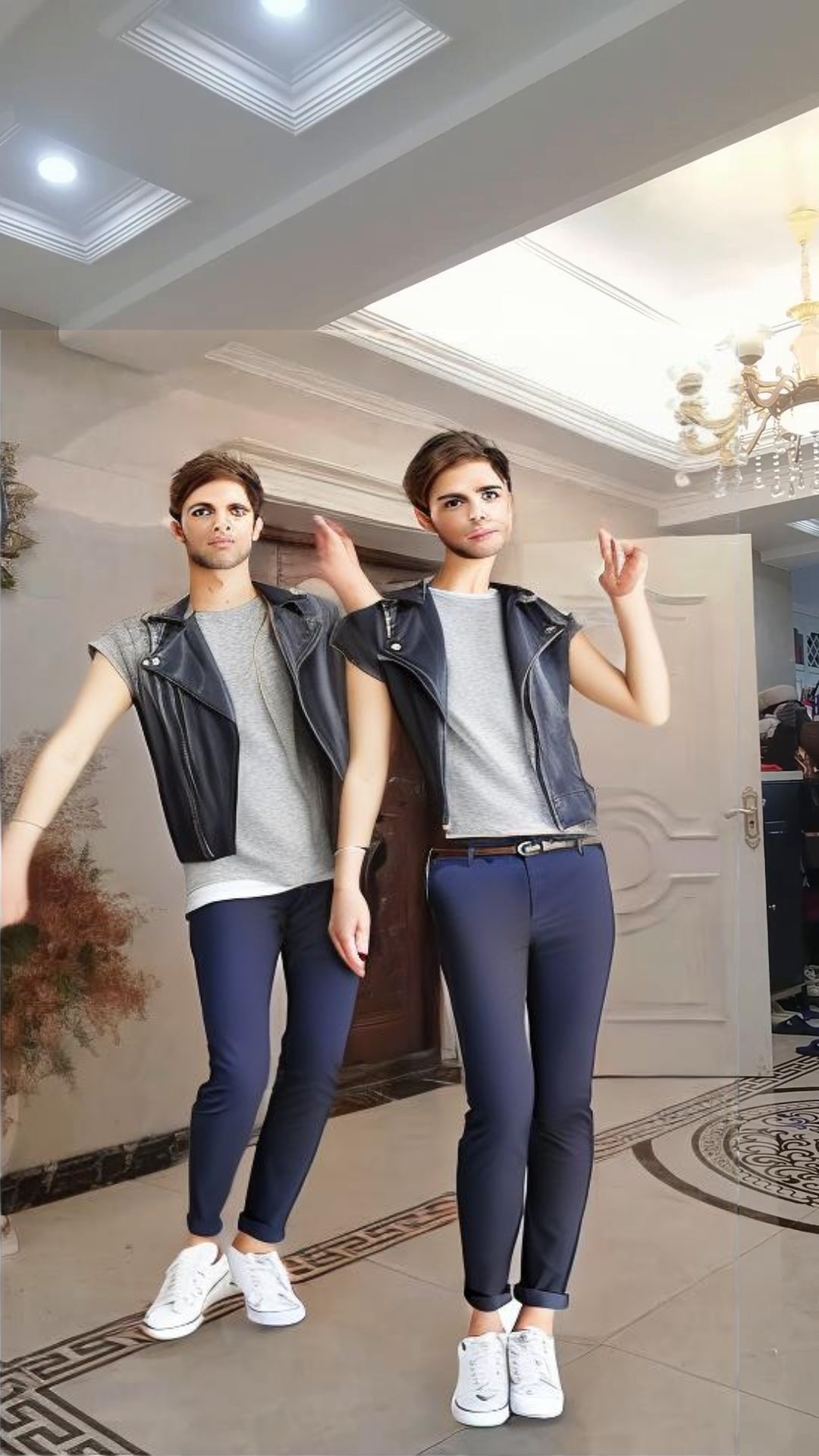}
    &\includegraphics[width=1.7cm]{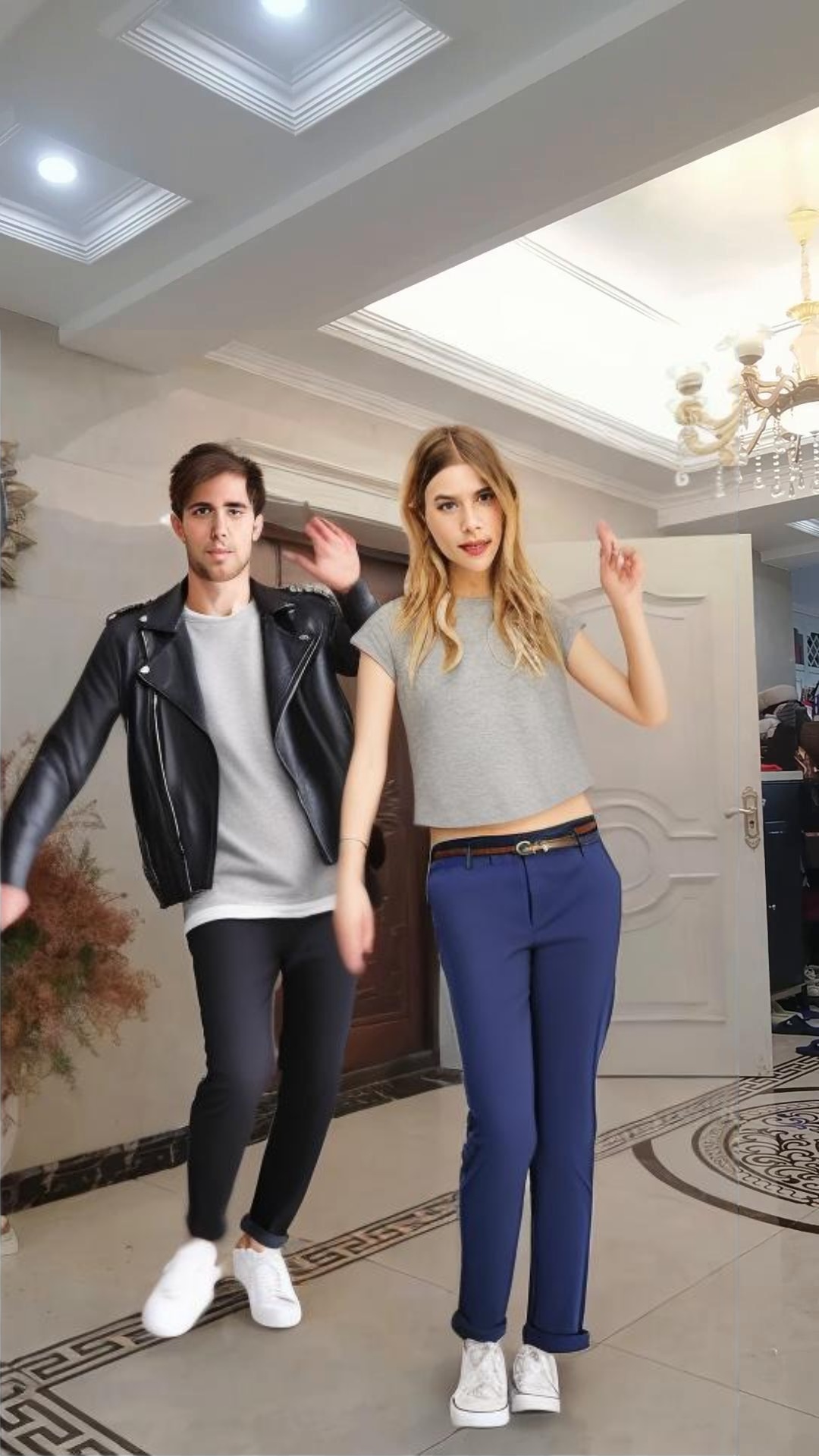} \\
    
    \end{tabular}
    \end{center}
    \caption{Ablation study on the identity-specific reference net: setting (a) and (b) exhibit identity and appearance ambiguities, whereas our proposed identity-specific reference net successfully differentiates and preserves distinct appearance characteristics from multiple reference images.} 
    \label{fig:ablation-pe}
\end{figure}

\subsubsection{Identity-Specific Reference Net.} Figure~\ref{fig:ablation-pe} demonstrates the efficacy of our identity-specific reference net through comparative analysis of three experimental configurations: (a) a baseline reference net that simply concatenate reference features, (b) a reference net with static position encoding using constant values, and (c) our proposed identity-specific reference net with learnable position encodings.

The experimental results demonstrate that both settings (a) and (b) exhibit significant identity and appearance ambiguities. Setting (a) fails due to the inherent spatial-agnostic nature of conventional reference net. Moreover, given the high dimensionality and complexity of reference features, the static position encoding using constant values in setting (b) proves insufficient for effective feature differentiation. In contrast, our proposed identity-specific reference net addresses these limitations through learnable position encoding, successfully establishing distinct feature representations from multiple reference images and enabling accurate spatial attention mapping in subsequent processing stages.

\begin{figure*}[tb]
    \begin{center}
    \setlength{\tabcolsep}{0.5pt}
    \begin{tabular}{m{1.75cm}<{\centering}m{1.75cm}<{\centering}m{0.2cm}<{\centering}m{1.75cm}<{\centering}m{1.75cm}<{\centering}m{0.2cm}<{\centering}m{1.75cm}<{\centering}m{1.75cm}<{\centering}}

    \scriptsize{Reference 1} & \scriptsize{Reference 2} &  & \multicolumn{2}{c}{\scriptsize{w/o. our pose encoder}} &  & \multicolumn{2}{c}{\scriptsize{w. our pose encoder}}\\
    \includegraphics[width=1.7cm]{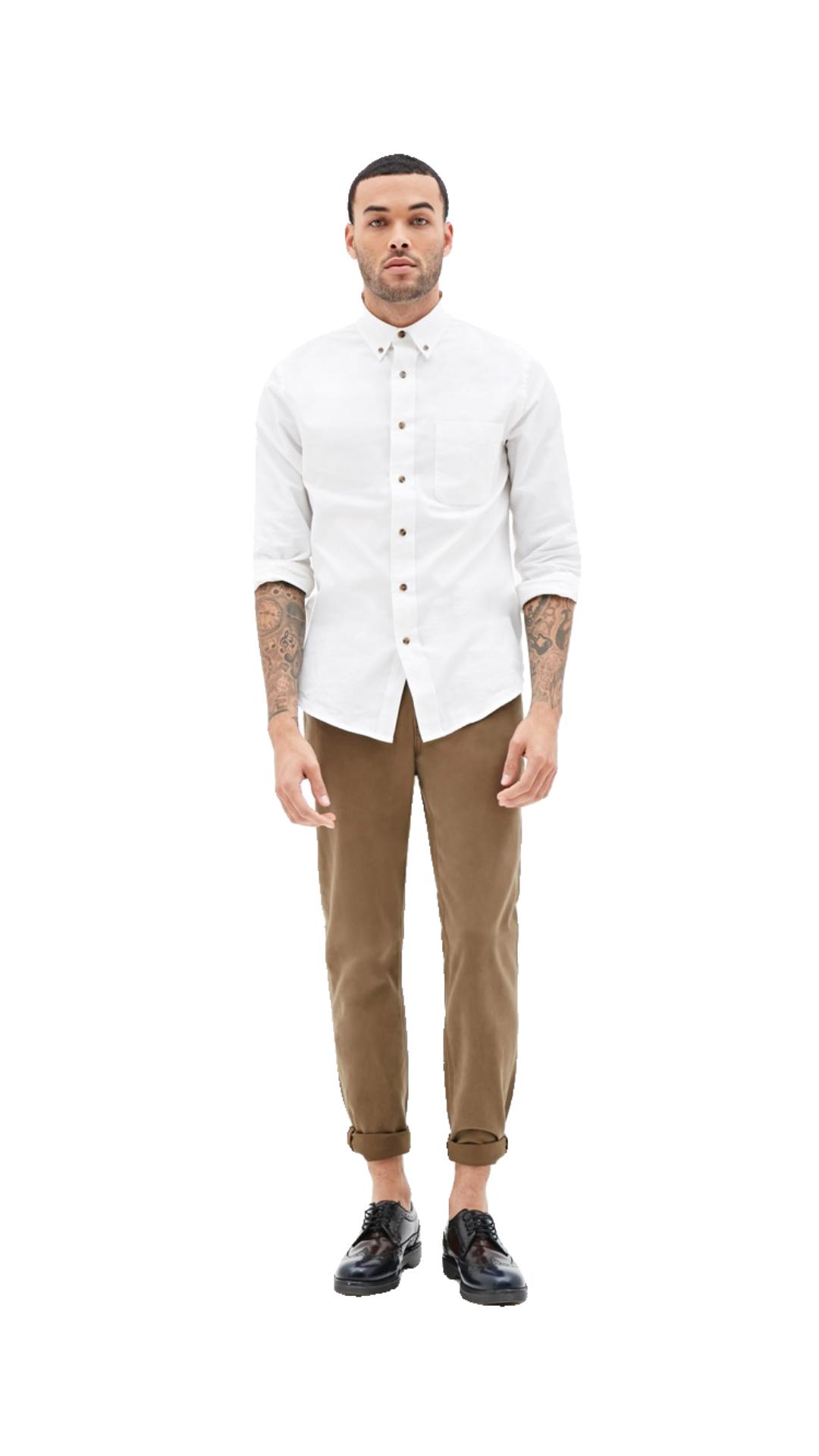}
    &\includegraphics[width=1.7cm]{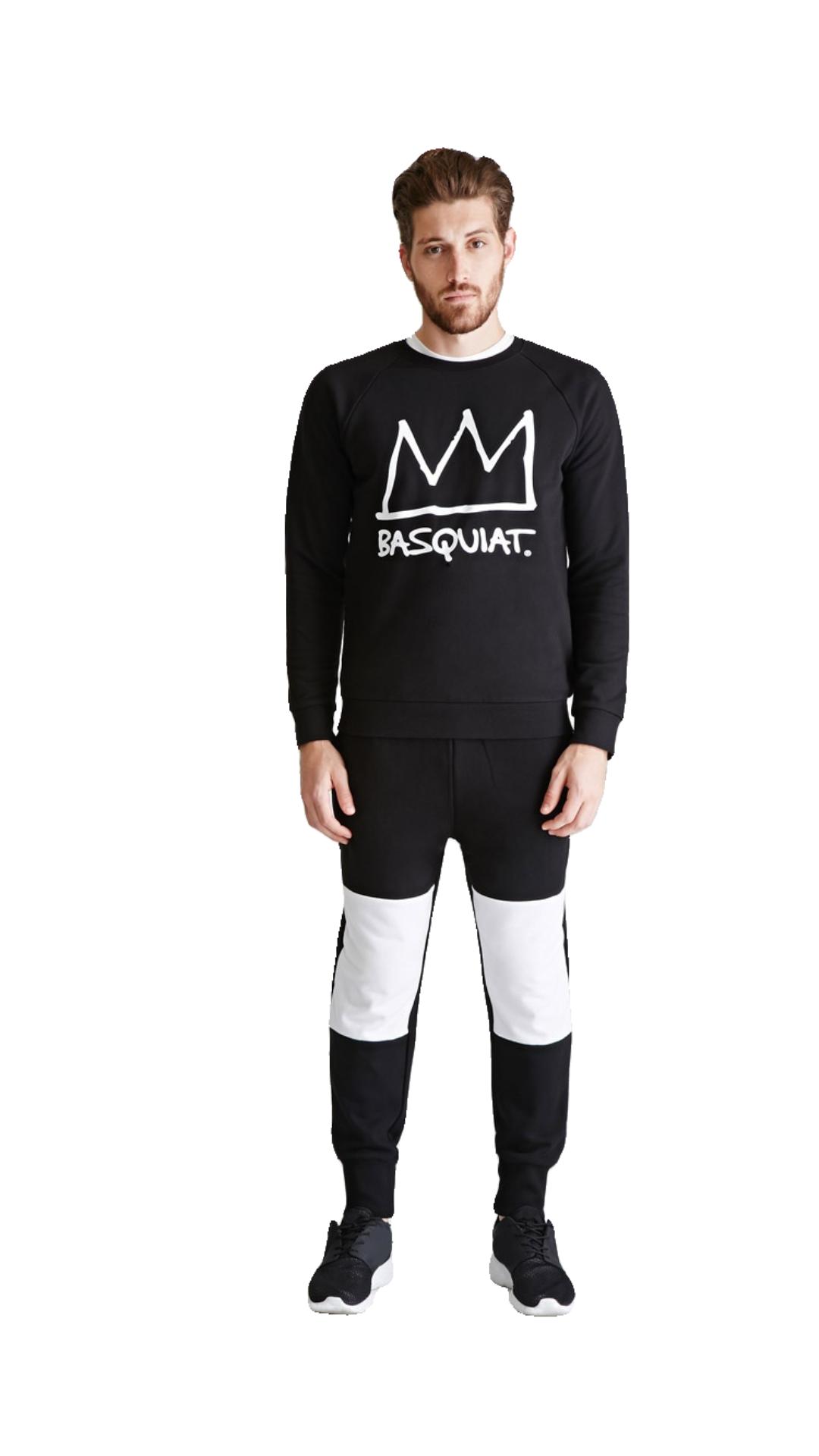}
    &
    &\includegraphics[width=1.7cm]{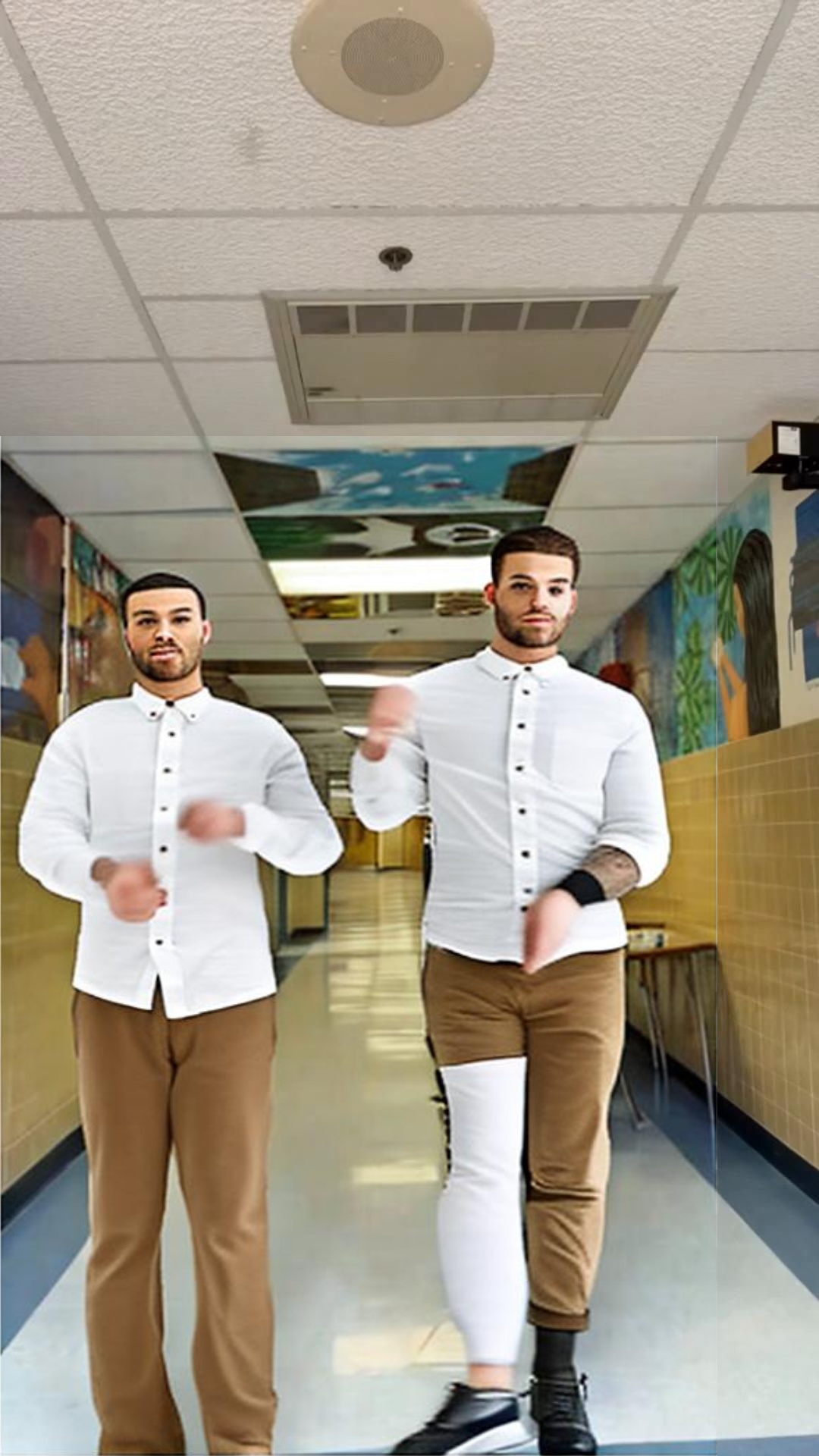}
    &\includegraphics[width=1.7cm]{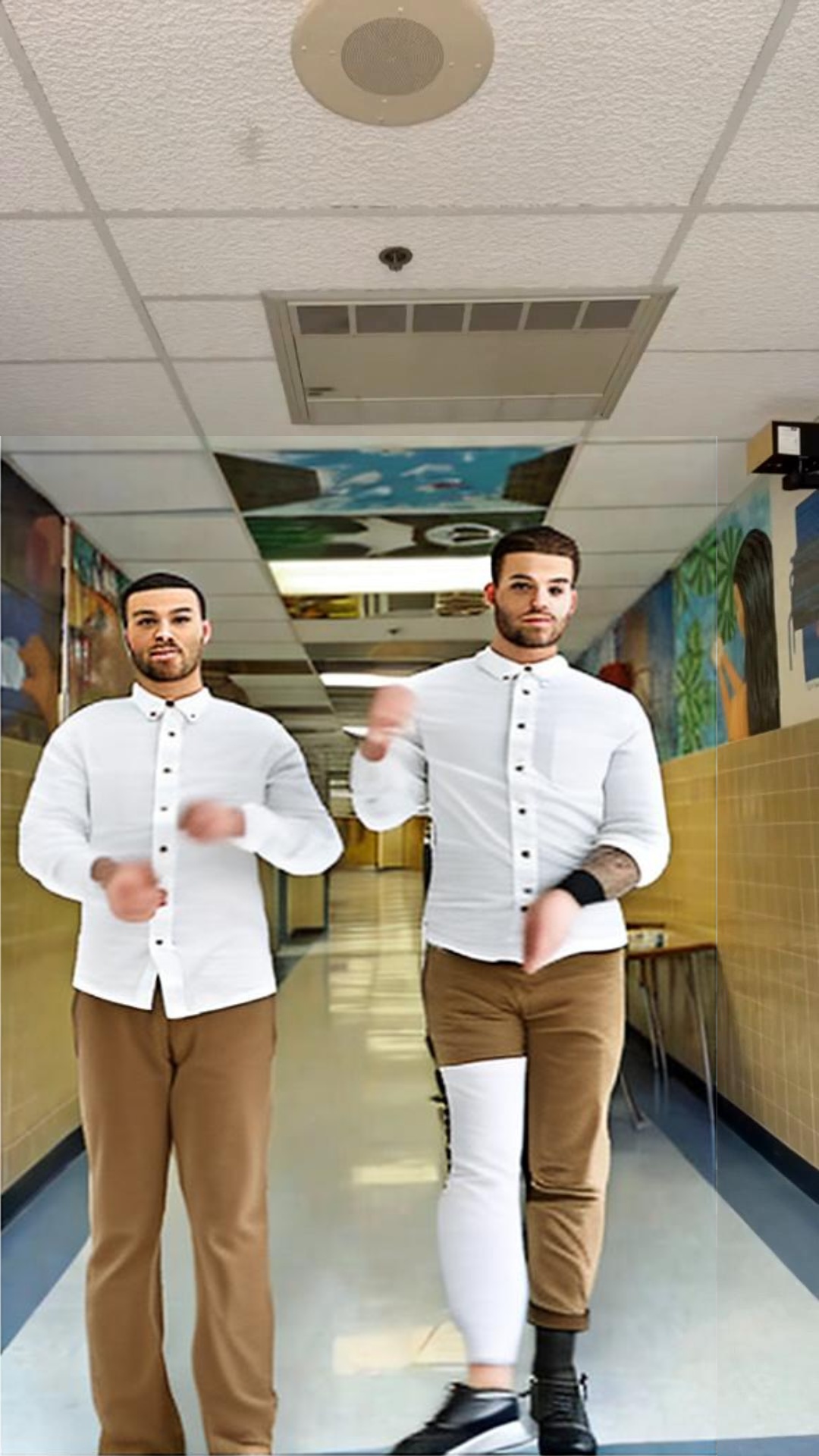}
    &
    &\includegraphics[width=1.7cm]{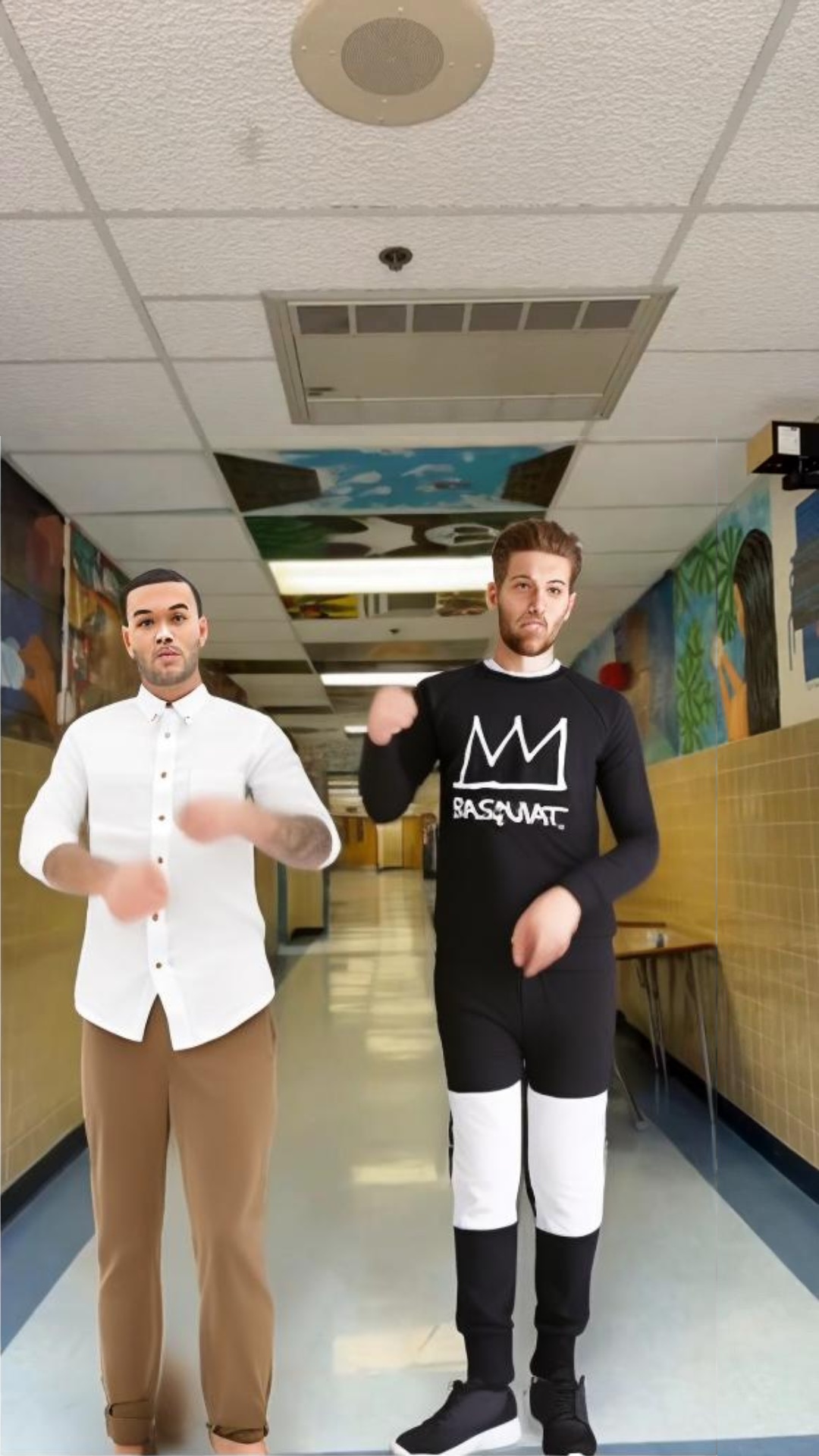}
    &\includegraphics[width=1.7cm]{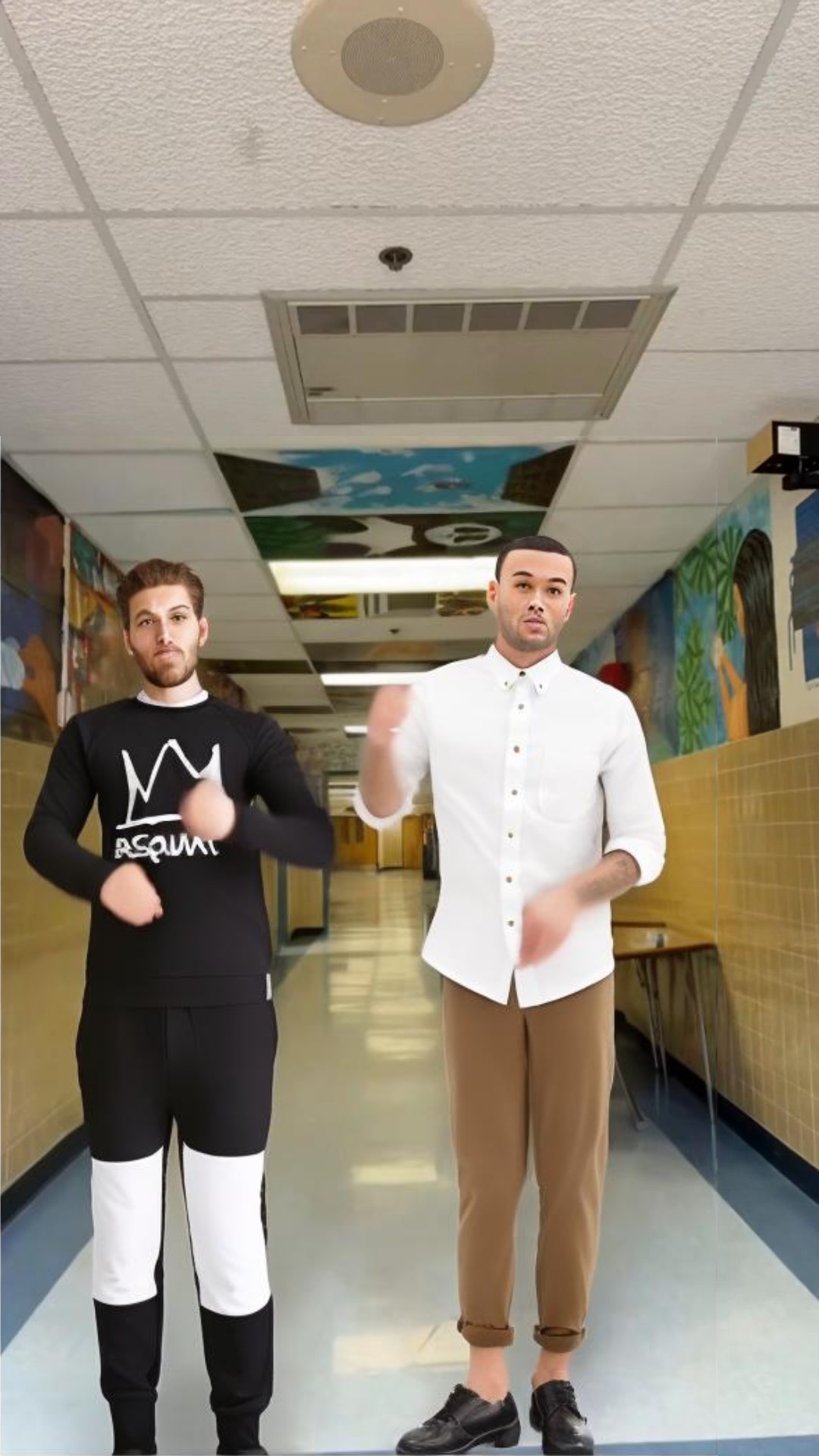} \\

    \includegraphics[width=1.7cm]{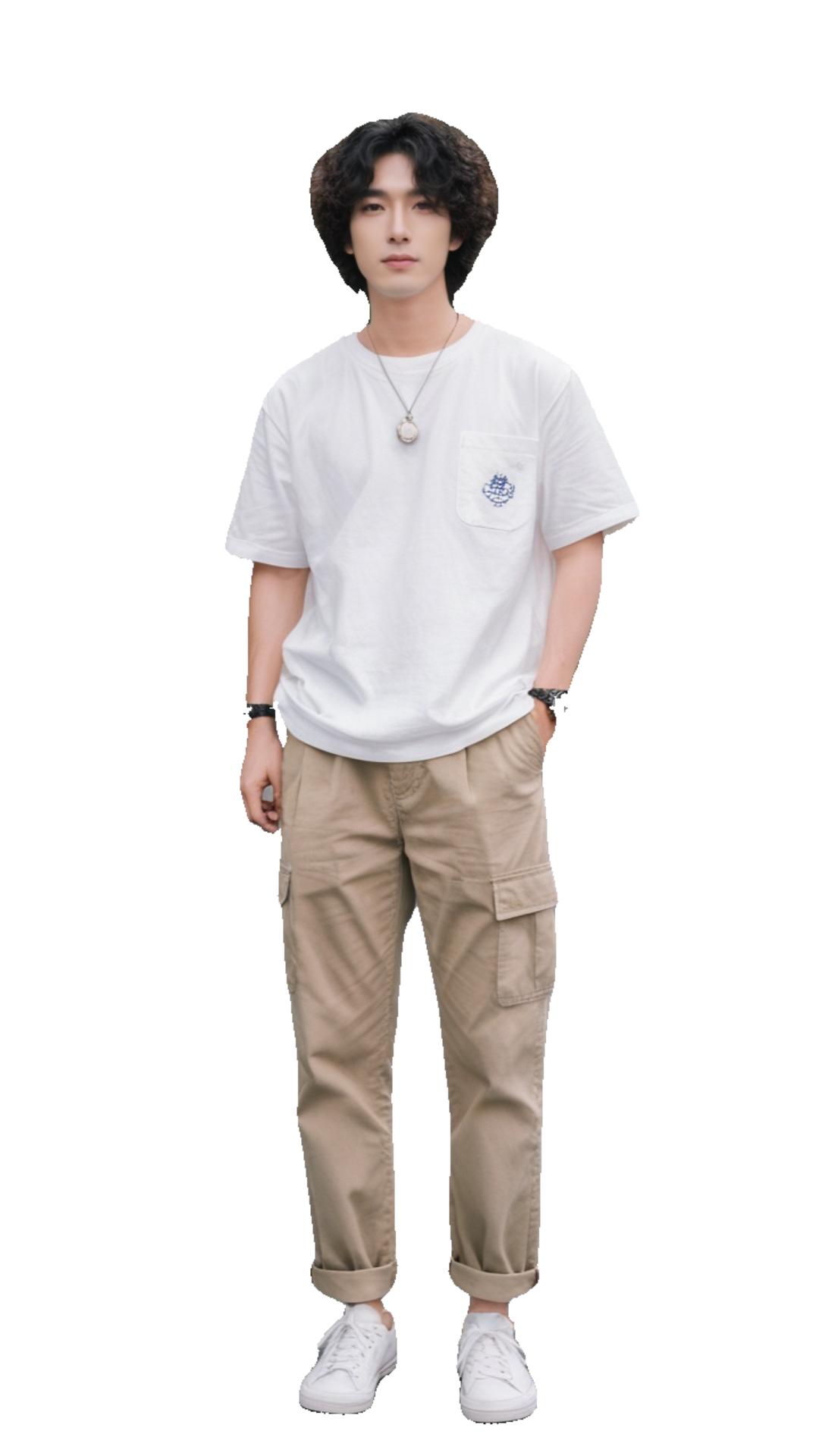}
    &\includegraphics[width=1.7cm]{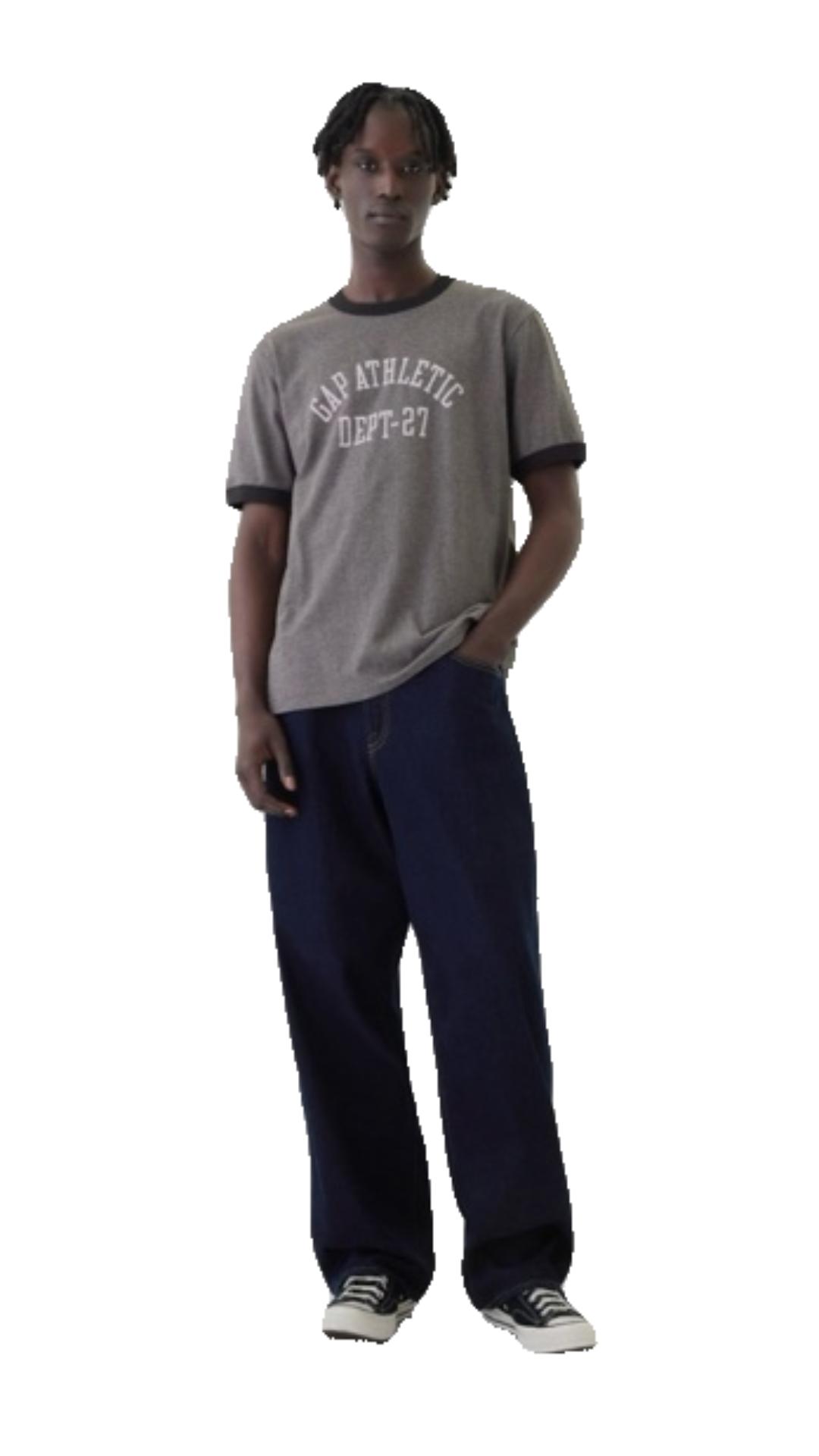}
    &
    &\includegraphics[width=1.7cm]{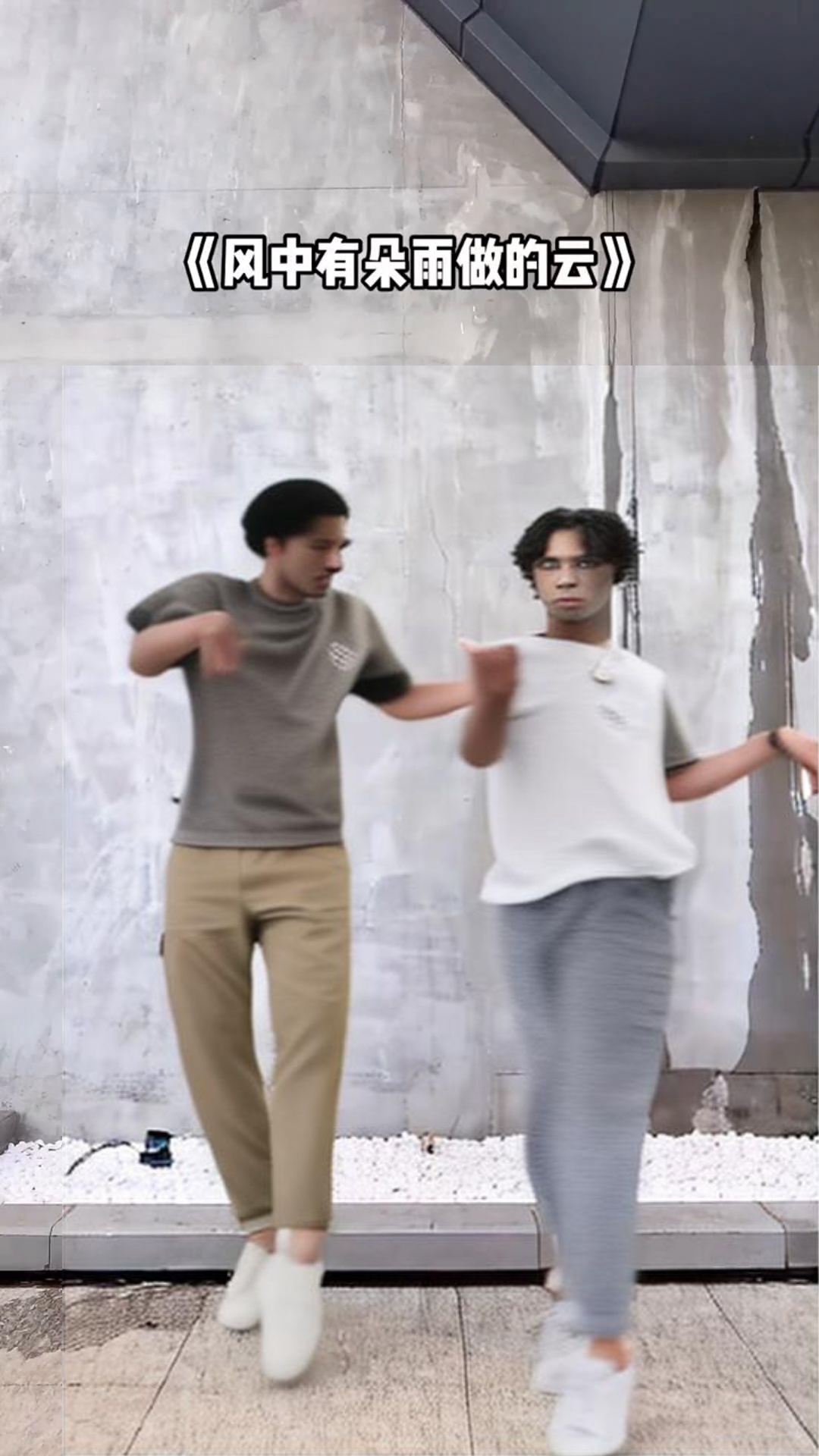}
    &\includegraphics[width=1.7cm]{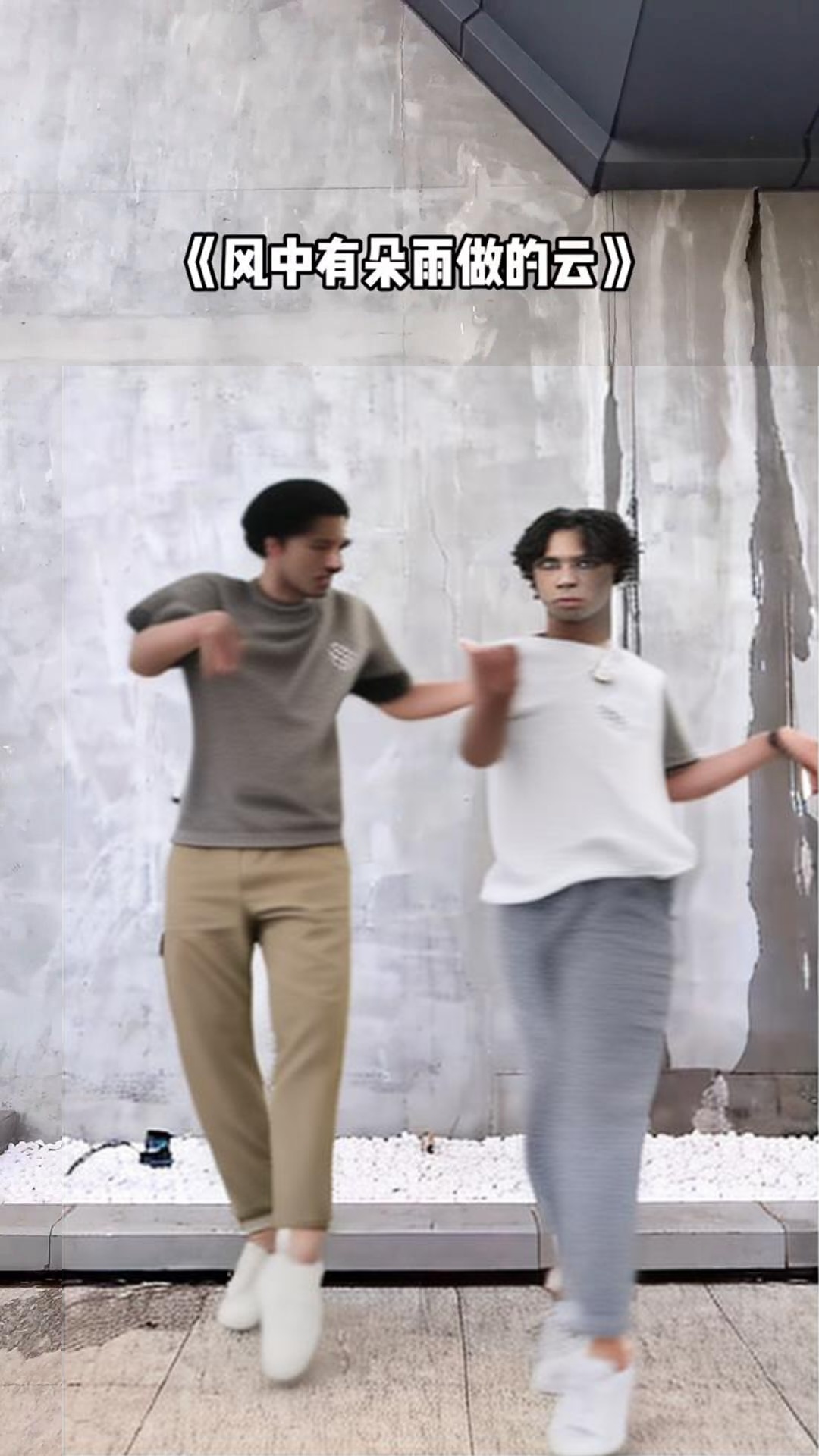}
    &
    &\includegraphics[width=1.7cm]{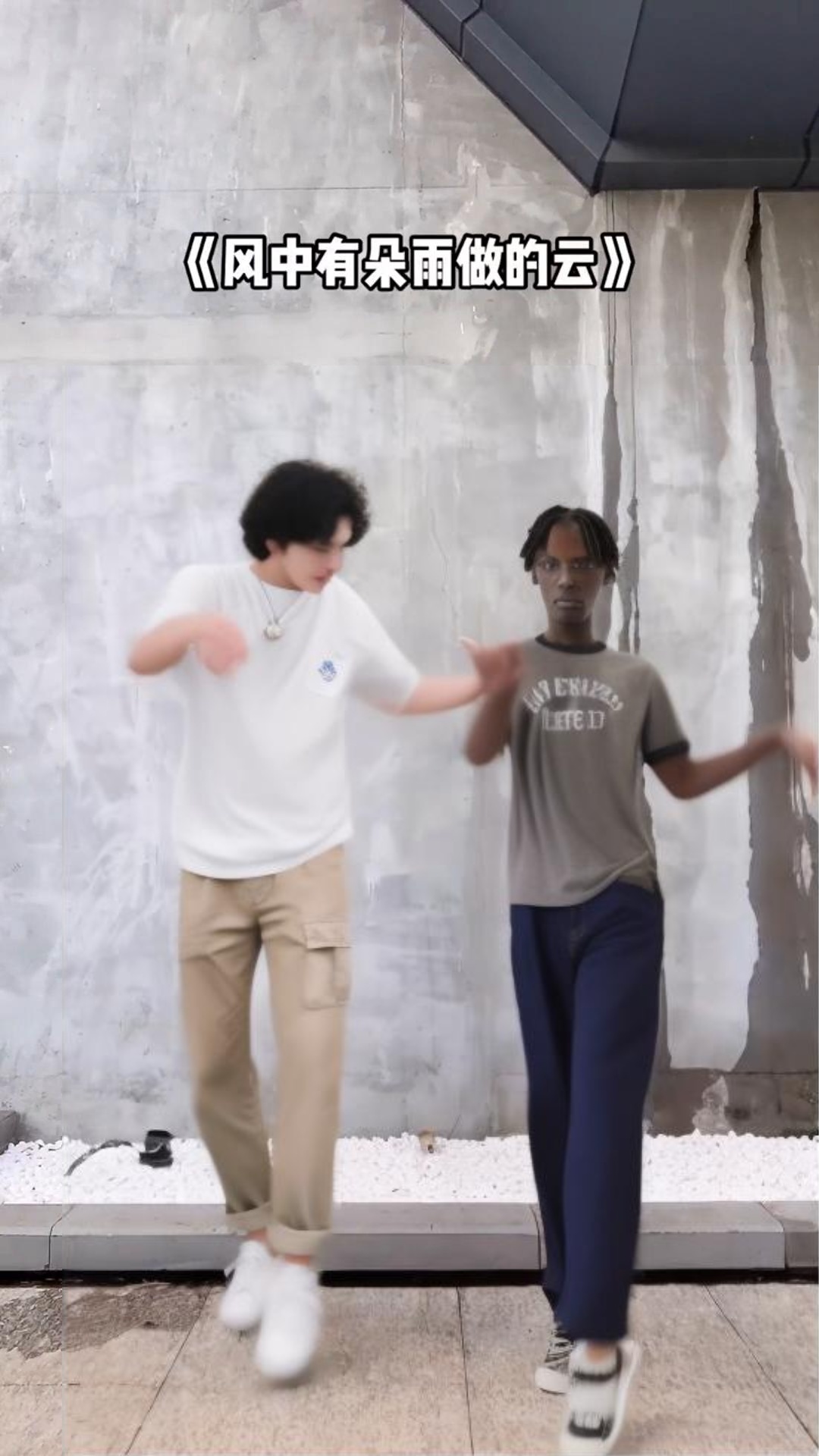}
    &\includegraphics[width=1.7cm]{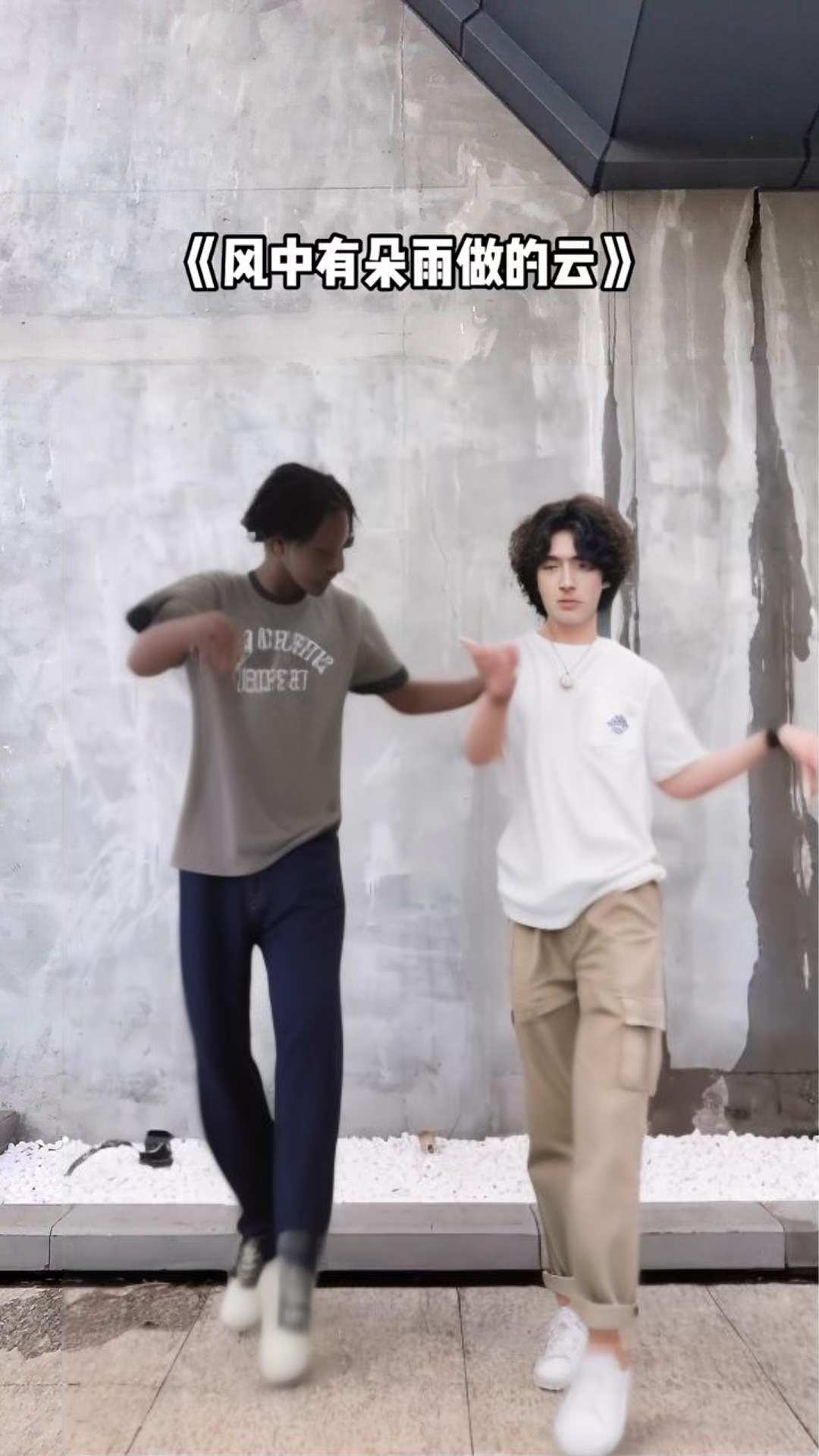} \\
    
    \end{tabular}
    \end{center}
    \caption{Ablation study on our identity-aware pose encoder. Our framework demonstrates robust identity preservation when interchanging pose sequences between reference images, while maintaining distinct appearance characteristics without degradation or ambiguity.} 
    \label{fig:ablation-id}
\end{figure*}

In our approach, we add learnable position encoding to each layer of the UNet, drawing inspiration from state-of-the-art image generation models (e.g., Flux~\cite{labs2025flux1kontextflowmatching}) and video generation models (e.g., Wan~\cite{wan2025}), which all incorporate PE into every layer of their models. Additionally, FreeFlux~\cite{wei2025freefluxunderstandingexploitinglayerspecific} demonstrates that existing SOTA models such as Flux use Rotary Position Encoding (RoPE) to achieve better performance than those that only inject PE in the first layer. We further illustrate this through experiments. When we only add this PE to the first layer of UNet, the performance of our framework degrades. The quantitative experimental results are shown in Table~\ref{tab:quan-pe}.

\begin{table*}
  \caption{Quantitative comparison on different position encoding designs. The one layer position encoding demonstrates performance degradation.}
  \label{tab:quan-pe}
  \centering
  \begin{tabular}{cccccc}
  \hline
  & SSIM $\uparrow$ & PSNR $\uparrow$ & LPIPS $\downarrow$ & MSE ($\times10^5$) $\downarrow$ & FID $\downarrow$ \\
  \hline
  One-layer PE & 0.787 & 21.20 & 0.201 & 1.63 & 34.45 \\
  Ours & 0.801 & 21.71 & 0.192 & 1.58 & 32.09 \\

  \hline
  \end{tabular}
\end{table*}

\begin{figure*}[h]
    \begin{center}
    \setlength{\tabcolsep}{0.5pt}
    \begin{tabular}{m{0.4cm}<{\centering}m{1.53cm}<{\centering}m{1.53cm}<{\centering}m{1.53cm}<{\centering}m{1.53cm}<{\centering}m{1.53cm}<{\centering}m{1.53cm}<{\centering}m{1.53cm}<{\centering}m{1.53cm}<{\centering}}

    \raisebox{0.3cm}{\rotatebox[origin=c]{90}{\footnotesize{{Ours}}}}
    &\includegraphics[width=1.5cm]{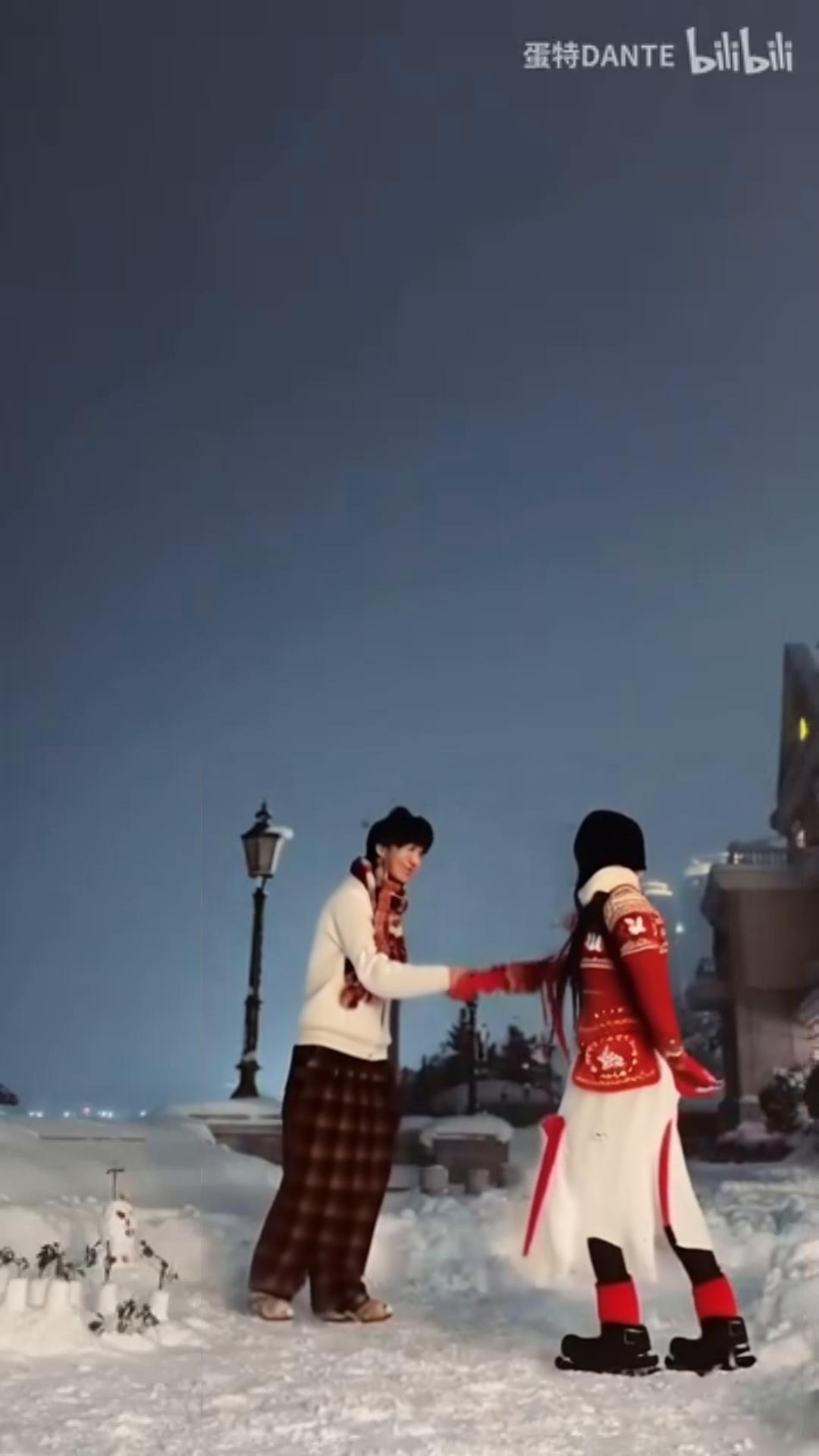}
    &\includegraphics[width=1.5cm]{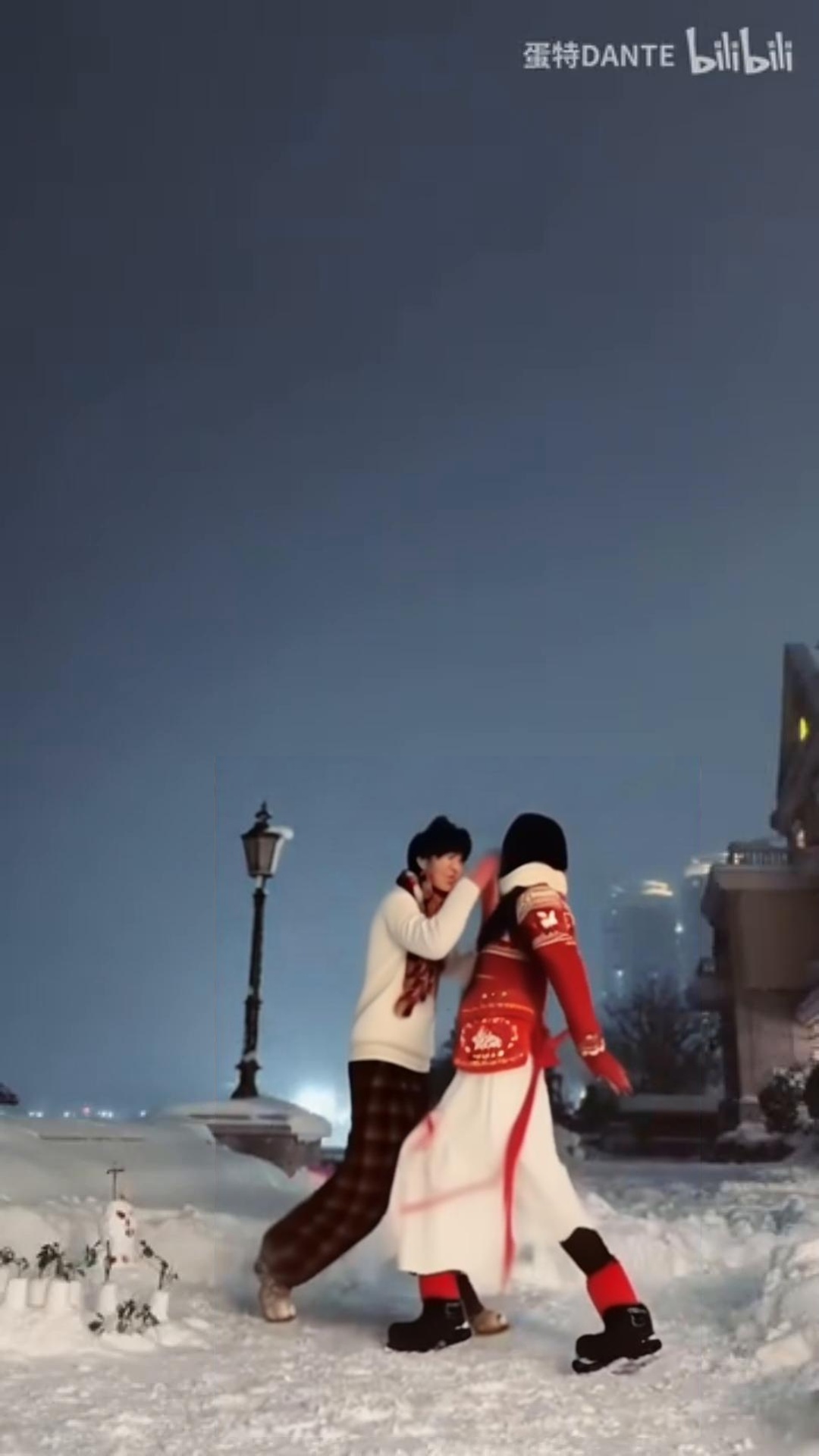}
    &\includegraphics[width=1.5cm]{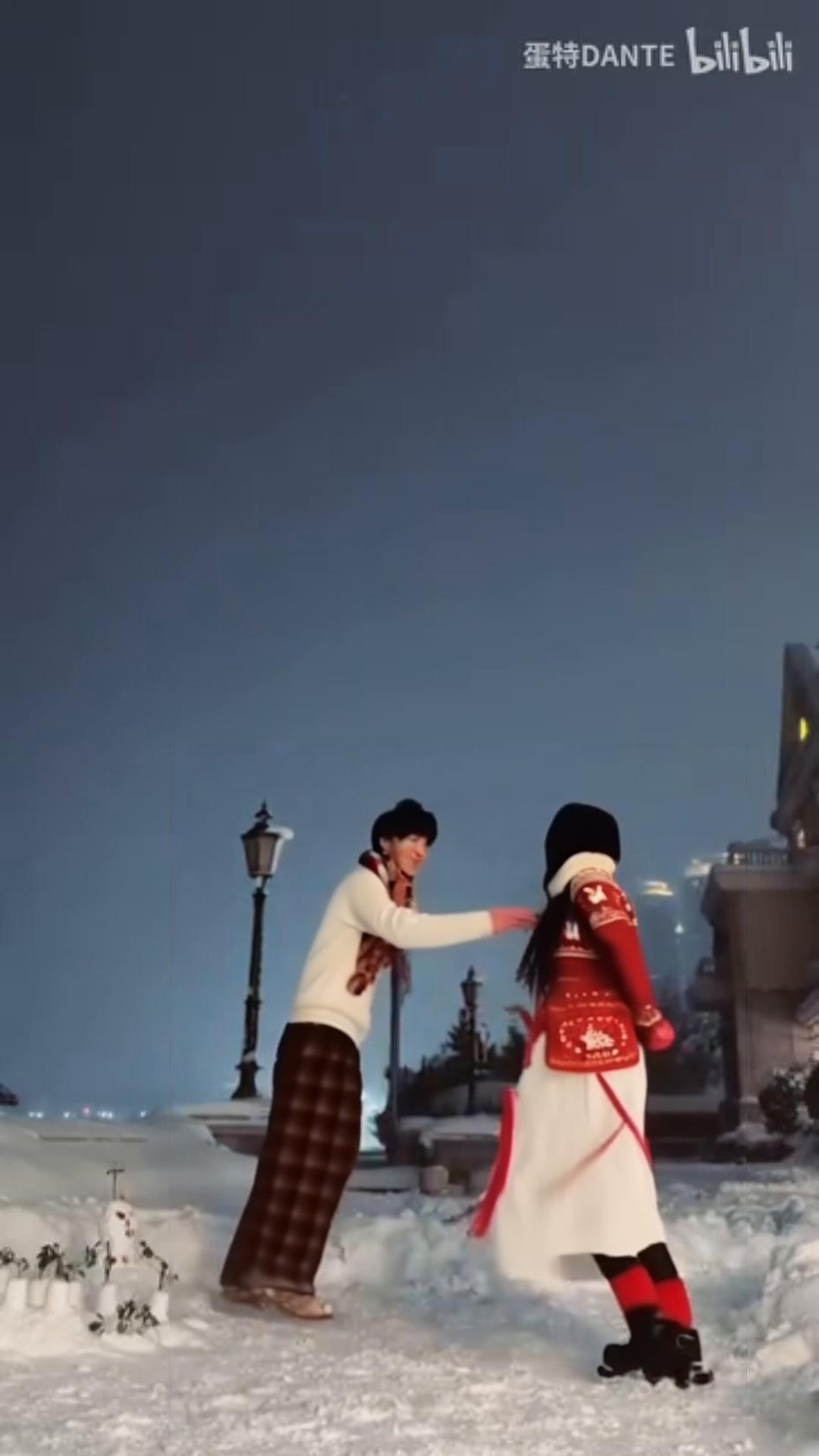}
    &\includegraphics[width=1.5cm]{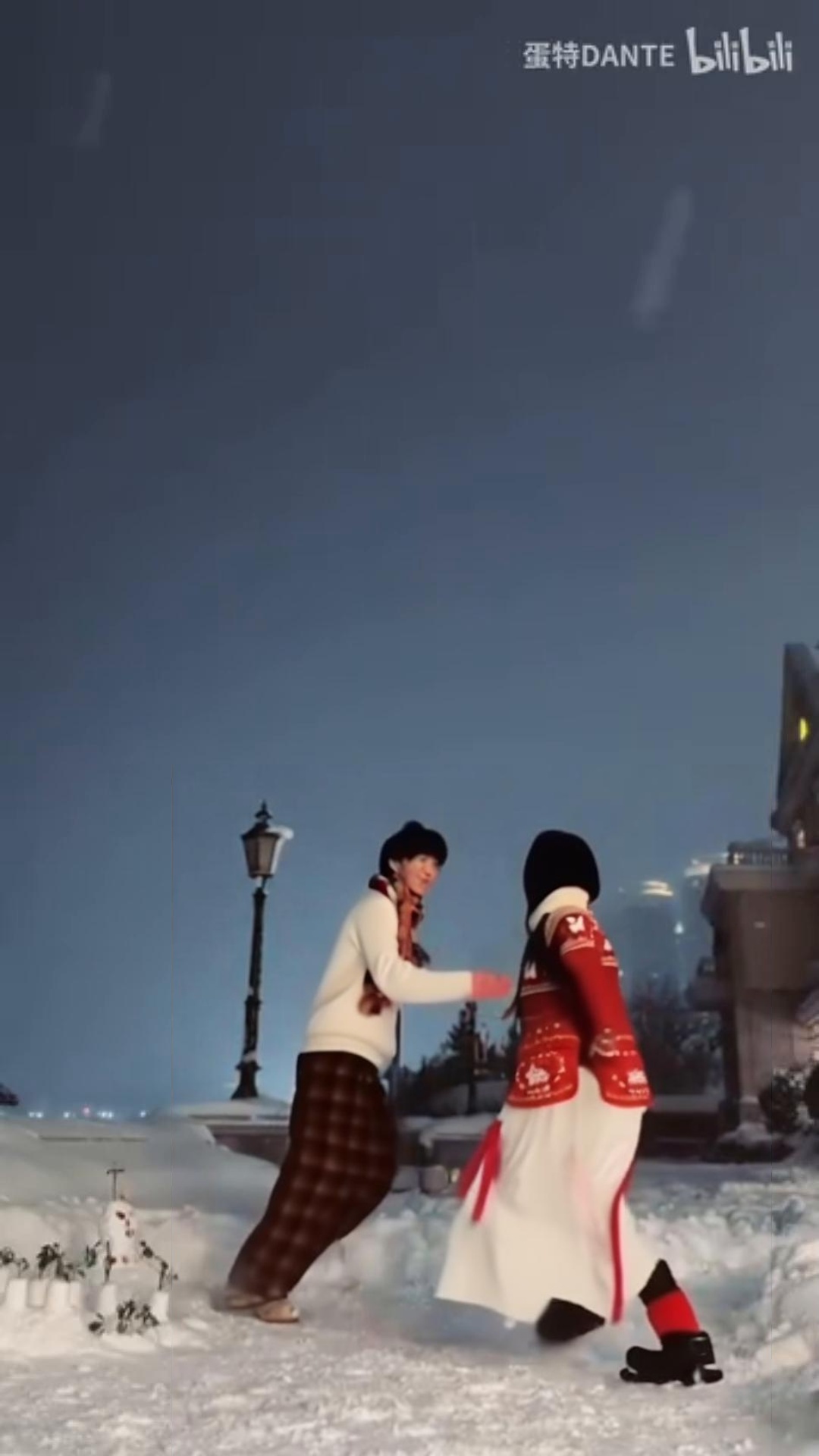}
    &\includegraphics[width=1.5cm]{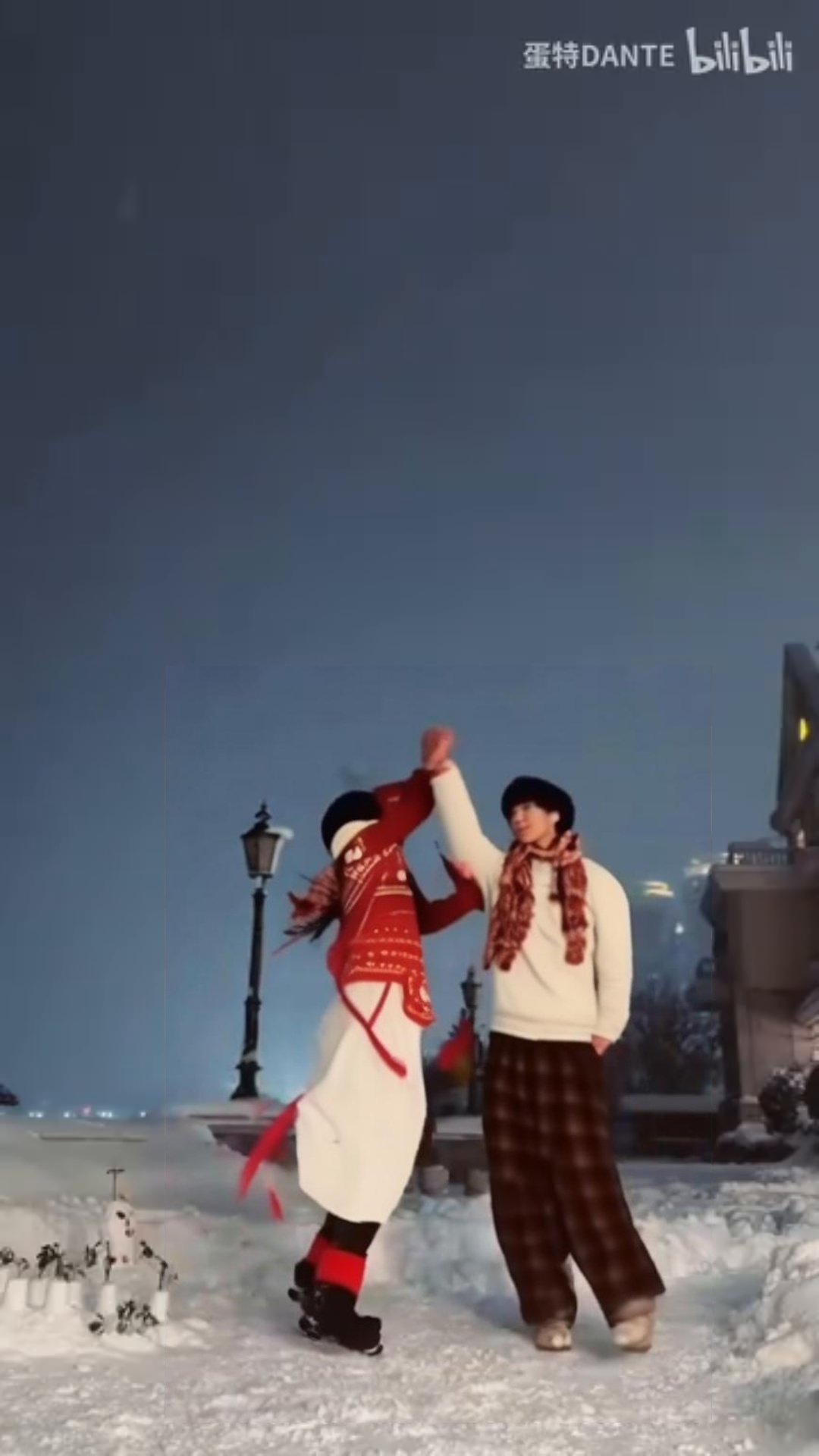}
    &\includegraphics[width=1.5cm]{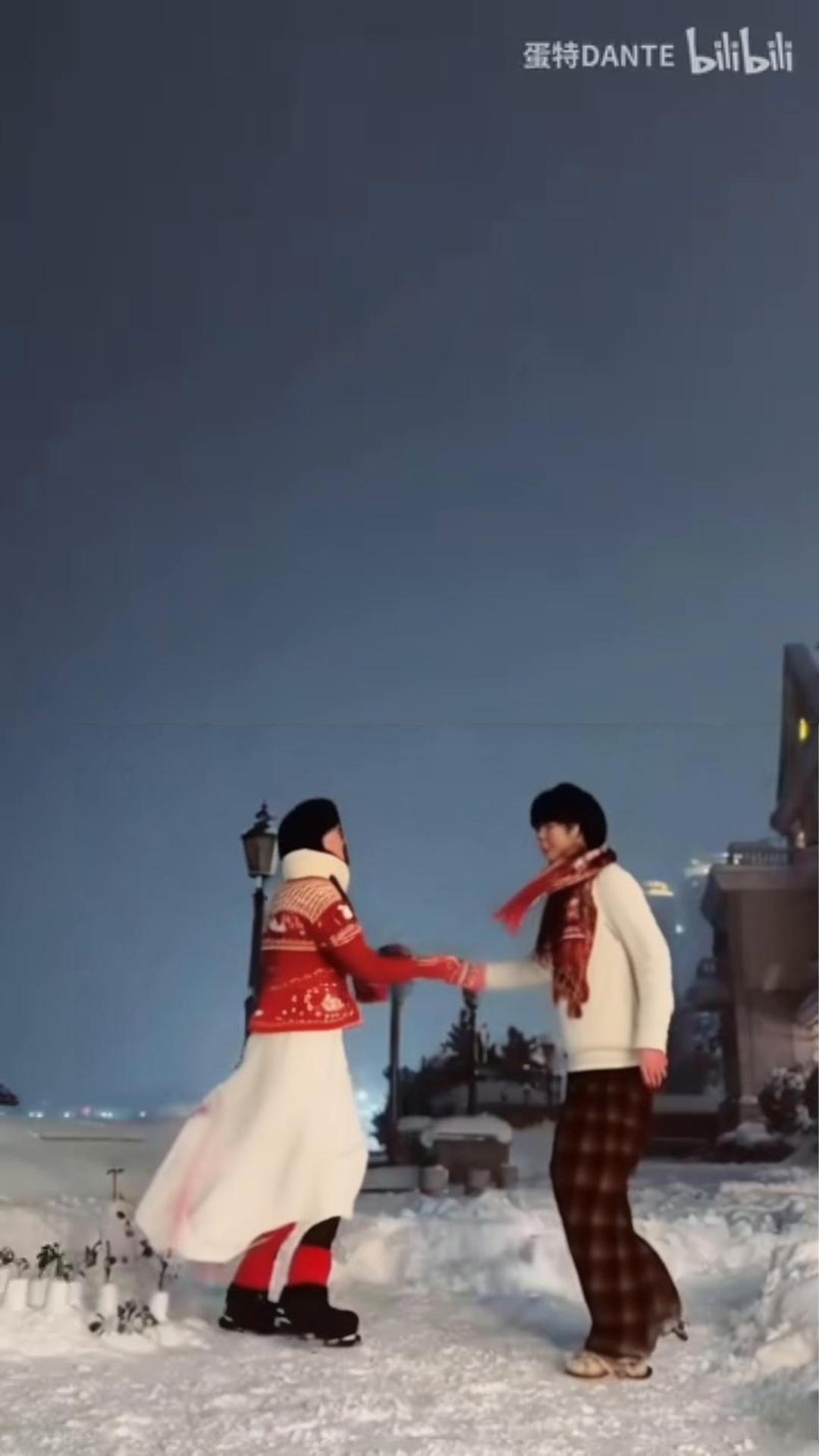}
    &\includegraphics[width=1.5cm]{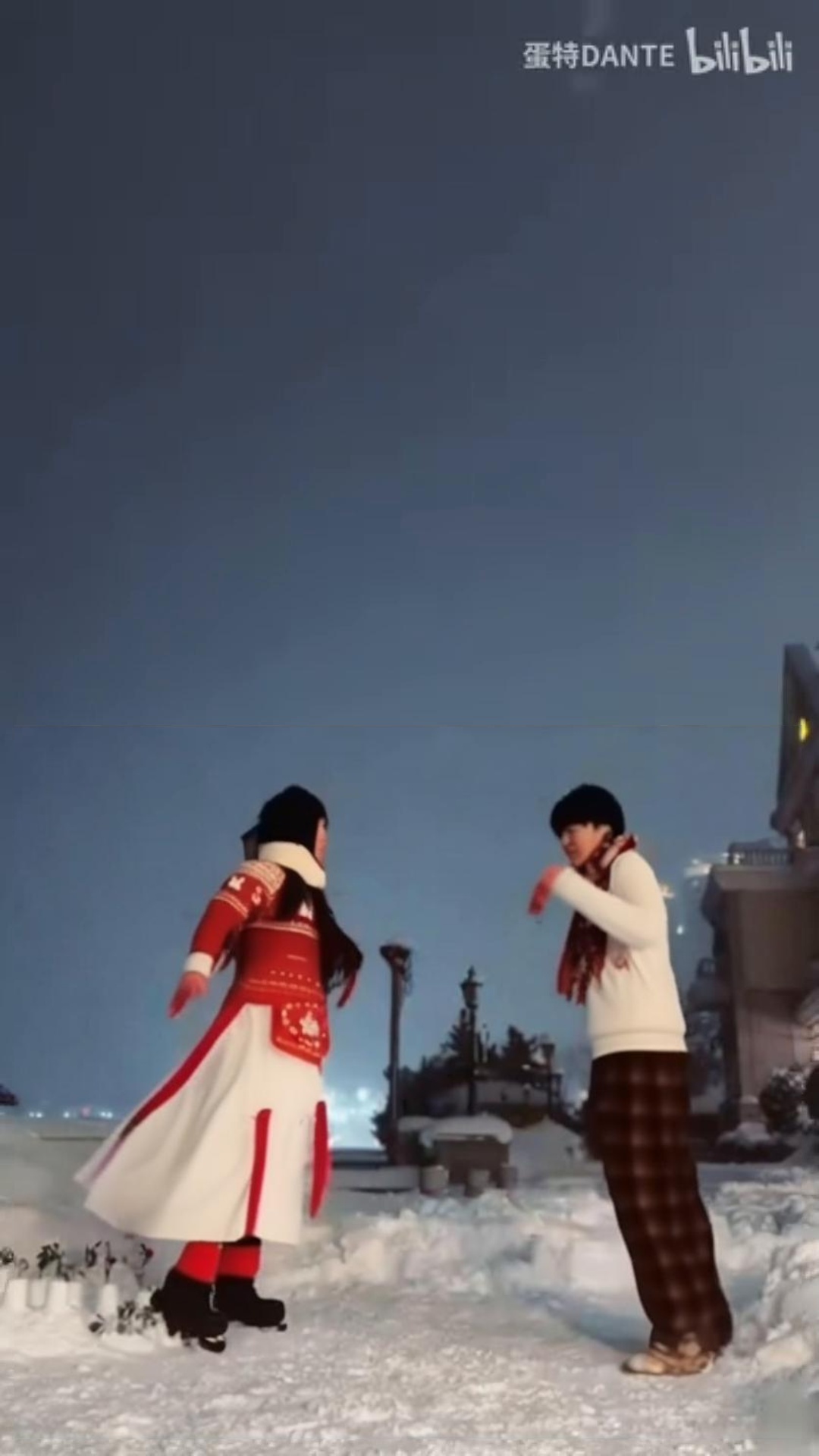}
    &\includegraphics[width=1.5cm]{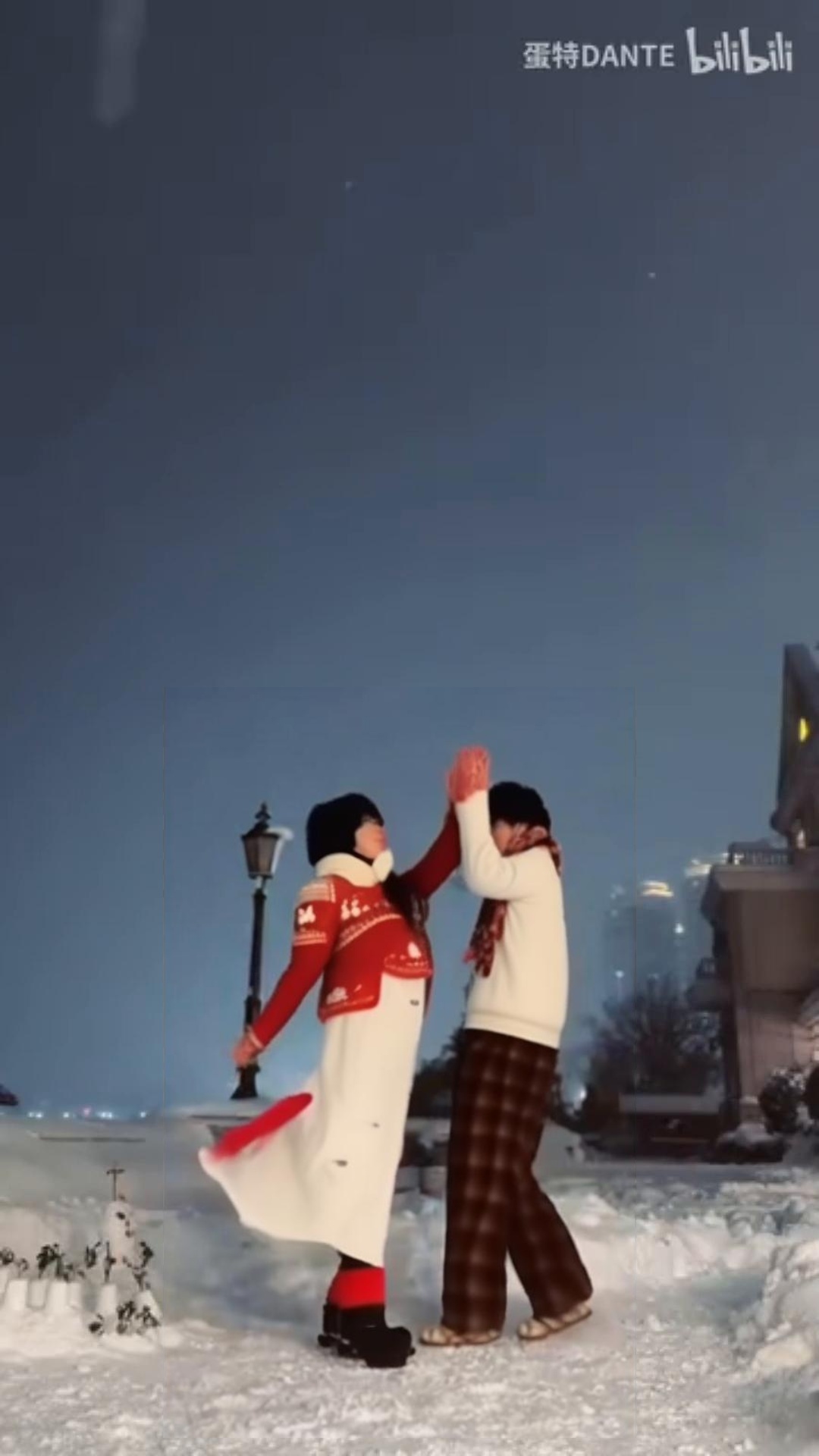}\\

    \raisebox{0.8cm}{\rotatebox[origin=c]{90}{\footnotesize{{Conv. Encoder}}}}
    &\includegraphics[width=1.5cm]{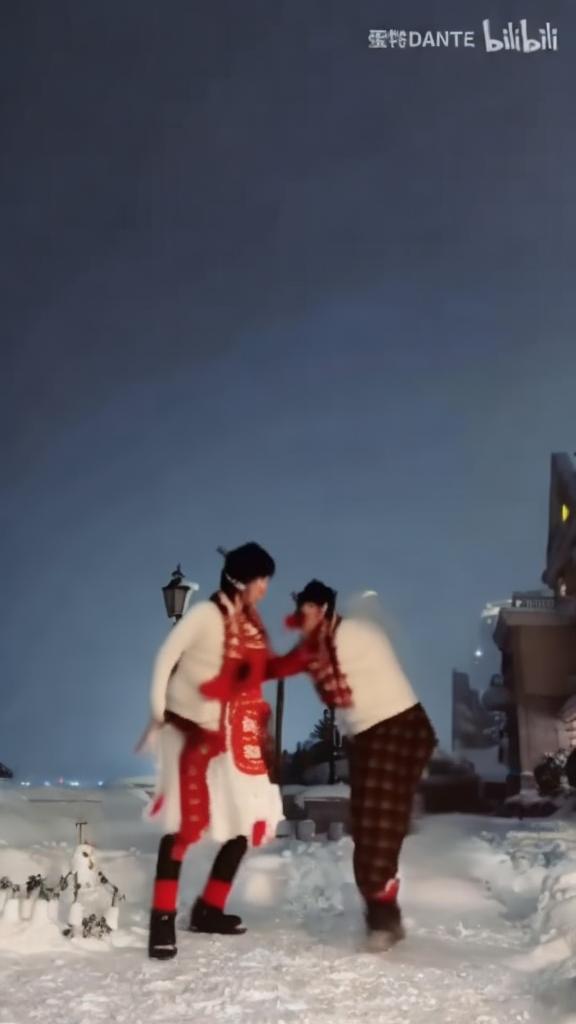}
    &\includegraphics[width=1.5cm]{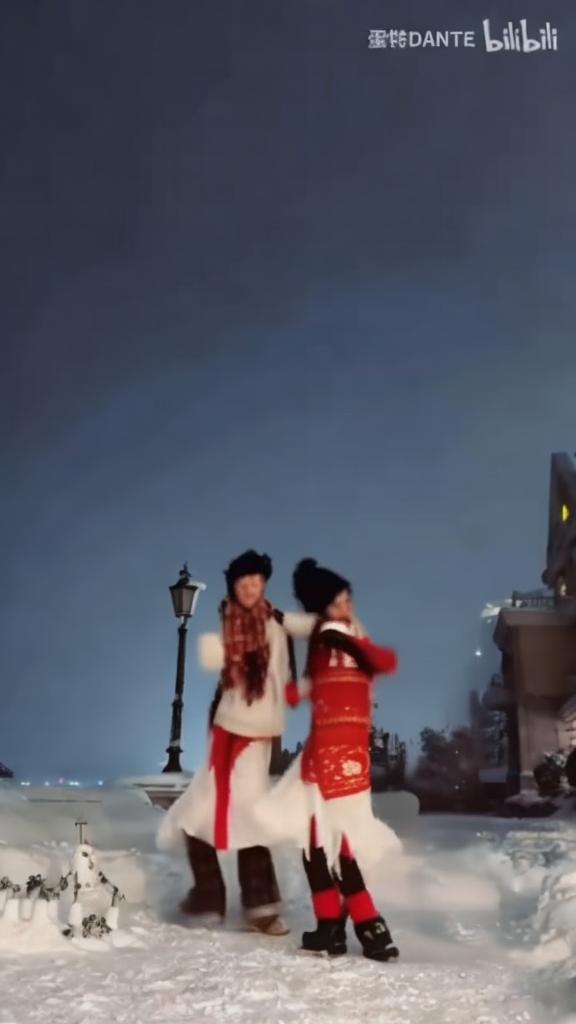}
    &\includegraphics[width=1.5cm]{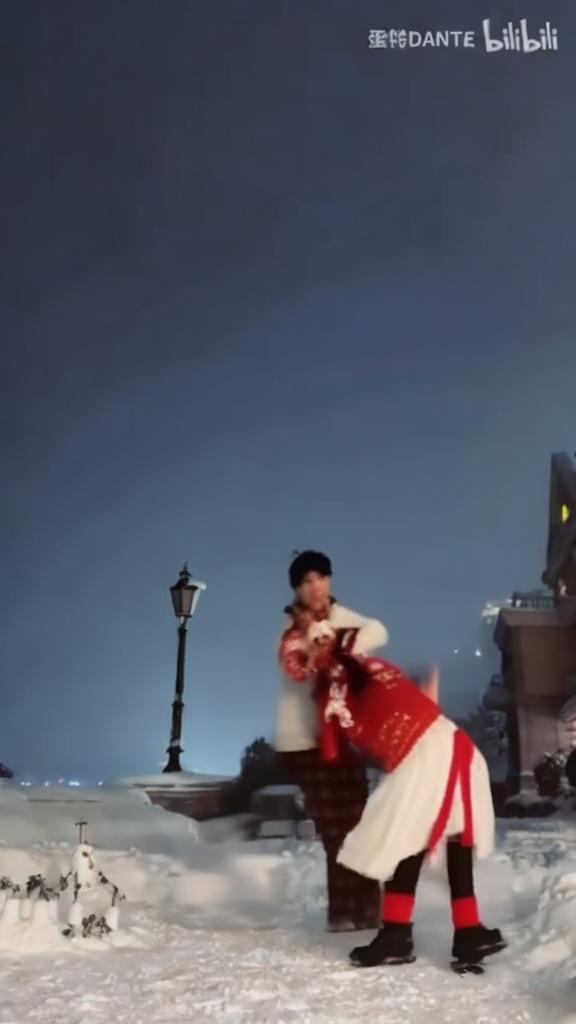}
    &\includegraphics[width=1.5cm]{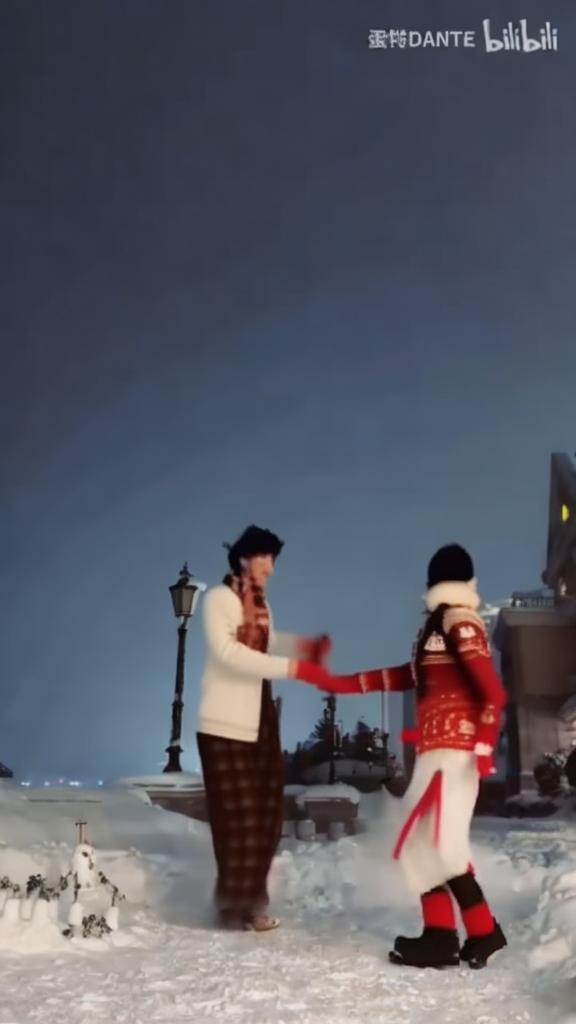}
    &\includegraphics[width=1.5cm]{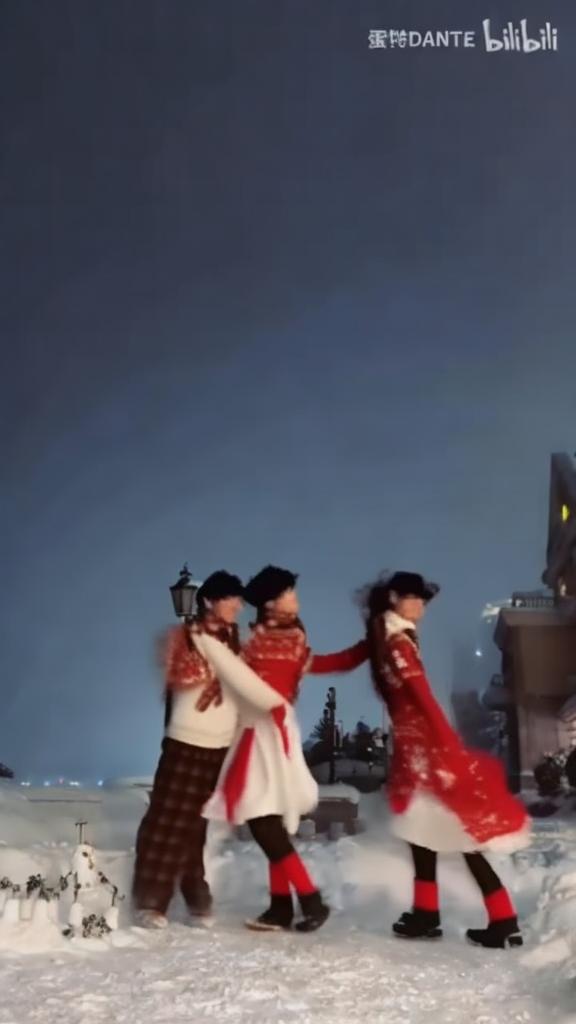}
    &\includegraphics[width=1.5cm]{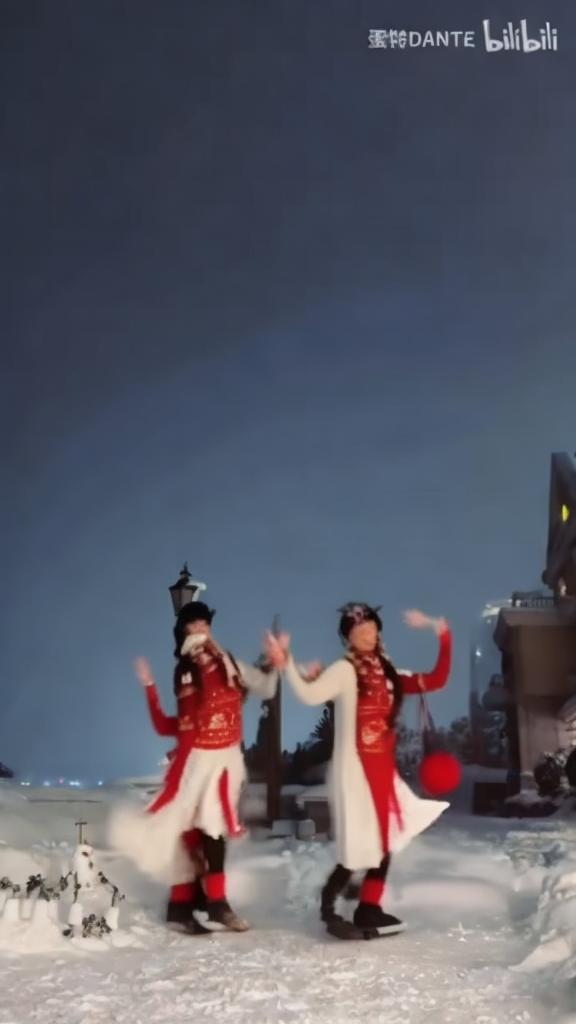}
    &\includegraphics[width=1.5cm]{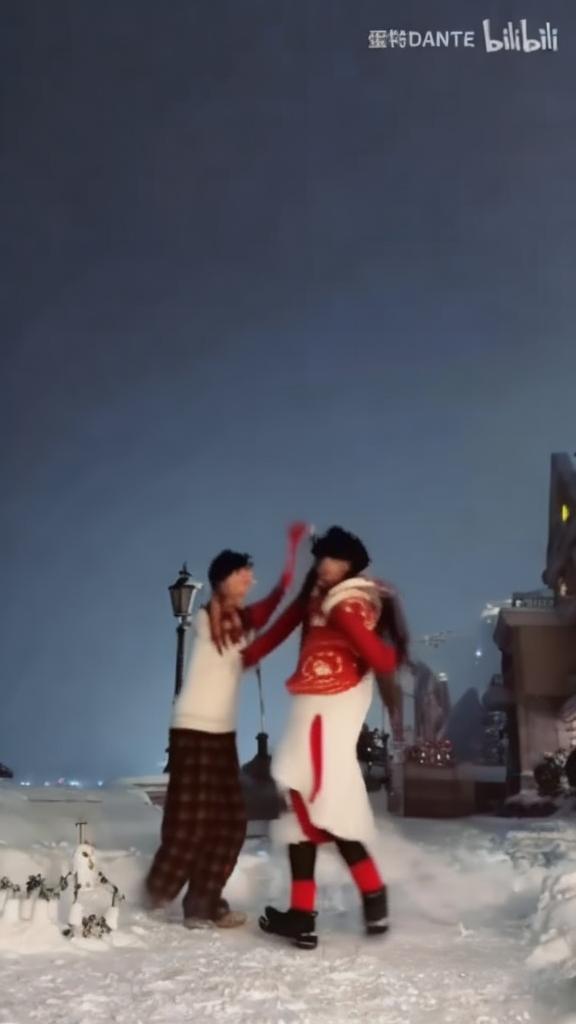}
    &\includegraphics[width=1.5cm]{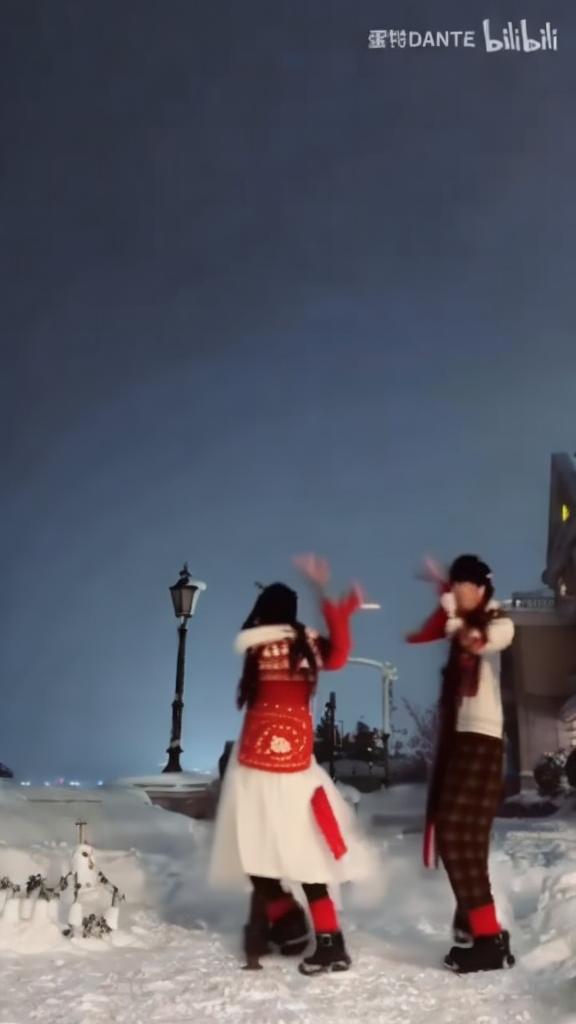}\\

    \raisebox{0.8cm}{\rotatebox[origin=c]{90}{\footnotesize{{Ground Truth}}}}
    &\includegraphics[width=1.5cm]{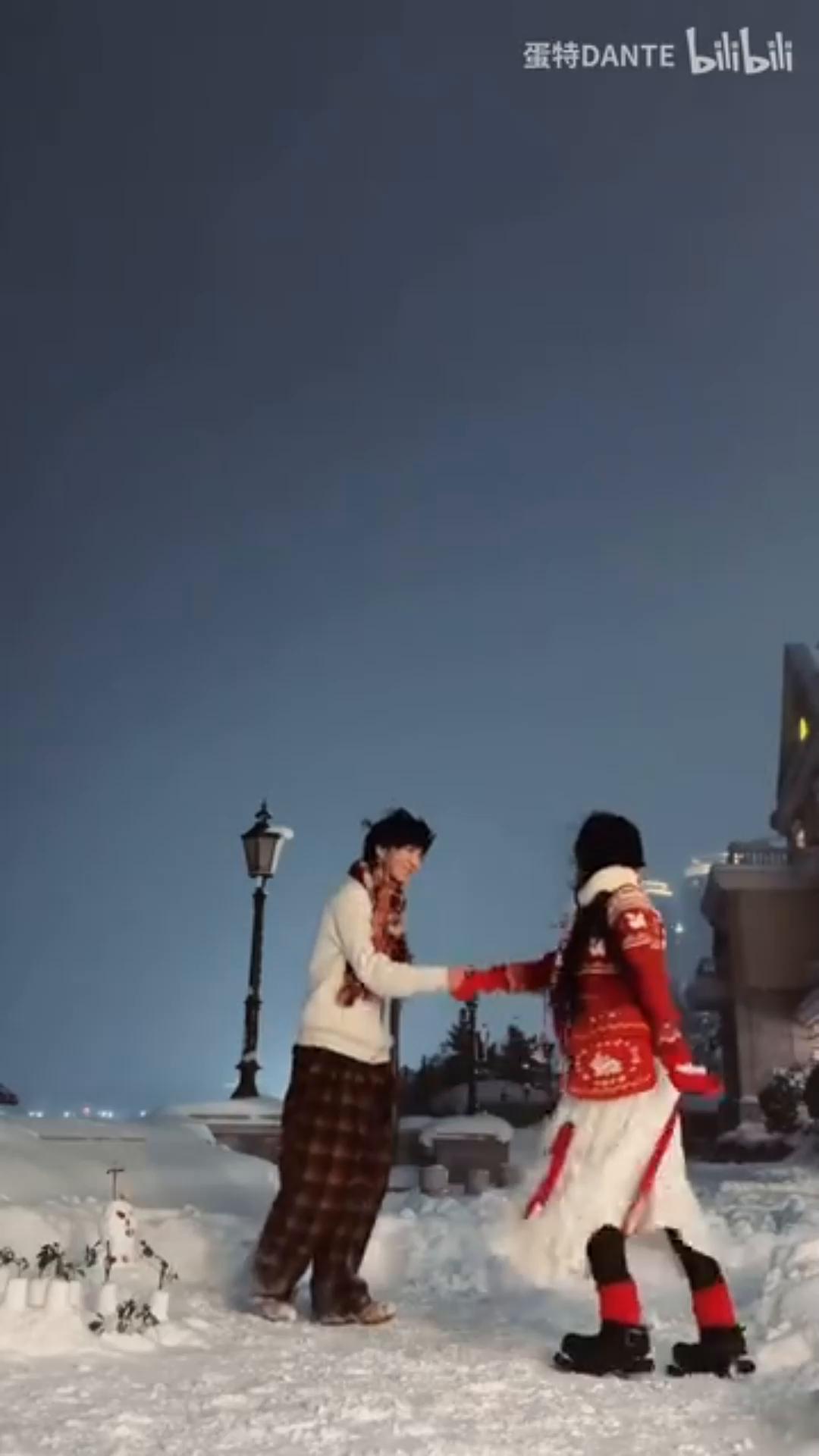}
    &\includegraphics[width=1.5cm]{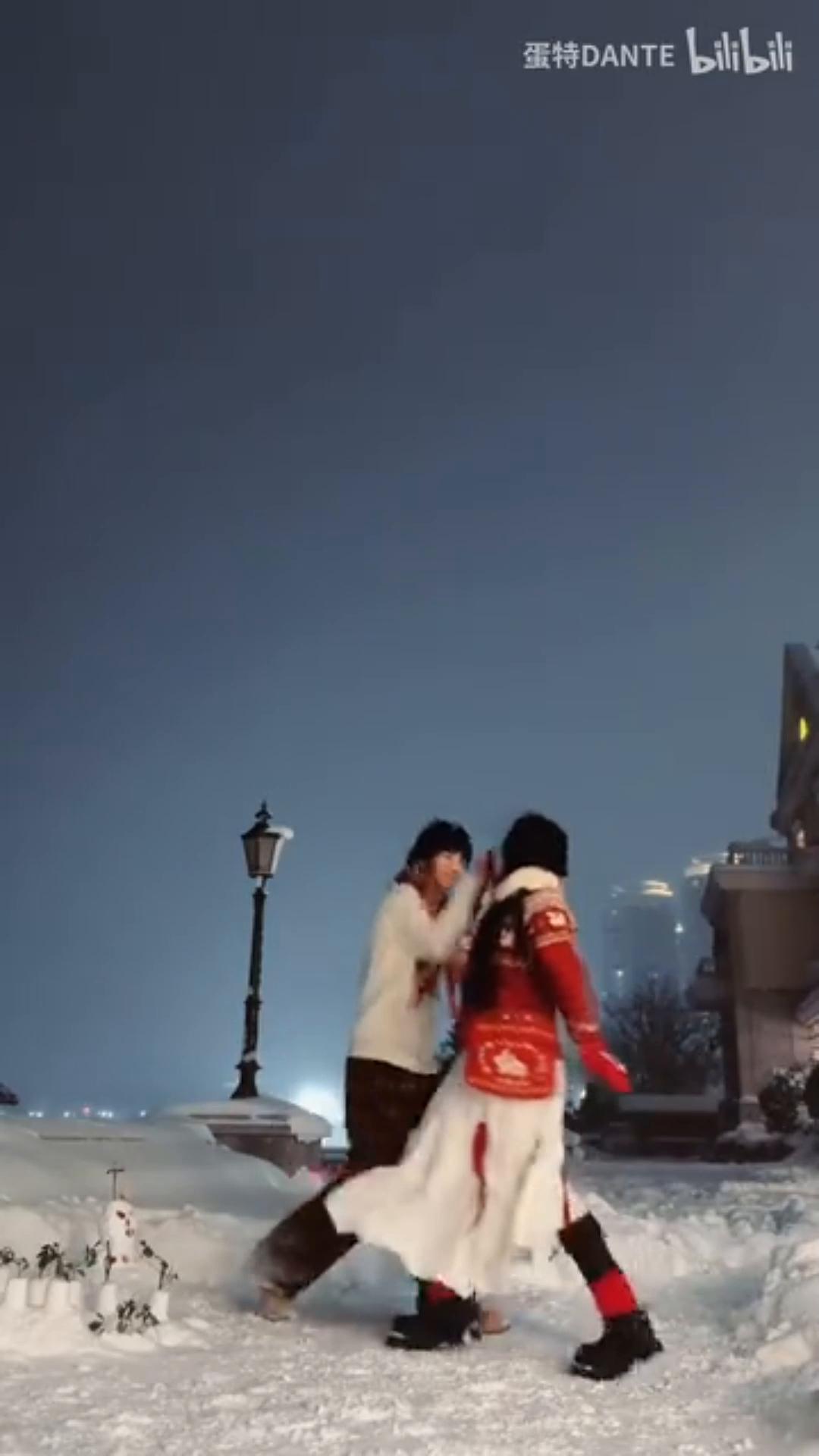}
    &\includegraphics[width=1.5cm]{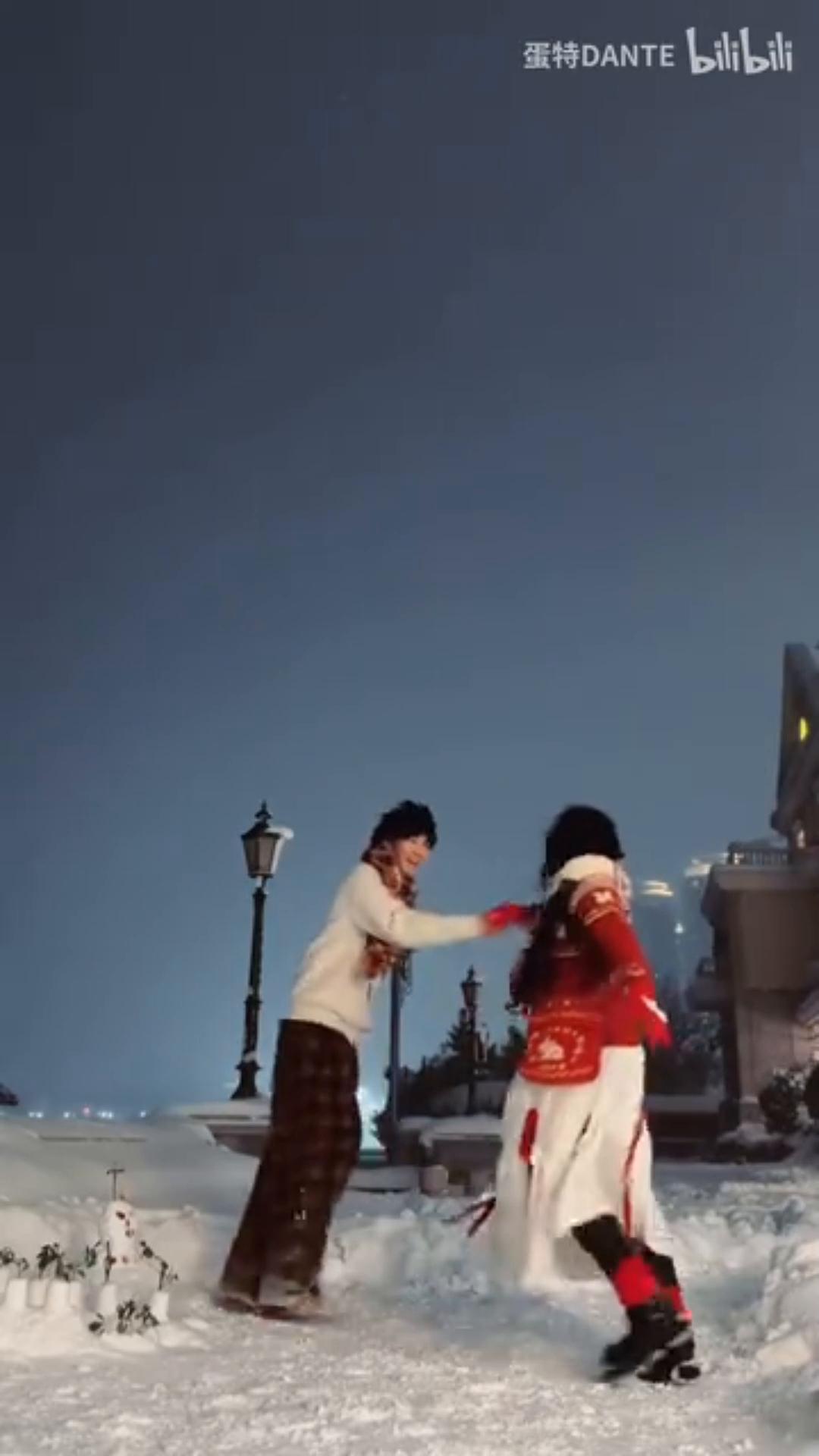}
    &\includegraphics[width=1.5cm]{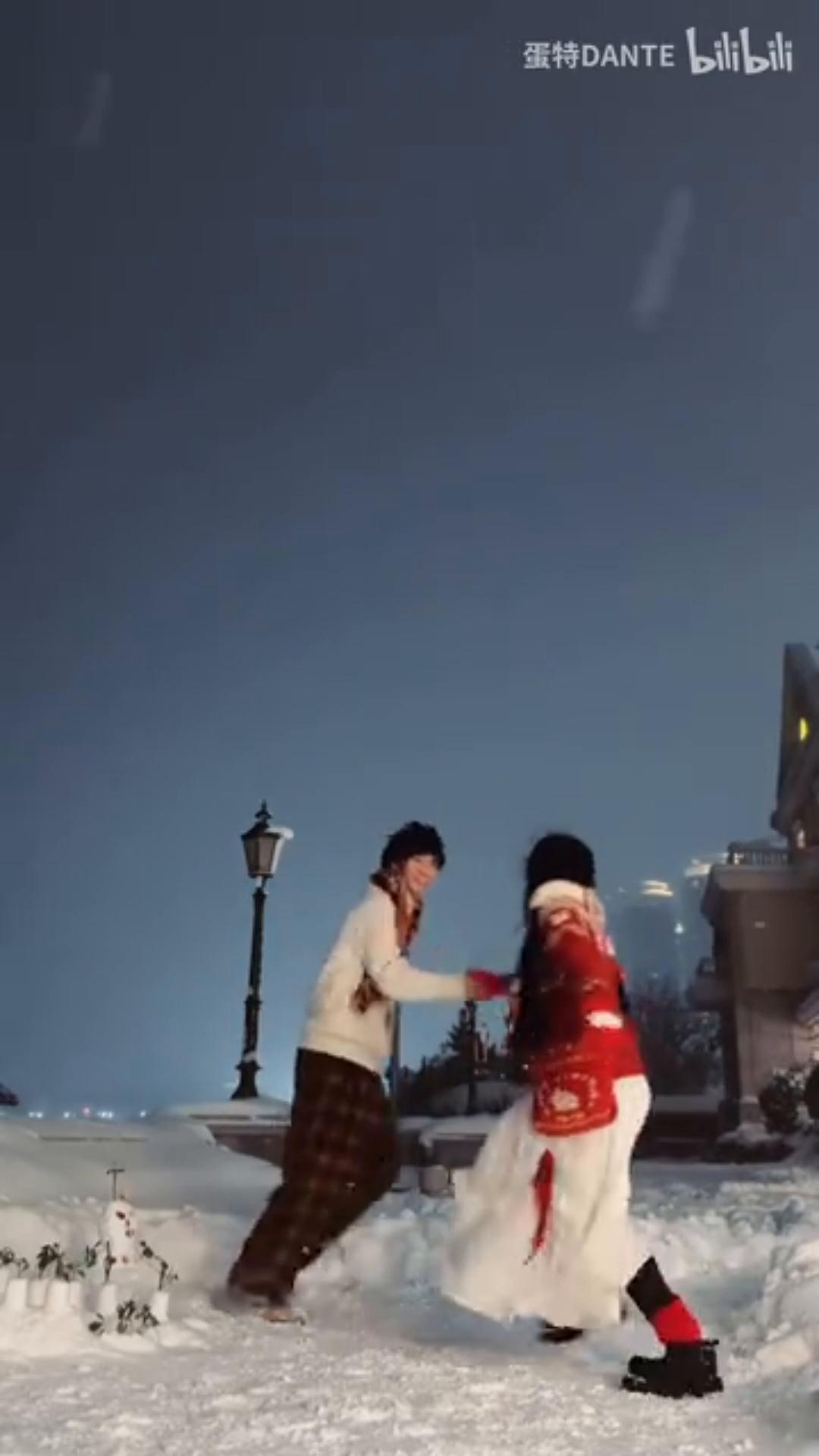}
    &\includegraphics[width=1.5cm]{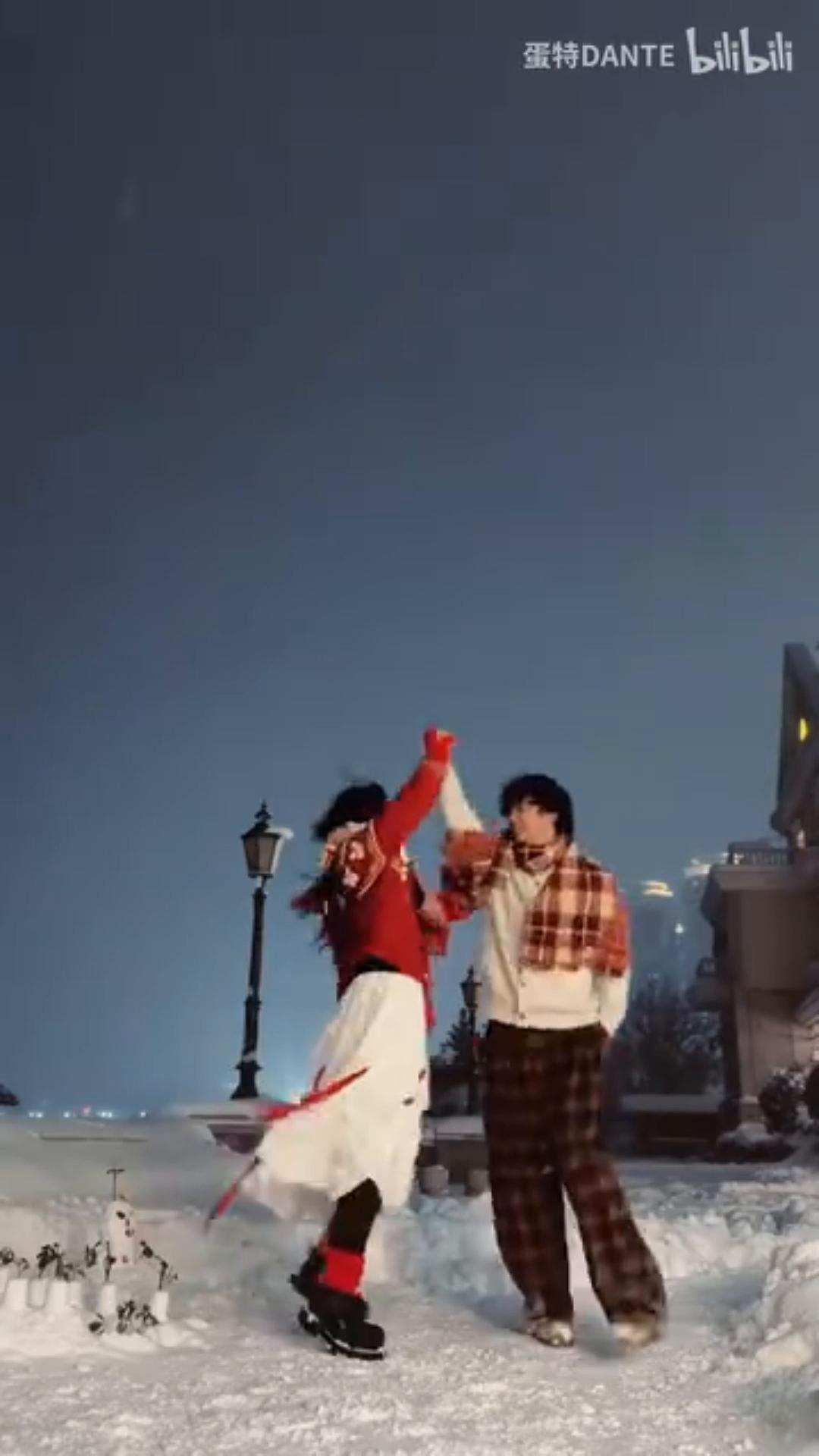}
    &\includegraphics[width=1.5cm]{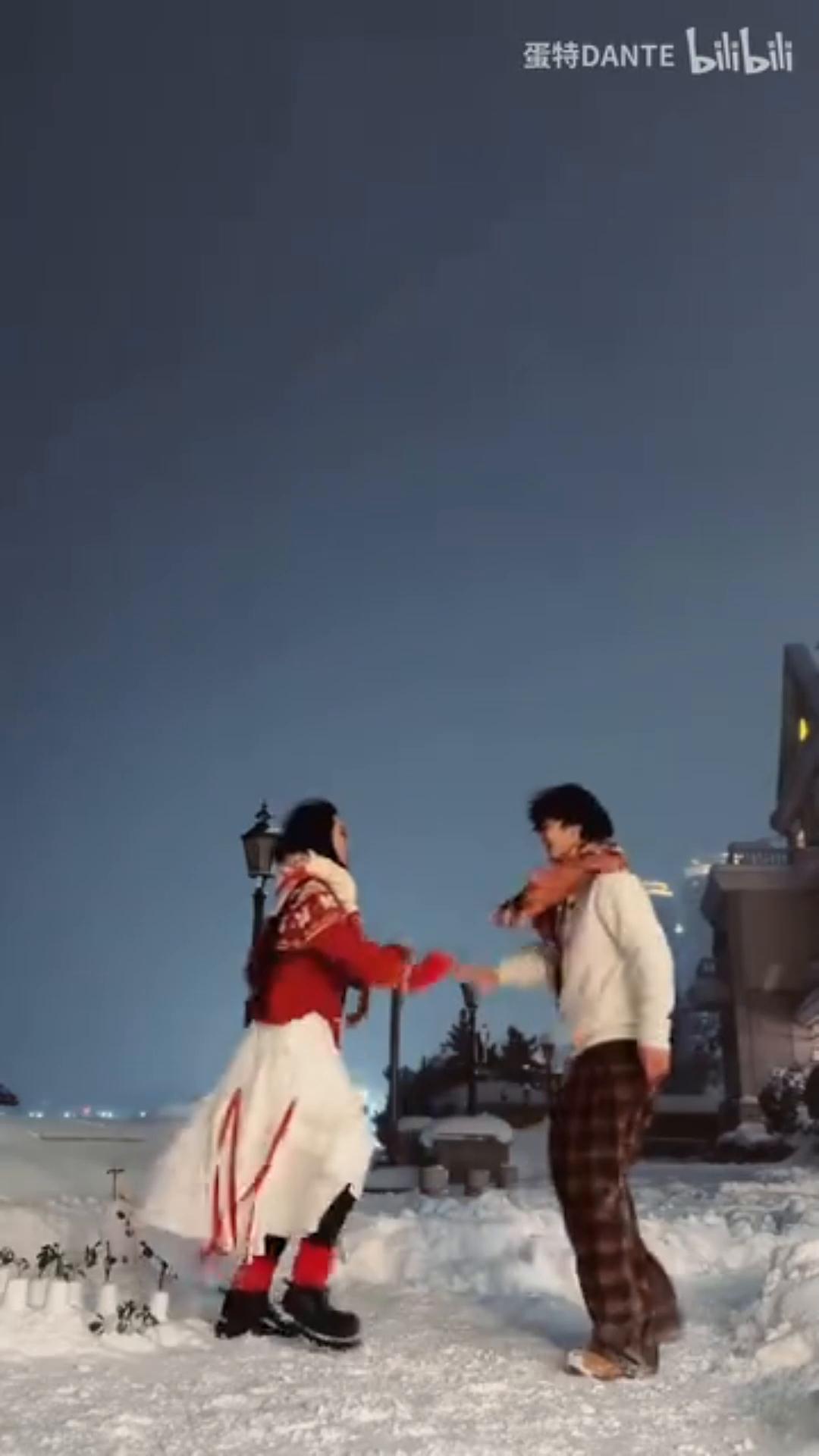}
    &\includegraphics[width=1.5cm]{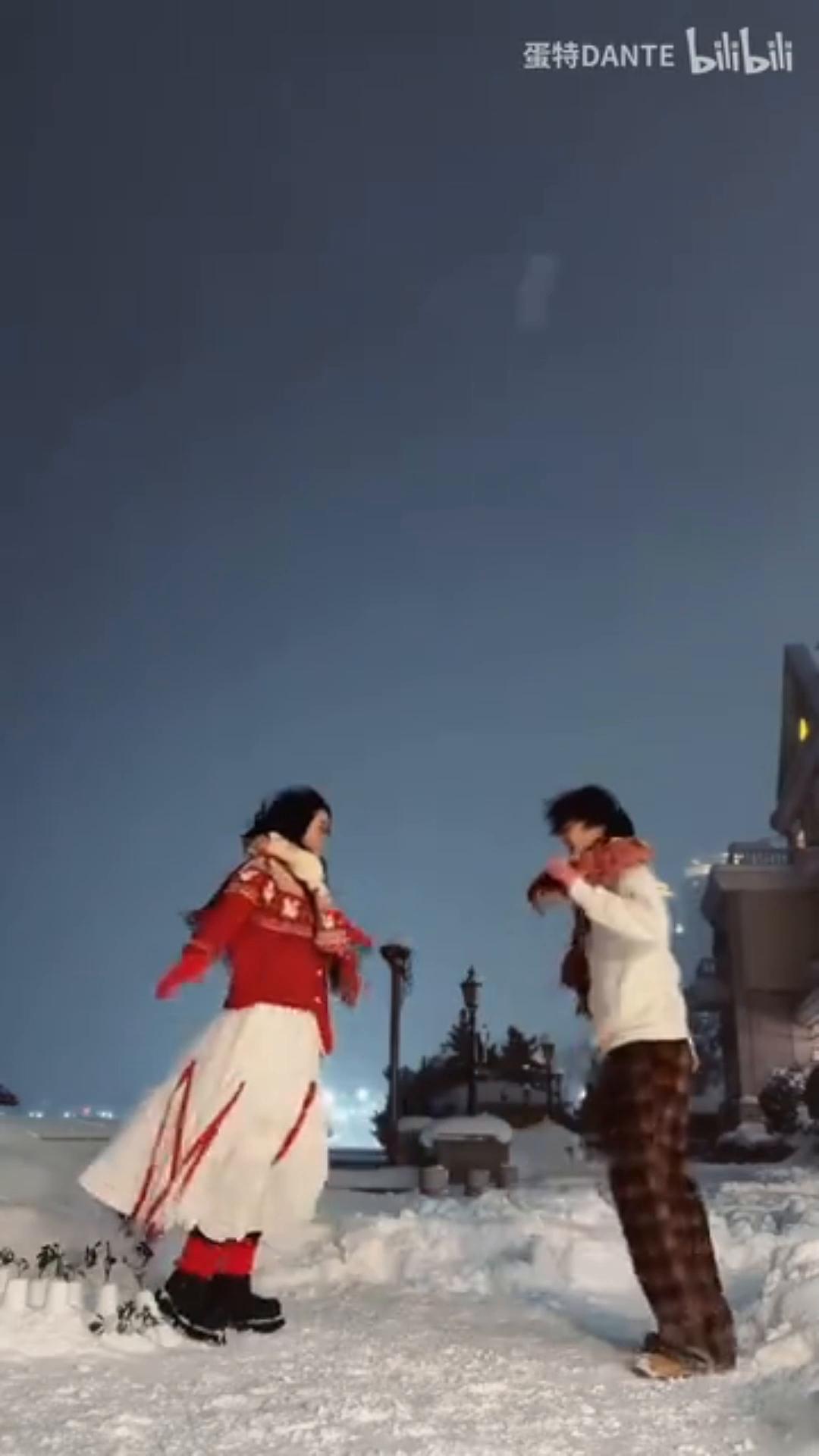}
    &\includegraphics[width=1.5cm]{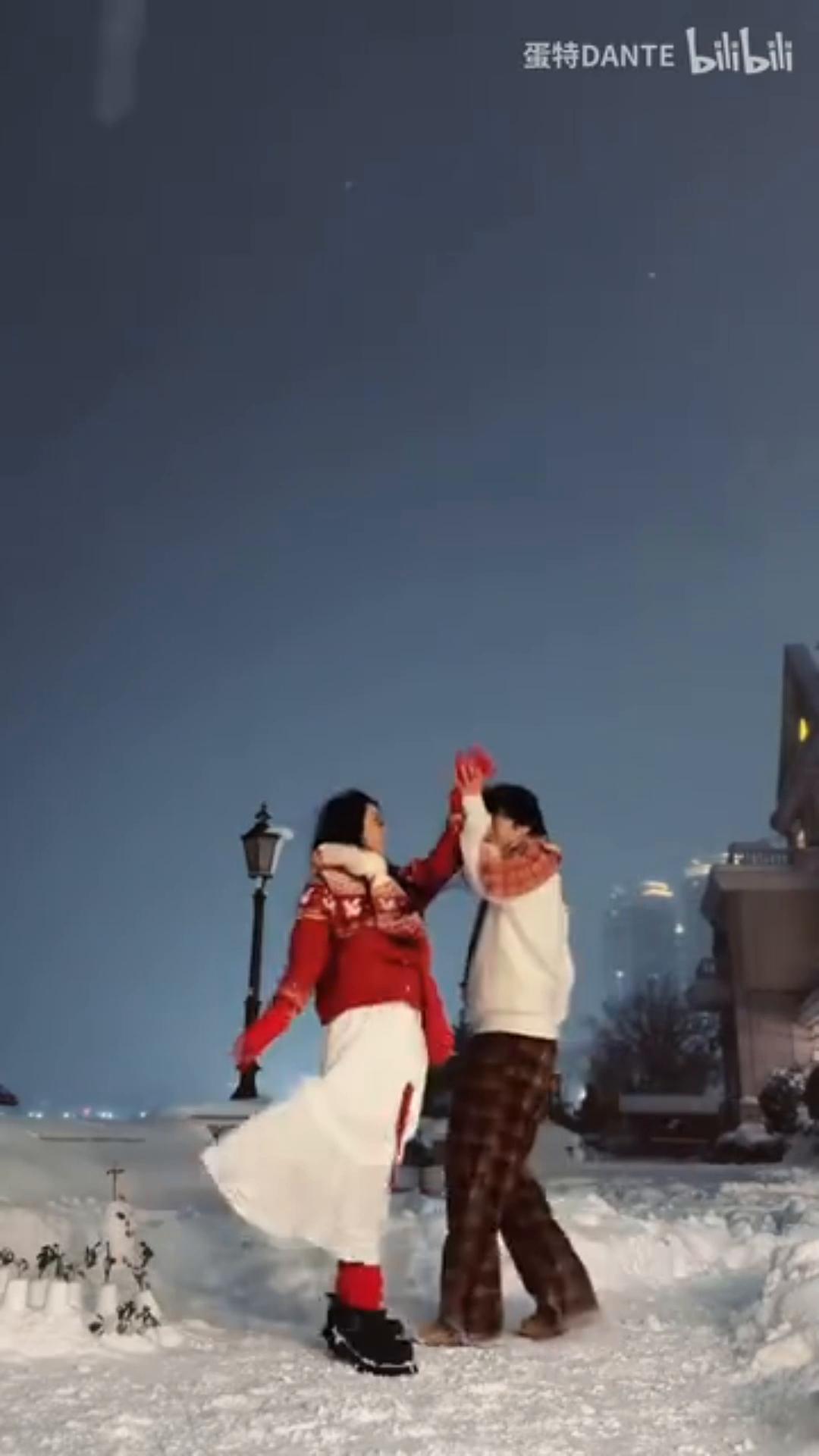}\\

    \end{tabular}
    \end{center}
    \caption{Comparison on replacing the identity-aware pose encoder with convolutional pose encoder. Artifacts appear when animation complex interaction such as position interchange without explicit identity-pose binding.} 
    \label{fig:ab-follow1}
\end{figure*}

\subsubsection{Identity-Aware Pose Encoder.} We present the ablation study on the identity-aware pose encoder in Figure~\ref{fig:ablation-id}. In experiments without the identity-aware pose encoder, we employ a conventional approach implemented in prior works, where pose sequences are directly aggregated with noise following feature extraction through convolutional networks. For comparative analysis, we conduct pose sequence interchange experiments, where we swap the corresponding pose sequences between reference images to generate paired results. 

The experimental results demonstrate that without the identity-aware pose encoder, the generated outputs exhibit identity ambiguities, as the pose sequences share a common feature extraction network lacking the capability to differentiate between distinct pose features. Furthermore, in this configuration, interchanging pose sequences between reference images becomes ineffective, as these features are merely aggregated within the latent space. In contrast, our proposed identity-aware pose encoder efficiently processes multiple pose sequences while maintaining distinct identity associations, enabling accurate character animation even during pose sequence interchange operations. This demonstrates the successful establishment of robust pose-identity bindings.

As Xue et al.~\cite{xue2025multiplecharacterimageanimation} utilizes pose sequences as images that contain poses of all characters without explicit identity differentiation. We substituted our identity-aware pose encoder with their approach, and the comparative results are presented in Figure~\ref{fig:ab-follow1}. The experimental results demonstrate that the absence of explicit identity-pose binding leads to significant limitations, particularly in animating complex character interactions such as position interchange. Specifically, the inability to bind identities with their corresponding pose sequences results in animation artifacts and incorrect character behaviors. In contrast, our proposed identity-aware pose encoder successfully maintains identity consistency and accurately handles these complex interactive scenarios.

\begin{figure}[]
    \begin{center}
    \setlength{\tabcolsep}{0.5pt}
    \begin{tabular}{m{1.85cm}<{\centering}m{1.85cm}<{\centering}m{1.85cm}<{\centering}m{1.85cm}<{\centering}m{1.85cm}<{\centering}}

    \scriptsize{Reference 1} & \scriptsize{Reference 2} & \scriptsize{Ground Truth Relation} & \scriptsize{w/o. Interaction Guider} &\scriptsize{w. Interaction Guider} \\
    \includegraphics[width=1.8cm]{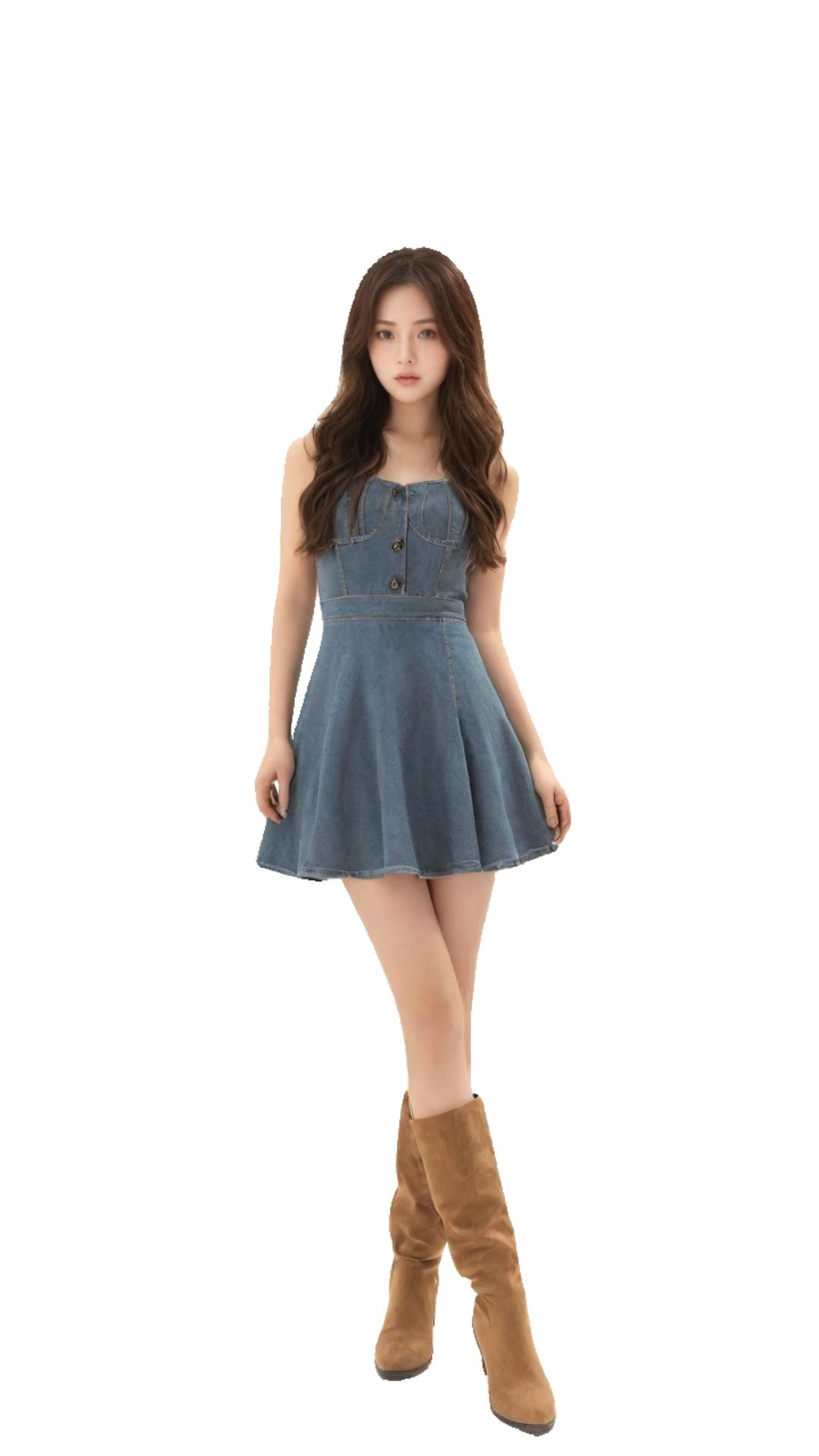}
    % &\includegraphics[width=1.5cm]{images/ablation/position/pose_0-res-1.jpg}
    &\includegraphics[width=1.8cm]{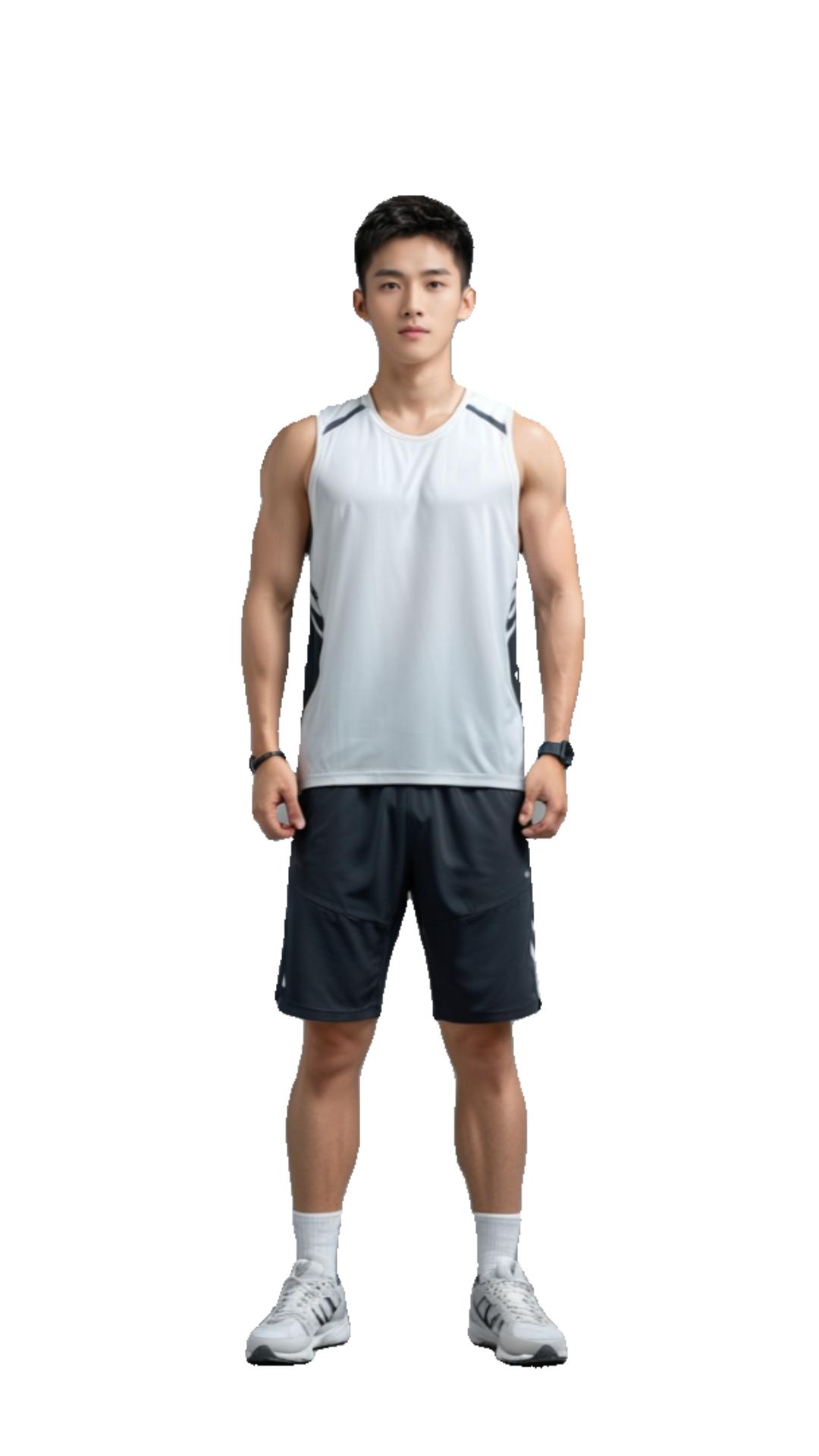}
    % &\includegraphics[width=1.5cm]{images/ablation/position/pose_1-res-1.jpg}
    &\includegraphics[width=1.8cm]{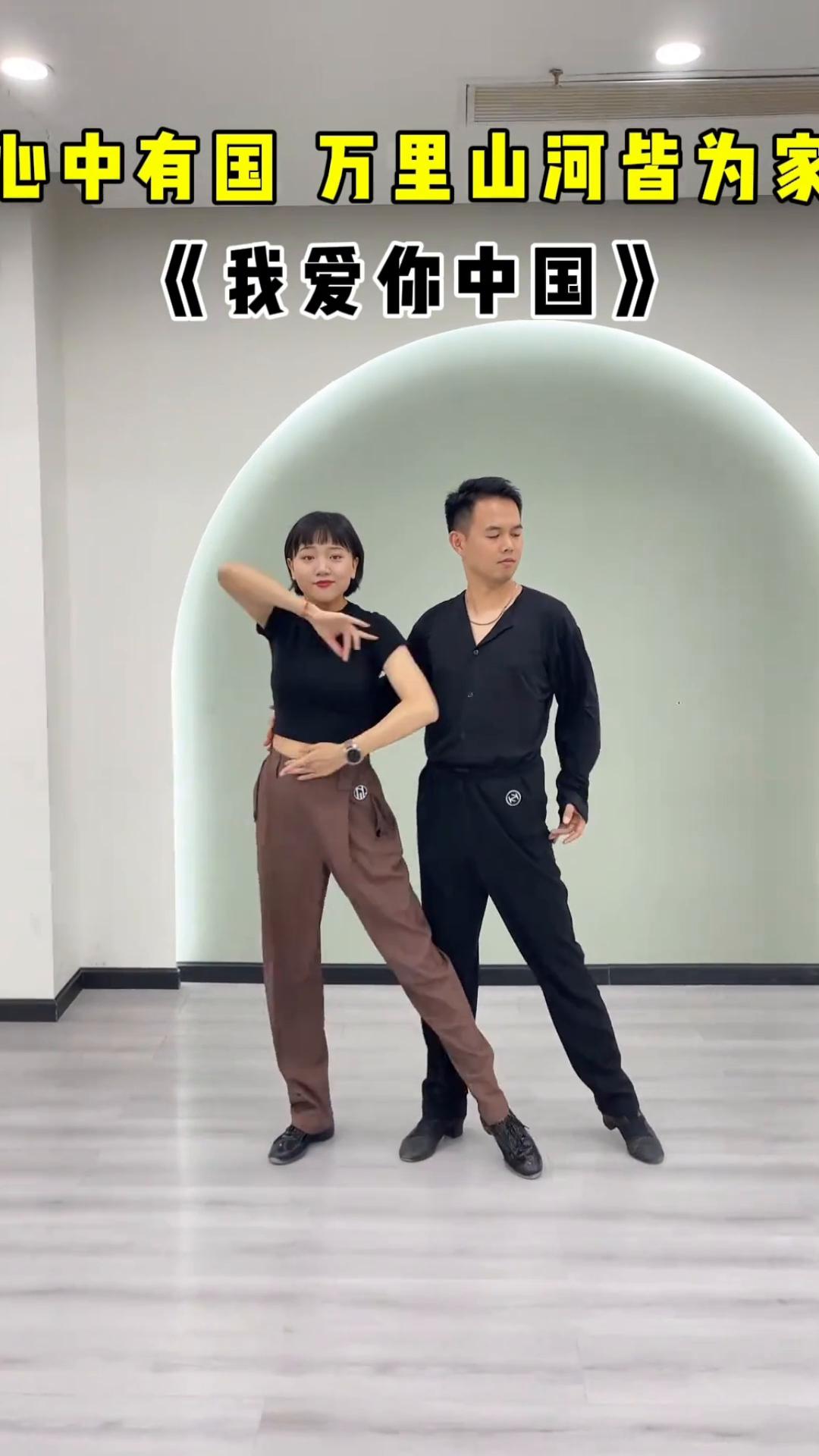}
    &\includegraphics[width=1.8cm]{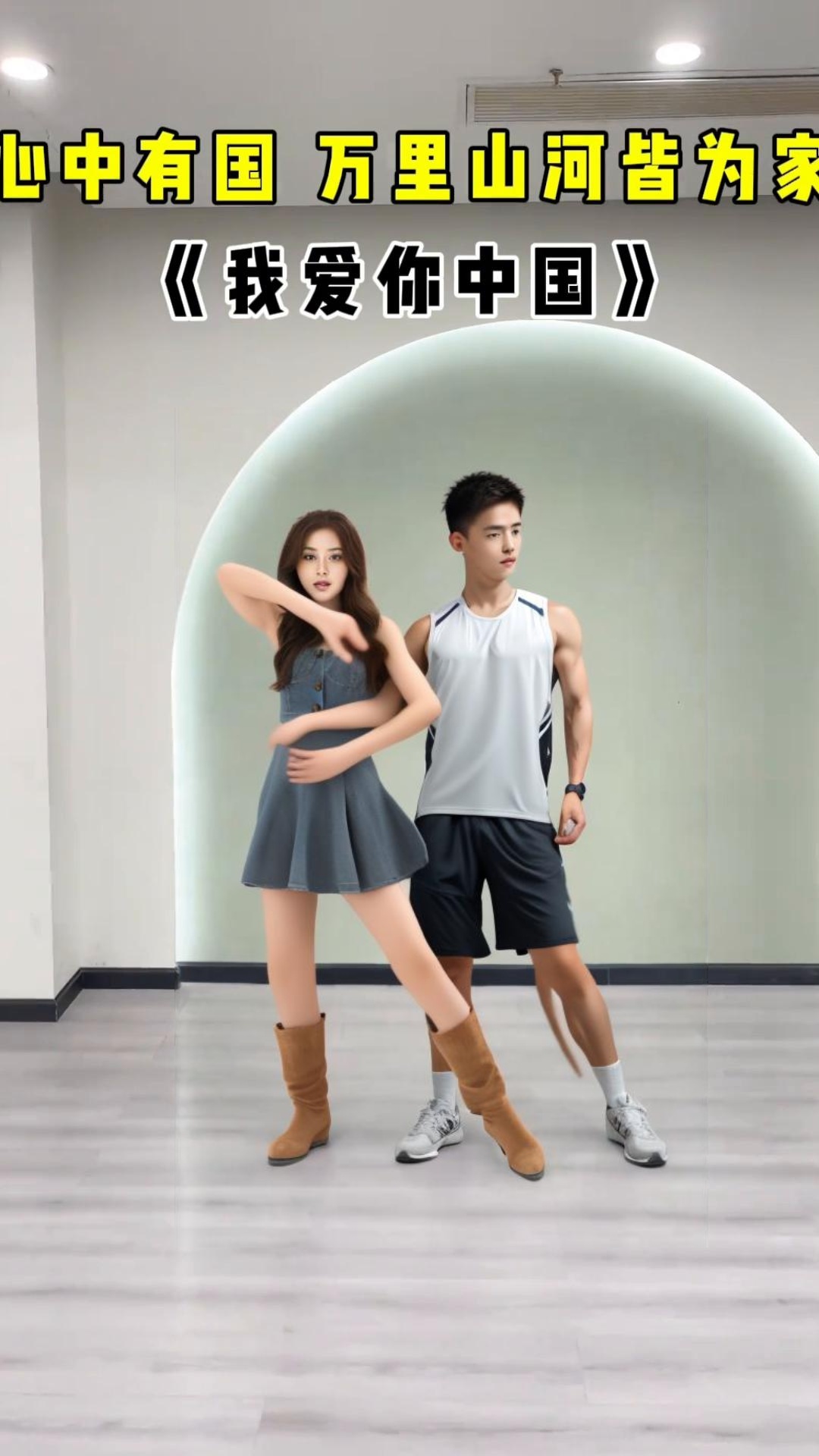}
    &\includegraphics[width=1.8cm]{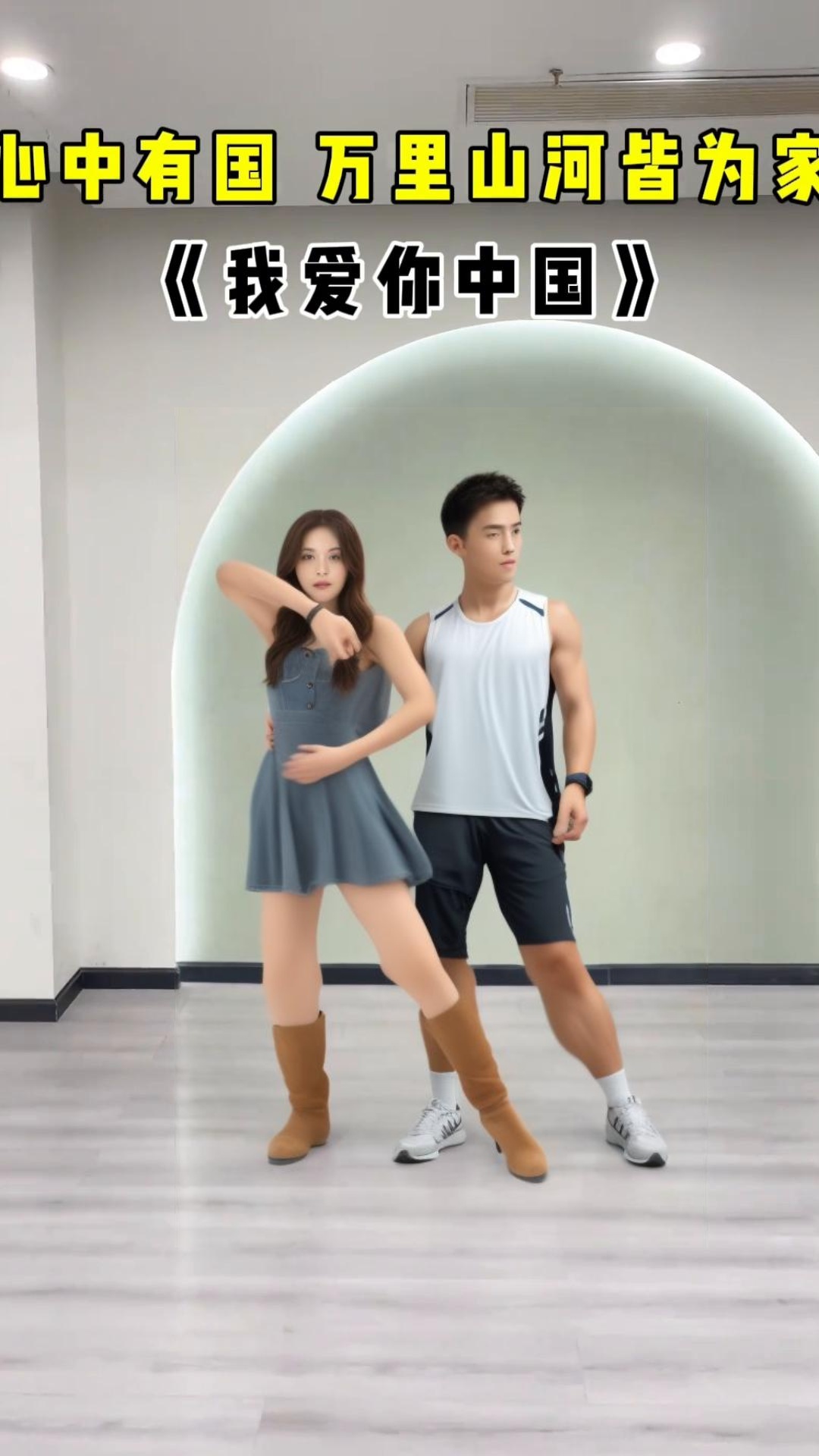} \\
    
    \end{tabular}
    \end{center}
    \caption{Ablation study on the interaction guider module. Interaction guider module enhances the framework's capability to generate accurate inter-character interactions, particularly in scenarios involving complex spatial relationships.} 
    \label{fig:ablation-mask}
\end{figure}

\subsubsection{Interaction Guider.} Figure~\ref{fig:ablation-mask} demonstrates the impact of the interaction guider through ablation experiments. Without the interaction guider, the model exhibits deficiencies in handling complex inter-character interactions, particularly in resolving occlusion relationships. The incorporation of the interaction guider enables the model to effectively distinguish masked regions, resulting in accurate character interaction synthesis.

\begin{figure}[h]
    \begin{center}
    \setlength{\tabcolsep}{0.5pt}
    \begin{tabular}{m{1.85cm}<{\centering}m{1.85cm}<{\centering}m{1.85cm}<{\centering}m{1.85cm}<{\centering}m{1.85cm}<{\centering}}

    \scriptsize{Reference 1} & \scriptsize{Reference 2} & \scriptsize{Ground Truth Relation} & \scriptsize{w. Interaction Guider} &\scriptsize{w. Depth Order Guider} \\
    \includegraphics[width=1.8cm]{images/ablation/guider/res-1-ref-1.jpg}
    &\includegraphics[width=1.8cm]{images/ablation/guider/res-1-ref-2.jpg}
    &\includegraphics[width=1.8cm]{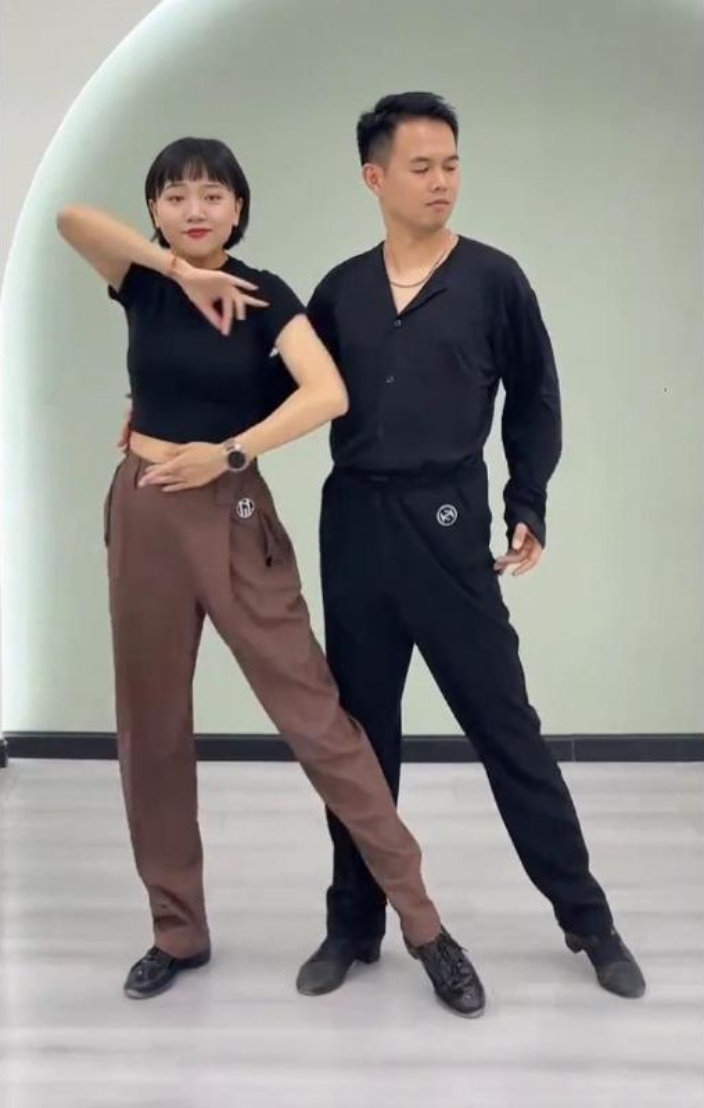}
    &\includegraphics[width=1.8cm]{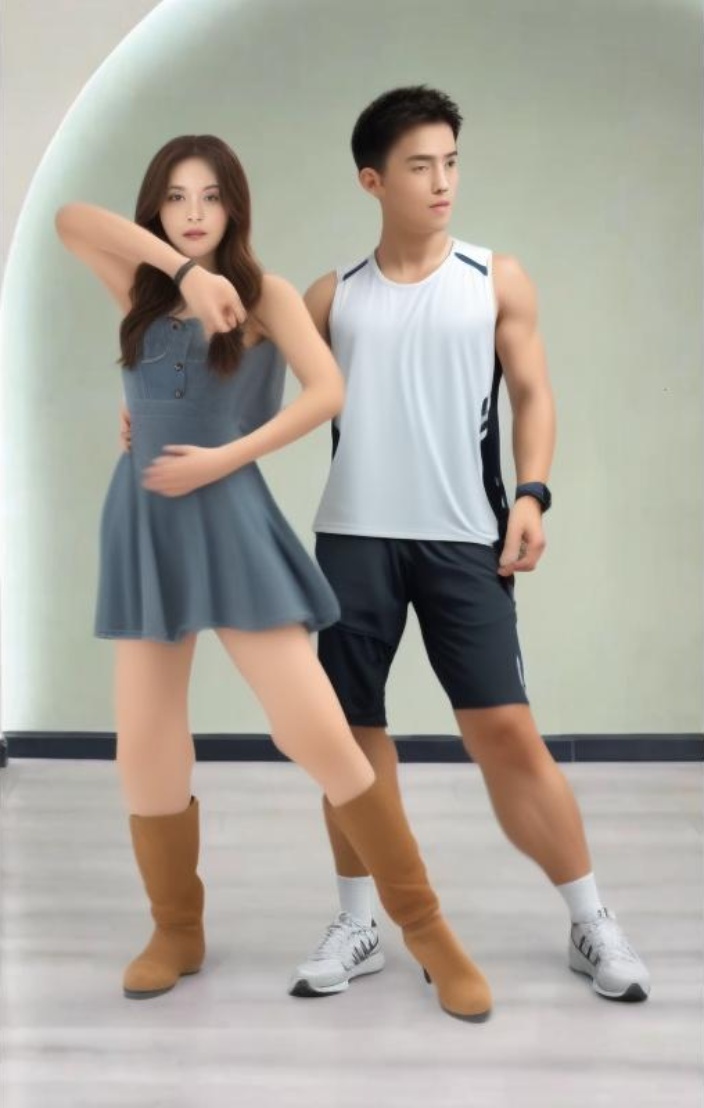}
    &\includegraphics[width=1.8cm]{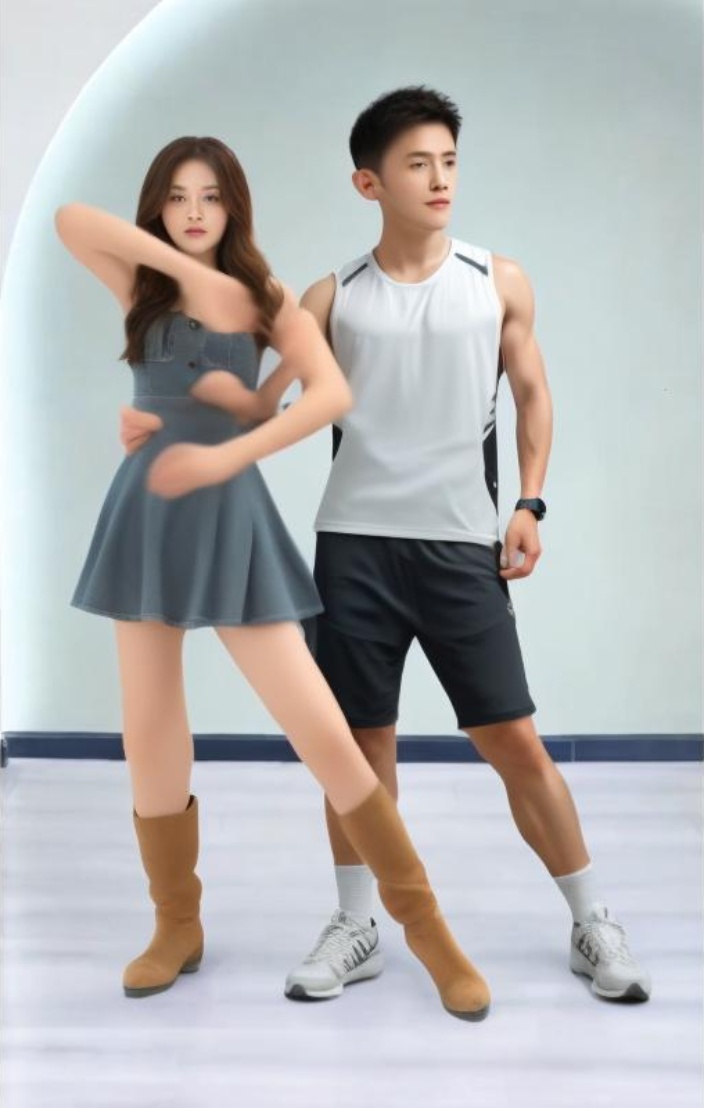} \\
    
    \end{tabular}
    \end{center}
    \caption{Ablation study on the interaction guider module. Interaction guider module enhances the framework's capability to generate accurate inter-character interactions, while depth order guider  fails and causes degradation on the background.} 
    \label{fig:ab-follow2}
\end{figure}

As Xue et al.~\cite{xue2025multiplecharacterimageanimation} utilized a depth order guider to establish plausible inter-character spatial relationships. This mechanism initially generates masks through the dilation of skeleton representations for distinct characters, subsequently producing a depth-order image derived from a depth map that determines the relative positioning of these mask regions. Notably, this approach does not explicitly encode identity information within the individual mask regions. We substituted our interaction guider with their approach, and the comparative results are presented in Figure~\ref{fig:ab-follow2}. 

We can observe from the figure that while the depth order guider provides spatial relationship representation, its guidance remains insufficiently granular. Specifically, when processing intricate character interactions, such as the hugging action illustrated in the figure, the depth order guider fails to provide accurate spatial guidance. Furthermore, this mechanism introduces interference with background generation in our framework, presumably because its multi-colored compositional nature, spanning a broad value range, potentially affects the latent background representation when incorporated into the initial Gaussian noise. In contrast, our proposed interaction guider minimizes perturbations to both skeleton representations and overall frame composition, as well as enabling accurate synthesis of inter-character interactions.

\subsection{Evaluation on Single Character Benchmark}

To better assess the performance of our method, we conducted experiments on single-character animation using the TikTok dataset~\cite{Jafarian_2021_CVPR_TikTok}. The quantitative results are presented in Table~\ref{tab:quan-single}. We obtained the results for the comparison methods directly from their original paper, since MimicMotion and Moore-Animate did not report their performance on TikTok dataset, we use their offical pre-trained model and conduct evaluation on the TikTok dataset. Moreover, UniAnimate did not report their FID on TikTok dataset. These results demonstrate that although our method is designed for handling multi-character animation, it can be seamlessly applied to single-character tasks and achieves performance comparable to state-of-the-art methods.

\begin{table}
  \caption{Quantitative comparison on single character animation, best results are in \textbf{bold}. Our method achieves performance comparable to state-of-the-art methods. The MSE results are presented after being multiplied by $10^5$.}
  \label{tab:quan-single}
  \centering
  \begin{tabular}{ccccc}
  \hline
  & SSIM $\uparrow$ & LPIPS $\downarrow$ & MSE  $\downarrow$ & FID $\downarrow$ \\
  \hline
  Moore-Animate~\cite{hu2023animateanyone} & 0.702 & \textbf{0.186} & 2.75 & 121.97\\
  MimicMotion~\cite{mimicmotion2024} & 0.615 & 0.339 & 5.59 & 107.32\\
  MagicPose~\cite{chang2023magicdance} & 0.752 & 0.292 & 8.10 & \textbf{25.50}\\
  MagicAnimate~\cite{xu2023magicanimate} & 0.714 & 0.239 & 3.13 & 32.09\\
  UniAnimate~\cite{wang2024unianimate} & \textbf{0.801} & 0.231 & 2.66 & - \\
  Ours & 0.759 & 0.195 & \textbf{2.42} & 42.03\\
  \hline
  \end{tabular}
\end{table}

\section{Discussion.} 
\label{discussion}
Our framework occasionally introduces artifacts, particularly in highly detailed regions such as hands, due to inherent limitations in the base model's generation capabilities. Recent video diffusion models~\cite{wan2025,kong2024hunyuanvideo,yang2025cogvideoxtexttovideodiffusionmodels} base on DiT~\cite{Peebles2022DiT} have demonstrated superior video generation capabilities compared to their UNet-based counterparts. We have been able to significantly mitigate these artifacts through simple post-processing methods, such as temporal smoothing. Future iterations of our work will explore the integration of our framework with these more advanced DiT-based architectures. Additionally, we acknowledge potential ethical concerns regarding the misuse of our framework for generating deceptive content. Therefore, we emphasize the importance of implementing appropriate usage controls and guidelines.

\section{Conclusion.} 
\label{sec:conclusion}
Our paper presents MultiAnimate, a comprehensive framework that enables concurrent animation of multiple characters within a shared environment using distinct reference images and their corresponding pose sequences. The framework extends the conventional reference net through the introduction of position encoding, enabling precise feature extraction and differentiation from multiple reference images. Our identity-aware pose encoder establishes robust character-pose bindings, ensuring consistent character animation throughout the generated sequence. Furthermore, we incorporate an interaction guider module that enhances the framework's capability to synthesize accurate inter-character interactions. Through extensive experimental evaluations and ablation studies, we demonstrate our framework's ability to generate high-quality animation results while validating the effectiveness of each architectural component.

\bmhead{Acknowledgements}

The authors would like to thank the members of our research laboratory for their helpful discussions and advices during this paper.

\section*{Declarations}

\subsection*{Funding}
This work was supported in part by the Natural Science Foundation of China under Grants 62372423,62121002,62072421, and was also supported by the Fundamental Research Funds for the Central Universities WK2100250070.

\subsection*{Conflict of Interest}
The authors declare that they have no competing interests.

\subsection*{Data Availability}
The multi-character dataset used in this paper was collected from publicly available online sources. We will release this dataset as open source upon acceptance of the paper to further benefit the research community. In addition, the TikTok dataset used in our experiments is publicly available and has been fully cited in the manuscript.

\subsection*{Code Availability}
The code used in this study will be released upon publication to support reproducibility and further research.

\subsection*{Author Contributions}
Zhongyi conceived the research idea, developed the proposed methodology, performed the experiments, and contributed to writing the manuscript. 
Guangyuan Wang contributed to the project through data collection and preprocessing. 
Li Hu, Tianyi Wei, Peng Zhang, Bang Zhang, Wenbo Zhou, Weiming Zhang and Nenghai Yu provided scientific guidance, supervised the research design and experimental validation, and assisted in revising the manuscript. All authors reviewed and approved the final version of the manuscript.

\subsection*{Reproducibility}
We ensure reproducibility by providing clear distinctions between results and interpretations, along with detailed reporting of implementation components such as evaluation metrics, and computing resources. 

\subsection*{Ethical Considerations in Character Image Animation}
The method proposed in this paper can be applied to generating videos of human motions, with broad application prospects in areas such as cinematic content creation and entertainment. However, if misused, it may raise ethical concerns, for example through the creation of non-existent or falsified videos for malicious purposes. We advocate the responsible development and deployment of such technologies, including safeguards such as embedding watermarks in generated content.

\bibliography{ref}% common bib file

@article{hu2023animateanyone,
  title={Animate Anyone: Consistent and Controllable Image-to-Video Synthesis for Character Animation},
  author={Li Hu and Xin Gao and Peng Zhang and Ke Sun and Bang Zhang and Liefeng Bo},
  journal={arXiv preprint arXiv:2311.17117},
  website={https://humanaigc.github.io/animate-anyone/},
  year={2023}
}

@inproceedings{
wang2022latent,
title={Latent Image Animator: Learning to Animate Images via Latent Space Navigation},
author={Yaohui Wang and Di Yang and Francois Bremond and Antitza Dantcheva},
booktitle={International Conference on Learning Representations},
year={2022}
}

@InProceedings{Siarohin_2019_NeurIPS,
  author={Siarohin, Aliaksandr and Lathuilière, Stéphane and Tulyakov, Sergey and Ricci, Elisa and Sebe, Nicu},
  title={First Order Motion Model for Image Animation},
  booktitle = {Conference on Neural Information Processing Systems (NeurIPS)},
  month = {December},
  year = {2019}
}

@InProceedings{Gao_2023_CVPR,
    author    = {Gao, Yue and Zhou, Yuan and Wang, Jinglu and Li, Xiao and Ming, Xiang and Lu, Yan},
    title     = {High-Fidelity and Freely Controllable Talking Head Video Generation},
    booktitle = {Proceedings of the IEEE/CVF Conference on Computer Vision and Pattern Recognition (CVPR)},
    month     = {June},
    year      = {2023},
    pages     = {5609-5619}
}

@misc{blattmann2023stablevideodiffusionscaling,
      title={Stable Video Diffusion: Scaling Latent Video Diffusion Models to Large Datasets}, 
      author={Andreas Blattmann and Tim Dockhorn and Sumith Kulal and Daniel Mendelevitch and Maciej Kilian and Dominik Lorenz and Yam Levi and Zion English and Vikram Voleti and Adam Letts and Varun Jampani and Robin Rombach},
      year={2023},
      eprint={2311.15127},
      archivePrefix={arXiv},
      primaryClass={cs.CV},
      url={https://arxiv.org/abs/2311.15127}, 
}

@article{guo2023animatediff,
  title={AnimateDiff: Animate Your Personalized Text-to-Image Diffusion Models without Specific Tuning},
  author={Guo, Yuwei and Yang, Ceyuan and Rao, Anyi and Liang, Zhengyang and Wang, Yaohui and Qiao, Yu and Agrawala, Maneesh and Lin, Dahua and Dai, Bo},
  journal={International Conference on Learning Representations},
  year={2024}
}

@misc{xue2025multiplecharacterimageanimation,
      title={Towards Multiple Character Image Animation Through Enhancing Implicit Decoupling}, 
      author={Jingyun Xue and Hongfa Wang and Qi Tian and Yue Ma and Andong Wang and Zhiyuan Zhao and Shaobo Min and Wenzhe Zhao and Kaihao Zhang and Heung-Yeung Shum and Wei Liu and Mengyang Liu and Wenhan Luo},
      year={2025},
      eprint={2406.03035},
      archivePrefix={arXiv},
      primaryClass={cs.CV},
      url={https://arxiv.org/abs/2406.03035}, 
}

@article{ma2023follow,
  title={Follow Your Pose: Pose-Guided Text-to-Video Generation using Pose-Free Videos},
  author={Ma, Yue and He, Yingqing and Cun, Xiaodong and Wang, Xintao and Shan, Ying and Li, Xiu and Chen, Qifeng},
  journal={arXiv preprint arXiv:2304.01186},
  year={2023}
}

@misc{karras2023dreamposefashionimagetovideosynthesis,
      title={DreamPose: Fashion Image-to-Video Synthesis via Stable Diffusion}, 
      author={Johanna Karras and Aleksander Holynski and Ting-Chun Wang and Ira Kemelmacher-Shlizerman},
      year={2023},
      eprint={2304.06025},
      archivePrefix={arXiv},
      primaryClass={cs.CV},
      url={https://arxiv.org/abs/2304.06025}, 
}

@article{wang2023disco,
  title={Disco: Disentangled control for realistic human dance generation},
  author={Wang, Tan and Li, Linjie and Lin, Kevin and Zhai, Yuanhao and Lin, Chung-Ching and Yang, Zhengyuan and Zhang, Hanwang and Liu, Zicheng and Wang, Lijuan},
  journal={arXiv preprint arXiv:2307.00040},
  year={2023}
}

@article{chang2023magicdance,
  title={MagicDance: Realistic Human Dance Video Generation with Motions \& Facial Expressions Transfer},
  author={Chang, Di and Shi, Yichun and Gao, Quankai and Fu, Jessica and Xu, Hongyi and Song, Guoxian and Yan, Qing and Yang, Xiao and Soleymani, Mohammad},
  journal={arXiv preprint arXiv:2311.12052},
  year={2023}
}

@inproceedings{xu2023magicanimate,
    author    = {Xu, Zhongcong and Zhang, Jianfeng and Liew, Jun Hao and Yan, Hanshu and Liu, Jia-Wei and Zhang, Chenxu and Feng, Jiashi and Shou, Mike Zheng},
    title     = {MagicAnimate: Temporally Consistent Human Image Animation using Diffusion Model},
    booktitle = {arXiv},
    year      = {2023}
}

@inproceedings{zhu2024champ,
      title={Champ: Controllable and Consistent Human Image Animation with 3D Parametric Guidance},
      author={Shenhao Zhu and Junming Leo Chen and Zuozhuo Dai and Yinghui Xu and Xun Cao and Yao Yao and Hao Zhu and Siyu Zhu},
      booktitle={European Conference on Computer Vision (ECCV)},
      year={2024}
}

@article{mimicmotion2024,
  title={MimicMotion: High-Quality Human Motion Video Generation with Confidence-aware Pose Guidance},
  author={Yuang Zhang and Jiaxi Gu and Li-Wen Wang and Han Wang and Junqi Cheng and Yuefeng Zhu and Fangyuan Zou},
  journal={arXiv preprint arXiv:2406.19680},
  year={2024}
}

@misc{wang2025multiidentityhumanimageanimation,
      title={Multi-identity Human Image Animation with Structural Video Diffusion}, 
      author={Zhenzhi Wang and Yixuan Li and Yanhong Zeng and Yuwei Guo and Dahua Lin and Tianfan Xue and Bo Dai},
      year={2025},
      eprint={2504.04126},
      archivePrefix={arXiv},
      primaryClass={cs.CV},
      url={https://arxiv.org/abs/2504.04126}, 
}

@misc{kim2024tcananimatinghumanimages,
      title={TCAN: Animating Human Images with Temporally Consistent Pose Guidance using Diffusion Models}, 
      author={Jeongho Kim and Min-Jung Kim and Junsoo Lee and Jaegul Choo},
      year={2024},
      eprint={2407.09012},
      archivePrefix={arXiv},
      primaryClass={cs.CV},
      url={https://arxiv.org/abs/2407.09012}, 
}

@article{wang2024unianimate,
      title={UniAnimate: Taming Unified Video Diffusion Models for Consistent Human Image Animation},
      author={Wang, Xiang and Zhang, Shiwei and Gao, Changxin and Wang, Jiayu and Zhou, Xiaoqiang and Zhang, Yingya and Yan, Luxin and Sang, Nong},
      journal={arXiv preprint arXiv:2406.01188},
      year={2024}
}

@misc{yoon2025tpctesttimeprocrustescalibration,
      title={TPC: Test-time Procrustes Calibration for Diffusion-based Human Image Animation}, 
      author={Sunjae Yoon and Gwanhyeong Koo and Younghwan Lee and Chang D. Yoo},
      year={2025},
      eprint={2410.24037},
      archivePrefix={arXiv},
      primaryClass={cs.CV},
      url={https://arxiv.org/abs/2410.24037}, 
}

@misc{men2024mimocontrollablecharactervideo,
      title={MIMO: Controllable Character Video Synthesis with Spatial Decomposed Modeling}, 
      author={Yifang Men and Yuan Yao and Miaomiao Cui and Liefeng Bo},
      year={2024},
      eprint={2409.16160},
      archivePrefix={arXiv},
      primaryClass={cs.CV},
      url={https://arxiv.org/abs/2409.16160}, 
}

@misc{rombach2022highresolutionimagesynthesislatent,
      title={High-Resolution Image Synthesis with Latent Diffusion Models}, 
      author={Robin Rombach and Andreas Blattmann and Dominik Lorenz and Patrick Esser and Björn Ommer},
      year={2022},
      eprint={2112.10752},
      archivePrefix={arXiv},
      primaryClass={cs.CV},
      url={https://arxiv.org/abs/2112.10752}, 
}

@misc{gu2024mambalineartimesequencemodeling,
      title={Mamba: Linear-Time Sequence Modeling with Selective State Spaces}, 
      author={Albert Gu and Tri Dao},
      year={2024},
      eprint={2312.00752},
      archivePrefix={arXiv},
      primaryClass={cs.LG},
      url={https://arxiv.org/abs/2312.00752}, 
}

@incollection{loper2023smpl,
  title={SMPL: A skinned multi-person linear model},
  author={Loper, Matthew and Mahmood, Naureen and Romero, Javier and Pons-Moll, Gerard and Black, Michael J},
  booktitle={Seminal Graphics Papers: Pushing the Boundaries, Volume 2},
  pages={851--866},
  year={2023}
}

@misc{chan2019everybodydance,
      title={Everybody Dance Now}, 
      author={Caroline Chan and Shiry Ginosar and Tinghui Zhou and Alexei A. Efros},
      year={2019},
      eprint={1808.07371},
      archivePrefix={arXiv},
      primaryClass={cs.GR},
      url={https://arxiv.org/abs/1808.07371}, 
}

@misc{siarohin2021motionrepresentationsarticulatedanimation,
      title={Motion Representations for Articulated Animation}, 
      author={Aliaksandr Siarohin and Oliver J. Woodford and Jian Ren and Menglei Chai and Sergey Tulyakov},
      year={2021},
      eprint={2104.11280},
      archivePrefix={arXiv},
      primaryClass={cs.CV},
      url={https://arxiv.org/abs/2104.11280}, 
}

@misc{zhao2022thinplatesplinemotionmodel,
      title={Thin-Plate Spline Motion Model for Image Animation}, 
      author={Jian Zhao and Hui Zhang},
      year={2022},
      eprint={2203.14367},
      archivePrefix={arXiv},
      primaryClass={cs.CV},
      url={https://arxiv.org/abs/2203.14367}, 
}

@misc{zhang2022exploringdualtaskcorrelationpose,
      title={Exploring Dual-task Correlation for Pose Guided Person Image Generation}, 
      author={Pengze Zhang and Lingxiao Yang and Jianhuang Lai and Xiaohua Xie},
      year={2022},
      eprint={2203.02910},
      archivePrefix={arXiv},
      primaryClass={cs.CV},
      url={https://arxiv.org/abs/2203.02910}, 
}

@misc{liang2025movieweavertuningfreemulticoncept,
      title={Movie Weaver: Tuning-Free Multi-Concept Video Personalization with Anchored Prompts}, 
      author={Feng Liang and Haoyu Ma and Zecheng He and Tingbo Hou and Ji Hou and Kunpeng Li and Xiaoliang Dai and Felix Juefei-Xu and Samaneh Azadi and Animesh Sinha and Peizhao Zhang and Peter Vajda and Diana Marculescu},
      year={2025},
      eprint={2502.07802},
      archivePrefix={arXiv},
      primaryClass={cs.CV},
      url={https://arxiv.org/abs/2502.07802}, 
}

@misc{kumari2023multiconceptcustomizationtexttoimagediffusion,
      title={Multi-Concept Customization of Text-to-Image Diffusion}, 
      author={Nupur Kumari and Bingliang Zhang and Richard Zhang and Eli Shechtman and Jun-Yan Zhu},
      year={2023},
      eprint={2212.04488},
      archivePrefix={arXiv},
      primaryClass={cs.CV},
      url={https://arxiv.org/abs/2212.04488}, 
}

@inproceedings{Avrahami_2023, series={SA ’23},
   title={Break-A-Scene: Extracting Multiple Concepts from a Single Image},
   url={http://dx.doi.org/10.1145/3610548.3618154},
   DOI={10.1145/3610548.3618154},
   booktitle={SIGGRAPH Asia 2023 Conference Papers},
   publisher={ACM},
   author={Avrahami, Omri and Aberman, Kfir and Fried, Ohad and Cohen-Or, Daniel and Lischinski, Dani},
   year={2023},
   month=dec, pages={1–12},
   collection={SA ’23} }

@misc{kwon2024conceptweaverenablingmulticoncept,
      title={Concept Weaver: Enabling Multi-Concept Fusion in Text-to-Image Models}, 
      author={Gihyun Kwon and Simon Jenni and Dingzeyu Li and Joon-Young Lee and Jong Chul Ye and Fabian Caba Heilbron},
      year={2024},
      eprint={2404.03913},
      archivePrefix={arXiv},
      primaryClass={cs.CV},
      url={https://arxiv.org/abs/2404.03913}, 
}

@misc{jang2024identitydecouplingmultisubjectpersonalization,
      title={Identity Decoupling for Multi-Subject Personalization of Text-to-Image Models}, 
      author={Sangwon Jang and Jaehyeong Jo and Kimin Lee and Sung Ju Hwang},
      year={2024},
      eprint={2404.04243},
      archivePrefix={arXiv},
      primaryClass={cs.CV},
      url={https://arxiv.org/abs/2404.04243}, 
}

@misc{xiao2023fastcomposertuningfreemultisubjectimage,
      title={FastComposer: Tuning-Free Multi-Subject Image Generation with Localized Attention}, 
      author={Guangxuan Xiao and Tianwei Yin and William T. Freeman and Frédo Durand and Song Han},
      year={2023},
      eprint={2305.10431},
      archivePrefix={arXiv},
      primaryClass={cs.CV},
      url={https://arxiv.org/abs/2305.10431}, 
}

@misc{wang2024moamixtureofattentionsubjectcontextdisentanglement,
      title={MoA: Mixture-of-Attention for Subject-Context Disentanglement in Personalized Image Generation}, 
      author={Kuan-Chieh Wang and Daniil Ostashev and Yuwei Fang and Sergey Tulyakov and Kfir Aberman},
      year={2024},
      eprint={2404.11565},
      archivePrefix={arXiv},
      primaryClass={cs.CV},
      url={https://arxiv.org/abs/2404.11565}, 
}

@misc{kim2024instantfamilymaskedattentionzeroshot,
      title={InstantFamily: Masked Attention for Zero-shot Multi-ID Image Generation}, 
      author={Chanran Kim and Jeongin Lee and Shichang Joung and Bongmo Kim and Yeul-Min Baek},
      year={2024},
      eprint={2404.19427},
      archivePrefix={arXiv},
      primaryClass={cs.CV},
      url={https://arxiv.org/abs/2404.19427}, 
}

@misc{he2024uniportraitunifiedframeworkidentitypreserving,
      title={UniPortrait: A Unified Framework for Identity-Preserving Single- and Multi-Human Image Personalization}, 
      author={Junjie He and Yifeng Geng and Liefeng Bo},
      year={2024},
      eprint={2408.05939},
      archivePrefix={arXiv},
      primaryClass={cs.CV},
      url={https://arxiv.org/abs/2408.05939}, 
}

@misc{gu2023mixofshowdecentralizedlowrankadaptation,
      title={Mix-of-Show: Decentralized Low-Rank Adaptation for Multi-Concept Customization of Diffusion Models}, 
      author={Yuchao Gu and Xintao Wang and Jay Zhangjie Wu and Yujun Shi and Yunpeng Chen and Zihan Fan and Wuyou Xiao and Rui Zhao and Shuning Chang and Weijia Wu and Yixiao Ge and Ying Shan and Mike Zheng Shou},
      year={2023},
      eprint={2305.18292},
      archivePrefix={arXiv},
      primaryClass={cs.CV},
      url={https://arxiv.org/abs/2305.18292}, 
}

@misc{ronneberger2015unetconvolutionalnetworksbiomedical,
      title={U-Net: Convolutional Networks for Biomedical Image Segmentation}, 
      author={Olaf Ronneberger and Philipp Fischer and Thomas Brox},
      year={2015},
      eprint={1505.04597},
      archivePrefix={arXiv},
      primaryClass={cs.CV},
      url={https://arxiv.org/abs/1505.04597}, 
}

@misc{kingma2022autoencodingvariationalbayes,
      title={Auto-Encoding Variational Bayes}, 
      author={Diederik P Kingma and Max Welling},
      year={2022},
      eprint={1312.6114},
      archivePrefix={arXiv},
      primaryClass={stat.ML},
      url={https://arxiv.org/abs/1312.6114}, 
}

@misc{oord2018neuraldiscreterepresentationlearning,
      title={Neural Discrete Representation Learning}, 
      author={Aaron van den Oord and Oriol Vinyals and Koray Kavukcuoglu},
      year={2018},
      eprint={1711.00937},
      archivePrefix={arXiv},
      primaryClass={cs.LG},
      url={https://arxiv.org/abs/1711.00937}, 
}

@misc{tan2024animatexuniversalcharacterimage,
      title={Animate-X: Universal Character Image Animation with Enhanced Motion Representation}, 
      author={Shuai Tan and Biao Gong and Xiang Wang and Shiwei Zhang and Dandan Zheng and Ruobing Zheng and Kecheng Zheng and Jingdong Chen and Ming Yang},
      year={2024},
      eprint={2410.10306},
      archivePrefix={arXiv},
      primaryClass={cs.CV},
      url={https://arxiv.org/abs/2410.10306}, 
}

@article{Peebles2022DiT,
  title={Scalable Diffusion Models with Transformers},
  author={William Peebles and Saining Xie},
  year={2022},
  journal={arXiv preprint arXiv:2212.09748},
}

@inproceedings{yang2023effective,
  title={Effective whole-body pose estimation with two-stages distillation},
  author={Yang, Zhendong and Zeng, Ailing and Yuan, Chun and Li, Yu},
  booktitle={Proceedings of the IEEE/CVF International Conference on Computer Vision},
  pages={4210--4220},
  year={2023}
}

@inproceedings{Guler2018DensePose,
  title={DensePose: Dense Human Pose Estimation In The Wild},
  author={R{i}za Alp G{"u}ler, Natalia Neverova, Iasonas Kokkinos},
  journal={The IEEE Conference on Computer Vision and Pattern Recognition (CVPR)},
  year={2018}
}

@misc{cao2019openposerealtimemultiperson2d,
      title={OpenPose: Realtime Multi-Person 2D Pose Estimation using Part Affinity Fields}, 
      author={Zhe Cao and Gines Hidalgo and Tomas Simon and Shih-En Wei and Yaser Sheikh},
      year={2019},
      eprint={1812.08008},
      archivePrefix={arXiv},
      primaryClass={cs.CV},
      url={https://arxiv.org/abs/1812.08008}, 
}

@article{ravi2024sam2,
  title={SAM 2: Segment Anything in Images and Videos},
  author={Ravi, Nikhila and Gabeur, Valentin and Hu, Yuan-Ting and Hu, Ronghang and Ryali, Chaitanya and Ma, Tengyu and Khedr, Haitham and R{\"a}dle, Roman and Rolland, Chloe and Gustafson, Laura and Mintun, Eric and Pan, Junting and Alwala, Kalyan Vasudev and Carion, Nicolas and Wu, Chao-Yuan and Girshick, Ross and Doll{\'a}r, Piotr and Feichtenhofer, Christoph},
  journal={arXiv preprint arXiv:2408.00714},
  url={https://arxiv.org/abs/2408.00714},
  year={2024}
}

@INPROCEEDINGS{5596999,
  author={Horé, Alain and Ziou, Djemel},
  booktitle={2010 20th International Conference on Pattern Recognition}, 
  title={Image Quality Metrics: PSNR vs. SSIM}, 
  year={2010},
  volume={},
  number={},
  pages={2366-2369},
  doi={10.1109/ICPR.2010.579}}

@ARTICLE{1284395,
  author={Zhou Wang and Bovik, A.C. and Sheikh, H.R. and Simoncelli, E.P.},
  journal={IEEE Transactions on Image Processing}, 
  title={Image quality assessment: from error visibility to structural similarity}, 
  year={2004},
  volume={13},
  number={4},
  pages={600-612},
  doi={10.1109/TIP.2003.819861}}

@misc{zhang2018unreasonableeffectivenessdeepfeatures,
      title={The Unreasonable Effectiveness of Deep Features as a Perceptual Metric}, 
      author={Richard Zhang and Phillip Isola and Alexei A. Efros and Eli Shechtman and Oliver Wang},
      year={2018},
      eprint={1801.03924},
      archivePrefix={arXiv},
      primaryClass={cs.CV},
      url={https://arxiv.org/abs/1801.03924}, 
}

@misc{heusel2018ganstrainedtimescaleupdate,
      title={GANs Trained by a Two Time-Scale Update Rule Converge to a Local Nash Equilibrium}, 
      author={Martin Heusel and Hubert Ramsauer and Thomas Unterthiner and Bernhard Nessler and Sepp Hochreiter},
      year={2018},
      eprint={1706.08500},
      archivePrefix={arXiv},
      primaryClass={cs.LG},
      url={https://arxiv.org/abs/1706.08500}, 
}

@misc{unterthiner2019accurategenerativemodelsvideo,
      title={Towards Accurate Generative Models of Video: A New Metric \& Challenges}, 
      author={Thomas Unterthiner and Sjoerd van Steenkiste and Karol Kurach and Raphael Marinier and Marcin Michalski and Sylvain Gelly},
      year={2019},
      eprint={1812.01717},
      archivePrefix={arXiv},
      primaryClass={cs.CV},
      url={https://arxiv.org/abs/1812.01717}, 
}

@article{wan2025,
      title={Wan: Open and Advanced Large-Scale Video Generative Models}, 
      author={Ang Wang and Baole Ai and Bin Wen and Chaojie Mao and Chen-Wei Xie and Di Chen and Feiwu Yu and Haiming Zhao and Jianxiao Yang and Jianyuan Zeng and Jiayu Wang and Jingfeng Zhang and Jingren Zhou and Jinkai Wang and Jixuan Chen and Kai Zhu and Kang Zhao and Keyu Yan and Lianghua Huang and Mengyang Feng and Ningyi Zhang and Pandeng Li and Pingyu Wu and Ruihang Chu and Ruili Feng and Shiwei Zhang and Siyang Sun and Tao Fang and Tianxing Wang and Tianyi Gui and Tingyu Weng and Tong Shen and Wei Lin and Wei Wang and Wei Wang and Wenmeng Zhou and Wente Wang and Wenting Shen and Wenyuan Yu and Xianzhong Shi and Xiaoming Huang and Xin Xu and Yan Kou and Yangyu Lv and Yifei Li and Yijing Liu and Yiming Wang and Yingya Zhang and Yitong Huang and Yong Li and You Wu and Yu Liu and Yulin Pan and Yun Zheng and Yuntao Hong and Yupeng Shi and Yutong Feng and Zeyinzi Jiang and Zhen Han and Zhi-Fan Wu and Ziyu Liu},
      journal = {arXiv preprint arXiv:2503.20314},
      year={2025}
}

@article{kong2024hunyuanvideo,
  title={Hunyuanvideo: A systematic framework for large video generative models},
  author={Kong, Weijie and Tian, Qi and Zhang, Zijian and Min, Rox and Dai, Zuozhuo and Zhou, Jin and Xiong, Jiangfeng and Li, Xin and Wu, Bo and Zhang, Jianwei and others},
  journal={arXiv preprint arXiv:2412.03603},
  year={2024}
}

@misc{yang2025cogvideoxtexttovideodiffusionmodels,
      title={CogVideoX: Text-to-Video Diffusion Models with An Expert Transformer}, 
      author={Zhuoyi Yang and Jiayan Teng and Wendi Zheng and Ming Ding and Shiyu Huang and Jiazheng Xu and Yuanming Yang and Wenyi Hong and Xiaohan Zhang and Guanyu Feng and Da Yin and Yuxuan Zhang and Weihan Wang and Yean Cheng and Bin Xu and Xiaotao Gu and Yuxiao Dong and Jie Tang},
      year={2025},
      eprint={2408.06072},
      archivePrefix={arXiv},
      primaryClass={cs.CV},
      url={https://arxiv.org/abs/2408.06072}, 
}

@article{yolox2021,
  title={YOLOX: Exceeding YOLO Series in 2021},
  author={Ge, Zheng and Liu, Songtao and Wang, Feng and Li, Zeming and Sun, Jian},
  journal={arXiv preprint arXiv:2107.08430},
  year={2021}
}

@InProceedings{Jafarian_2021_CVPR_TikTok,
    author    = {Jafarian, Yasamin and Park, Hyun Soo},
    title     = {Learning High Fidelity Depths of Dressed Humans by Watching Social Media Dance Videos},
    booktitle = {Proceedings of the IEEE/CVF Conference on Computer Vision and Pattern Recognition (CVPR)},
    month     = {June},
    year      = {2021},
    pages     = {12753-12762}}

@misc{wei2025freefluxunderstandingexploitinglayerspecific,
      title={FreeFlux: Understanding and Exploiting Layer-Specific Roles in RoPE-Based MMDiT for Versatile Image Editing}, 
      author={Tianyi Wei and Yifan Zhou and Dongdong Chen and Xingang Pan},
      year={2025},
      eprint={2503.16153},
      archivePrefix={arXiv},
      primaryClass={cs.CV},
      url={https://arxiv.org/abs/2503.16153}, 
}

@misc{labs2025flux1kontextflowmatching,
      title={FLUX.1 Kontext: Flow Matching for In-Context Image Generation and Editing in Latent Space},
      author={Black Forest Labs and Stephen Batifol and Andreas Blattmann and Frederic Boesel and Saksham Consul and Cyril Diagne and Tim Dockhorn and Jack English and Zion English and Patrick Esser and Sumith Kulal and Kyle Lacey and Yam Levi and Cheng Li and Dominik Lorenz and Jonas Müller and Dustin Podell and Robin Rombach and Harry Saini and Axel Sauer and Luke Smith},
      year={2025},
      eprint={2506.15742},
      archivePrefix={arXiv},
      primaryClass={cs.GR},
      url={https://arxiv.org/abs/2506.15742},
}

@article{vougioukas2020realistic,
  title={Realistic speech-driven facial animation with gans},
  author={Vougioukas, Konstantinos and Petridis, Stavros and Pantic, Maja},
  journal={International Journal of Computer Vision},
  volume={128},
  number={5},
  pages={1398--1413},
  year={2020},
  publisher={Springer}
}

@article{pumarola2020ganimation,
  title={Ganimation: One-shot anatomically consistent facial animation},
  author={Pumarola, Albert and Agudo, Antonio and Martinez, Aleix M and Sanfeliu, Alberto and Moreno-Noguer, Francesc},
  journal={International Journal of Computer Vision},
  volume={128},
  number={3},
  pages={698--713},
  year={2020},
  publisher={Springer}
}

@article{wang2025leo,
  title={Leo: Generative latent image animator for human video synthesis},
  author={Wang, Yaohui and Ma, Xin and Chen, Xinyuan and Chen, Cunjian and Dantcheva, Antitza and Dai, Bo and Qiao, Yu},
  journal={International Journal of Computer Vision},
  volume={133},
  number={3},
  pages={1277--1289},
  year={2025},
  publisher={Springer}
}

@article{xiao2025fastcomposer,
  title={Fastcomposer: Tuning-free multi-subject image generation with localized attention},
  author={Xiao, Guangxuan and Yin, Tianwei and Freeman, William T and Durand, Fr{\'e}do and Han, Song},
  journal={International Journal of Computer Vision},
  volume={133},
  number={3},
  pages={1175--1194},
  year={2025},
  publisher={Springer}
}

@article{zhao2026freercustom,
  title={FreerCustom: Training-Free Multi-Concept Customization for Image and Video Generation: C. Zhao et al.},
  author={Zhao, Canyu and Ding, Ganggui and Wang, Wen and Yang, Zhen and Liu, Zide and Chen, Hao and Shen, Chunhua},
  journal={International Journal of Computer Vision},
  volume={134},
  number={1},
  pages={17},
  year={2026},
  publisher={Springer}
}

@article{hu2025lamd,
  title={LaMD: Latent Motion Diffusion for Image-Conditional Video Generation},
  author={Hu, Yaosi and Chen, Zhenzhong and Luo, Chong},
  journal={International Journal of Computer Vision},
  pages={1--17},
  year={2025},
  publisher={Springer}
}

\end{document}